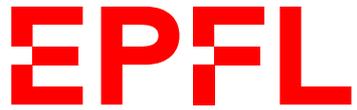 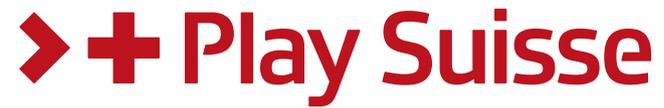

# Automating Video Thumbnails Selection and Generation with Multimodal and Multistage Analysis

A comprehensive AI approach for selecting varied engaging thumbnails for VOD Services

by Elia Fantini

## Master Thesis

Prof. Dr. Sabine Süsstrunk
Thesis Advisor

Dr. Gabriel Autès
External Expert

# Abstract

This thesis presents an innovative approach to automate the selection of potential thumbnails for videos, encompassing movies, documentaries, and TV series, with a focus on traditional broadcast video production content. Our methodology is predicated on the establishment of stringent criteria that prioritize a broad and diverse array of proposals, ensuring that the selected thumbnails are not only aesthetically pleasing but also faithfully representative of the video content. Important factors in our selection process include ensuring sufficient space for logo placement, incorporating vertical aspect ratios, and accurately recognizing facial identities and emotions.

In order to achieve these objectives, we introduce a sophisticated multistage pipeline. This pipeline is designed to meticulously select candidate frames from the video while also generating novel images by blending different foregrounds and backgrounds—either sourced directly from the video or synthesized using diffusion models. The pipeline incorporates a suite of state-of-the-art models, including downsampling, redundancy reduction, automated cropping, face recognition, closed-eye and emotion detection, shot scale and aesthetic prediction, segmentation, matting, and harmonization models. Furthermore, it leverages large language models and visual transformers to ensure semantic consistency. A graphical user interface (GUI) tool is developed to facilitate a rapid and intuitive navigation of the pipeline's output, significantly streamlining the selection process.

To address the inherently subjective nature of thumbnail evaluation, we conducted comprehensive experiments. In an initial study comprising 69 videos, findings revealed that 53.6% of our proposed set included thumbnails chosen by professional designers. Moreover, in 73.9% of instances, the proposed thumbnails contained images at least resembling the professionally selected ones. A subsequent survey involving 82 participants indicated a preference for our method 45.77% of the time, compared to 37.99% for manually chosen thumbnails and 16.36% for an alternative thumbnail selection method. Tests from professional thumbnail designers highlighted a 3.57-fold increase in the percentage of valid candidates found among the proposed set compared to the alternative method, and their feedback indicates that the method effectively fulfills the established criteria.

In conclusion, the findings of this thesis affirm that our proposed method not only accelerates the thumbnail creation process but also adheres to high-quality standards, thereby fostering greater user engagement.

# Résumé

Cette thèse présente une approche innovante pour automatiser la sélection de vignettes vidéos pour les films, documentaires et séries télévisées, avec un accent sur le contenu de production vidéo télévisuelle. Notre méthodologie repose sur l'établissement de critères stricts qui priorisent un large éventail de propositions diversifiées, garantissant que les vignettes sélectionnées sont non seulement esthétiquement agréables mais aussi fidèlement représentatives du contenu vidéo. Les facteurs importants dans notre processus de sélection incluent la présence d'un espace suffisant pour le placement du logo, l'incorporation de formats verticaux, et la reconnaissance précise des personnages et de leurs émotions.

Afin d'atteindre ces objectifs, nous introduisons un pipeline à plusieurs étapes. Ce pipeline est conçu pour sélectionner méticuleusement des vignettes er générer de nouvelles images en mélangeant différents premiers plans et arrière-plans—soit issus directement de la vidéo ou synthétisés en utilisant des modèles de diffusion. Le pipeline incorpore une sélection de modèles IA utilisés pour l'échantillonnage, la réduction de redondance, le recadrage automatisé, la reconnaissance faciale, la détection d'yeux fermés et d'émotions, la prédiction de l'échelle des plans et de l'esthétique, la segmentation, le détourage, et l'harmonisation. De plus, il tire parti des grands modèles de langage et des transformateurs visuels pour assurer une cohérence sémantique. Un outil d'interface graphique utilisateur (GUI) est développé pour faciliter une navigation rapide et intuitive des résultats du pipeline, simplifiant considérablement le processus de sélection.

Pour aborder la nature intrinsèquement subjective de l'évaluation des vignettes, nous avons mené des expériences exhaustives. Dans une étude initiale comprenant 69 vidéos, les résultats ont révélé que 53,6% de notre ensemble proposé comprenait des vignettes choisies par des créateurs de vignette professionnels. De plus, dans 73,9% des cas, les vignettes proposées contenaient des images ressemblant à celles sélectionnées professionnellement. Une enquête impliquant 82 participants a indiqué une préférence pour notre méthode 45,77% du temps, comparée à 37,99% pour les vignettes choisies manuellement et 16,36% pour une méthode alternative de sélection de vignettes. Les créateurs de vignettes professionnels ayant eu accès à notre outils ont reporté 3.57 fois plus de vignettes candidates valides trouvées comparé à une méthode alternative, et leurs retours indiquent que la méthode remplit efficacement les critères établis.

En conclusion, les résultats de cette thèse confirment que notre méthode proposée non seulement accélère le processus de création de vignettes mais adhère également à des normes de haute qualité, favorisant ainsi un plus grande engagement des utilisateurs.

# Acknowledgements

I am deeply thankful for the opportunity to conduct my master's thesis research within the vibrant and dynamic industries of entertainment and media creation, areas that I am genuinely passionate about. My experience at a forward-thinking company, where I felt immediately welcomed, has been nothing short of exhilarating.

I owe a special word of thanks to my company supervisor, Dr. Gabriel Autès, for his unwavering support and insightful feedback throughout my thesis project. He was a pillar of guidance and a key factor in the project's success. Equally, I am indebted to my academic supervisor at EPFL, Prof. Dr. Sabine Süsstrunk, for providing me with the necessary tools to kickstart my thesis and for her readiness to offer feedback and advice when needed.

I would also like to express my gratitude to Gaëtan Reine, the professional thumbnail designer, for his detailed feedback, which steered my thesis towards delivering maximum value to the company. Thanks are due to Francois Andries for his assistance with the downsampling deployment, and to the entire Play Suisse team for their varied contributions to my project.

My appreciation extends to EPFL for being an excellent source of education, networking, events, and opportunities, enriching both my academic journey and personal growth.

A heartfelt thanks to my family—my parents and my sister—for their unwavering belief in me and support throughout this challenging yet rewarding journey abroad. I am also immensely grateful for the incredible friends I've made in Switzerland, who have made this experience truly memorable through the highs and lows. I won't go through each one individually to keep this thesis from getting even longer, but know that they all are very important to me. A special shoutout to Lorenzo for being crazy enough to join me on a wonderful adventure in New York. I'm thankful for my friends from San Marino and those I met during my Bachelor's in Milan, who have been a part of my journey thus far and, I hope, will continue to be so in the future. Last but certainly not least, I want to thank Luana, for making my time during the master's program unforgettable.

*Lausanne, February 29, 2024*                                                                 E. F.

v

# Contents

# List of Figures

# List of Tables

# 1 Introduction

*This chapter provides a comprehensive introduction to the thesis, beginning with a background overview in Section 1.1, which sets the stage for understanding the core issue. The specific problem under investigation is detailed in Section 1.2, elucidating the challenges that need to be tackled. The significance of this research is highlighted in Section 1.3, addressing the 'why' behind the study's importance. The objectives and aims of this research are clearly outlined in Section 1.4, presenting the 'how' in terms of our approach to addressing the identified problem. The advantages of the proposed method, including its impact on stakeholders and the community, alongside considerations of ethics and sustainability, are discussed in Section 1.5. The methodology employed in this research is explained in Section 1.6. Section 1.7 introduces the stakeholders involved, while Section 1.8 delineates the project's scope and boundaries. Finally, Section 1.9 offers a concise outline of the thesis's content, guiding the reader through the structure of the document.*

## 1.1 Background

In the last decade, we have witnessed an unprecedented surge in video consumption, a trend heavily influenced by the advent of social media and video-sharing platforms. YouTube, a forerunner in this domain, exemplifies this growth. With over two billion users, YouTube experiences a daily consumption of one billion hours of video, underscoring the ever-increasing production of video content. This phenomenon is not confined to YouTube alone; the rise of Video on Demand (VOD) services has revolutionized how users engage with video content. VOD allows viewers to access a vast array of videos anytime and anywhere via Internet-enabled devices. Consequently, the volume of available videos has expanded, as has the time invested in selecting desirable content [57].

The escalating quality of video production further complicates this landscape. Platforms like YouTube maintain diverse standards, with many videos boasting high production values. This trend reaches its zenith in streaming services such as Netflix, Prime Video, and Disney+, where the budgets for video content are substantially larger. Regardless of the production scale, from social media clips to high-budget streaming service offerings, all videos rely on a pivotal element for viewer engagement: the thumbnail[13].

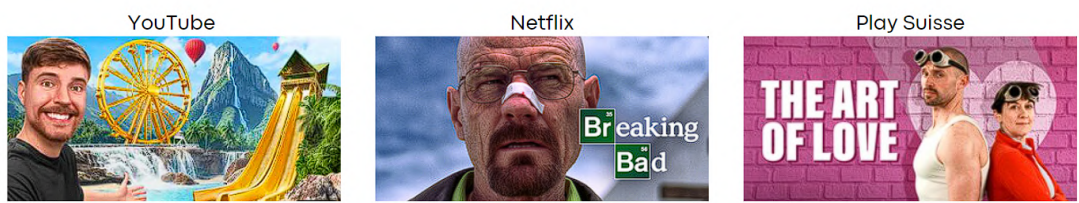

Figure 1.1: Examples of thumbnails and their different styles in different platforms: YouTube, Netflix, and Play Suisse.

Thumbnails serve as a critical gateway to video content, encapsulating a video's essence in a single image. Research indicates that thumbnails significantly influence viewers' decisions to watch or skip a video [22]. It is vital to catch interest at first sight, as the average attention span of viewers diminishes, partly due to the popularity of short-form platforms like TikTok, Instagram Reels, and YouTube Shorts. An attractive thumbnail not only increases the likelihood of a video being watched but also enhances ad revenue potential.

However, the appeal of a thumbnail can vary considerably based on the platform, the type of video, and the target audience. Thumbnails for movies, TV series, documentaries, and short-form content like YouTube videos differ markedly in their design and impact. A comparison of different styles used in different platforms is shown in Figure 1.1. In the high-stakes arena of streaming platforms, where content budgets soar, targeting a diverse audience with varying preferences is crucial.

The subjective nature of thumbnail appeal poses a significant challenge. The liking of an image varies from person to person, reflecting the broader theme of personalization that permeates various domains such as social media, AI, recommendation systems, and online shopping. Professional teams dedicated to thumbnail creation underscore the seriousness with which this aspect is treated in the industry.

Given the sheer volume of content and the time-intensive nature of selecting the best thumbnails tailored to user preferences, there is a growing interest in automating thumbnail extraction. While many methods have been proposed to address this challenge, the uniqueness and high stakes of representing traditional broadcast video production videos often necessitate manual selection by skilled professionals.

In this context, the development of an accurate, efficient, and automatic thumbnail selection system presents a promising avenue. Such a system could significantly streamline the process for video producers, editors, and hosting platforms, aligning with the evolving needs of the digital video landscape.

The domain of multimedia comprehension, especially concerning video content, constitutes a complex research domain that has undergone considerable evolution over time. Our literature review focuses on two pivotal aspects of multimedia comprehension: video summarization and thumbnail selection. Video summarization seeks to condense the essence of a video into concise formats, effectively encapsulating its narrative or thematic core. Thumbnail selection, the primary focus of our investigation, aims to capture the essence or allure of the video in a solitary, captivating image.

## 1.2 Problem

Play Suisse is a free Swiss video-on-demand service, encompassing productions and co-productions from the Swiss Radio and Television Company's (SRG) four corporate units: Radio Télévision Suisse (RTS), Schweizer Radio und Fernsehen (SRF), Radiotelevisione svizzera (RSI), and Radiotelevisiun Svizra Rumantscha (RTR). As a public media house, Play Suisse plays a crucial role in delivering a diverse audiovisual service to the public. The primary aim of Play Suisse is to foster cohesion between the different linguistic region of Switzerland. To realize this objective, our focus lies on enhancing user engagement and boosting overall viewership of Swiss content, through an innovative approach to thumbnail creation. This process involves automating the production of multiple thumbnails per video, aiming to captivate a broader audience segment by appealing to varied user interests, mirroring strategies used by major platforms like Netflix [28].

Contrary to the straightforward process of generating thumbnails for standard online videos, the task becomes significantly more challenging when dealing with traditional broadcast video production content. These productions not only require a higher standard of representation but also demand a nuanced approach to convey the quality and essence of the content effectively. In these cases, the thumbnail carries important weight: it must accurately represent the content, ensuring that foreground and background elements collectively provide a comprehensive preview of the video. This requirement stands in stark contrast to platforms like YouTube, where clickbait is more prevalent. Thumbnails for Play Suisse must truthfully represent the content quality, whether it showcases exceptional photography and direction or not, avoiding any exaggeration that misleads the viewer. Examples of thumbnails from Play Suisse are shown in Figure 1.2.

The task of generating the right number of thumbnail proposals presents a unique challenge. It's essential to provide the thumbnail designers at Play Suisse with a variety of options, enabling them to select the most appropriate thumbnails for different groups of viewers. However, the number of proposals must be carefully managed to avoid overwhelming the designers. Too many options could make the selection process tedious and counterproductive, while too few could limit the potential to accurately represent the diversity of content. This balance is critical in ensuring that the thumbnail generation tool is both efficient and effective.

In addition to the quantity of thumbnail proposals, the focus of their content is equally important. On different platforms, people-centric thumbnails have shown to be particularly effective in capturing viewer attention. This trend aligns with broader observations in the streaming industry, where human-focused imagery often drives higher engagement. However, this approach is not universally applicable. For instance, documentaries or other specialized genres may require a different strategy. Such content might benefit more from thumbnails that emphasize unique subject matter or thematic elements rather than focusing solely on human subjects.

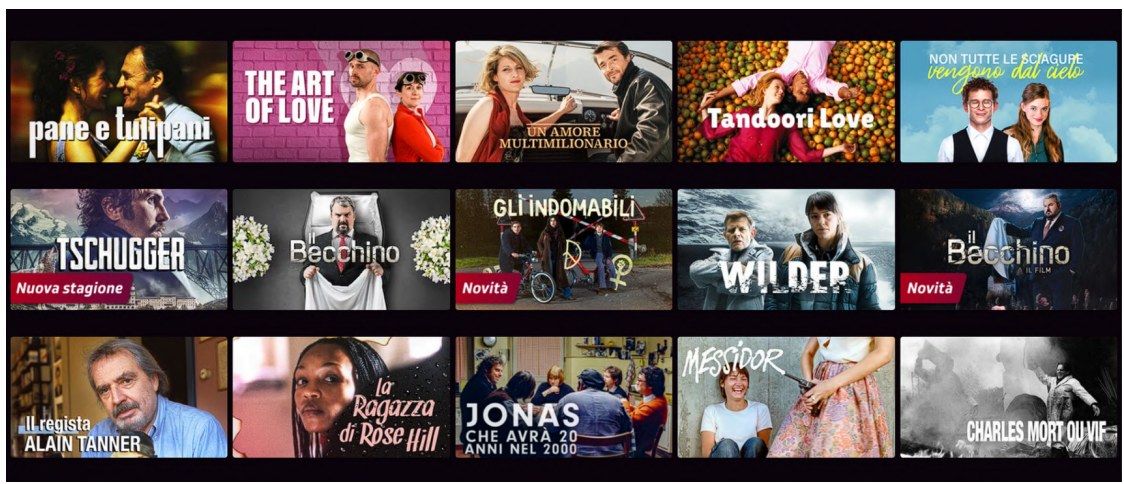

Figure 1.2: Examples of manually generated thumbnails from Play Suisse.

Furthermore, the evolving landscape of media consumption, particularly the rise of vertical content on social media platforms like Instagram and TikTok, has implications for thumbnail design. Play Suisse is considering the inclusion of vertical thumbnails to align with these changes in user behavior. This consideration is a response to the growing popularity of vertical viewing formats, which necessitates a different approach to thumbnail design—one that can effectively capture the essence of the content in a vertical frame.

Aesthetics play a fundamental role in thumbnail design. The initial visual appeal of a thumbnail can significantly influence user perception of content quality. The challenge lies in consistently extracting high-quality images from videos that vary in production quality.

Additionally, the integration of title placement into the thumbnail design is crucial. Professional designers at Play Suisse often create custom title logos to complement the selected image, necessitating a design that accommodates this additional element.

Lastly, the usability of the thumbnail generation tool for professional designers is paramount. Previous attempts at creating such tools have not been widely adopted due to their failure to meet the specific needs and criteria of the platform. This highlights the importance of developing a tool that is not only technologically advanced but also intuitive and aligned with the designers' workflow.

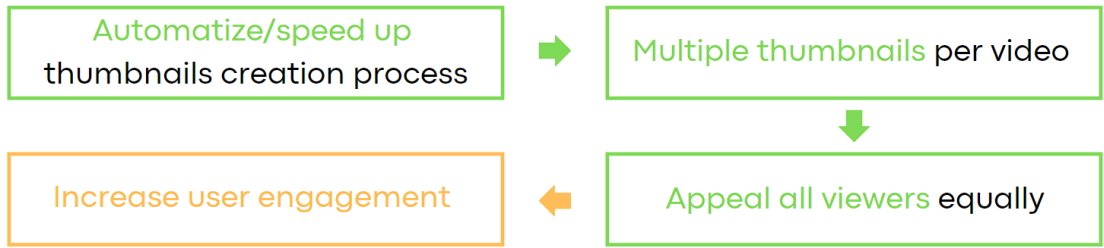

Figure 1.3: Visual representation of the thesis' purpose.

## 1.3 Purpose

The primary purpose of this research is to investigate and implement cutting-edge artificial intelligence methodologies to address an open problem in the domain of digital media: automatizing the thumbnail creation process for video content. By doing so, the research seeks to contribute significantly to both the technological and practical aspects of media presentation and viewer engagement.

The drive behind employing advanced AI techniques in this context lies in the potential to revolutionize how thumbnails are created and selected. Traditional methods, while effective, often lack the scalability and personalization that AI can offer. By automating the thumbnail generation process with state-of-the-art (SOTA) AI models, this research could pave the way for more dynamic and user-specific content presentation. This is particularly relevant for platforms like Play Suisse, where a diverse range of content necessitates equally diverse and appealing visual summaries.

The expected outcomes of this research extend beyond mere technological advancement. At its core, the goal is to attract more users through captivating thumbnails, which are often the first point of interaction between the viewer and the content. Engaging thumbnails are likely to lead to more clicks, increasing viewership and, subsequently, user engagement. This improvement is not only a key objective for Play Suisse but is also aligned with the goals of content creators who strive for broader viewership and engagement with their work. The purpose is visually represented in Figure 1.3.

To thoroughly explore and validate the effectiveness of the proposed AI-driven approach, this research will address several key questions:

- Can a tool leveraging different SOTA AI models automate most of the thumbnail creation process?

- How well does this AI-based approach perform in terms of accuracy and appeal?

- How does its performance compare to traditional manual methods?

- Do users and professional designers find the AI-generated thumbnails preferable or at

least comparable to manually created ones?

- Can this AI-based tool improve the workflow of professional designers, and will it be adopted in their regular practices?

These questions aim to establish not just the feasibility of the AI approach but also its practical implications in real-world settings. By comparing AI-generated thumbnails with traditional methods, both in terms of performance and user preference, the research will provide a comprehensive understanding of the impact of AI on thumbnail generation. Additionally, assessing its integration into professional workflows will offer insights into the adoption and usability of AI tools in creative industries.

## 1.4 Goal

The primary goal of this research is to develop a specialized tool designed to automate a significant portion of the thumbnail creation pipeline used by professional designers. Recognizing the complexity and unique requirements of traditional broadcast video production content, this tool focuses on efficiently executing a subset of the design process. Specifically, the emphasis is on the initial phase of thumbnail creation: the selection, extraction and preliminary editing of candidate images. This approach stems from the understanding that while complete automation of the entire process, including title placement and color grading, is conceivable, it is more practical and beneficial to excel in a specific segment of the workflow and thus provide a robust foundation for designers to build upon, rather than having an end-to-end tool that does not perform well enough.

To this end, the proposed tool will specialize in automating the process of thumbnail selection. It will provide designers with a curated set of images, both extracted and creatively generated, from which they can make their final selections. The extraction component will focus on identifying keyframes from the video content, capturing moments that are representative of the overall material and well-suited to be proposed as a thumbnail, following different criterias. In addition to this, the tool will also feature a creative generation aspect. This involves using automated editing techniques such as segmentation, background replacement, and image harmonization to enhance keyframes, thereby offering a diverse range of potential thumbnails.

A crucial aspect of this tool is its graphical user interface (GUI). The GUI is envisioned to be intuitive and efficient, accelerating the workflow by providing rapid access to a pre-computed database of image options. It will also allow for personalized filtering to meet specific needs when the standard proposals do not suffice. The tool's design will ensure that it integrates seamlessly into the existing workflow of designers, requiring only final adjustments such as title placement and color grading.

An high level scheme of the process applied to an example image is shown in Figure 1.4. The evaluation of this tool is a key part of the research. It aims to assess its usefulness, efficiency, and integration into the professional design workflow. The tool's contribution to the field is twofold: first, it offers a new pipeline that combines state-of-the-art models to address the unmet need for automated thumbnail selection for broadcast video production content on VOD streaming platforms. Second, it provides a novel approach to thumbnail generation from already existing video keyframes, enriching the diversity of available content and offering unseen perspectives.

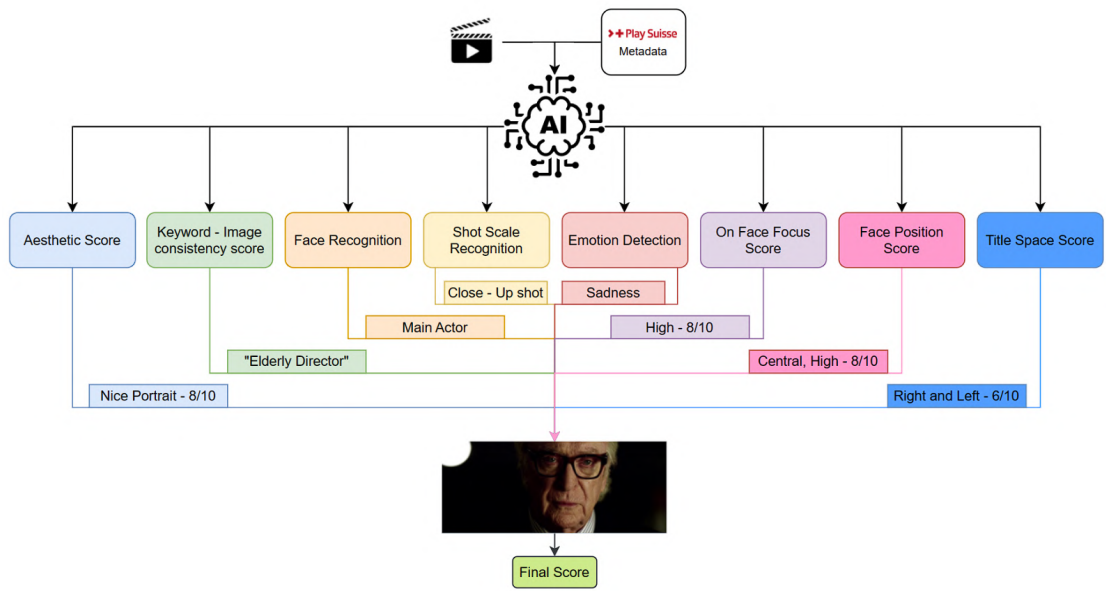

Figure 1.4: High level and visual representation of the proposed process for thumbnail selection. The process is leveraged as a starting point for thumbnail generation as well.

## 1.5 Benefits, Ethics, and Sustainability

The implementation of advanced AI techniques for thumbnail creation offers several benefits for Play Suisse and potentially to the Swiss Radio and Television Company's (SRG) four corporate units. The primary advantage is the acceleration of the thumbnail creation process. This technology, once proven effective, could be adapted to other use cases requiring daily thumbnail generation across SRG's units, thereby streamlining their workflows, reducing time requirements, and consequently, lowering operational costs.

For the broader community, this research introduces a novel framework. It leverages existing state-of-the-art (SOTA) technologies to achieve remarkable results in a specific use case where no prior solution existed. This framework has the potential to be extended to other areas, benefiting smaller content creators such as YouTubers and the creator community at large. In practical terms, this tool allows for the more laborious aspects of thumbnail creation to be managed by AI, freeing creators for more creative endeavors, although it may not necessarily be faster in every instance.

From an ethical standpoint, it's vital to acknowledge that the proposed process does not eliminate the need for human expertise. The AI tool is designed to supplement, not replace, human judgment in the final selection and adjustments of thumbnails. This approach ensures that the human element remains integral in the creative process.

As a public service broadcaster, SRG SSR is entrusted with maintaining the utmost ethical standards. It is imperative that we refrain from utilizing artificial intelligence in any manner that could deceive the public, including the avoidance of clickbait thumbnails and similar tactics.

The ethical concerns in AI, particularly its "black box" nature, are pertinent to this thesis' focus on thumbnail generation using state-of-the-art models. These models, known for their opacity, can make understanding their decision-making processes challenging, potentially leading to uninterpretable outcomes and inherent biases. While these issues are significant, given AI's tendency to reflect biases present in training data, this thesis primarily concentrates on enhancing the thumbnail generation process. It acknowledges these ethical considerations but does not delve deeply into AI interpretability or bias mitigation, as these represent broader challenges in the field of AI beyond this research's scope.

Although sustainability concerns are not the primary focus of the discussion, it is important to highlight that our approach diverges from others in the field by abstaining from additional training of AI models, which would be unnecessary for our specific use case. By leveraging pre-trained methods, we not only mitigate the environmental footprint of our research but also expedite the development process.

## 1.6   Research methodology

In alignment with the precepts outlined in the Methodology Chapter 3, thorough investigation was conducted for each challenge or subtask within the project. This involved an in-depth exploration of the respective fields pertaining to these topics, encompassing an examination of previous methodologies, contemporary state-of-the-art techniques, their respective advantages and limitations, and their applicability to our specific use case. Furthermore, an exhaustive review of analogous methodologies in thumbnail extraction was undertaken to discern similarities in task objectives and methodological approaches. In order to determine the optimal solution, rigorous comparison experiments were systematically conducted, as elucidated in the Implementation Chapter 4.

## 1.7   Stakeholders

This Master's Thesis, conducted under the initiative of Play Suisse, is presented as part of the Master of Science in Data Science program at Ecole Polytechnique Federale de Lausanne (EPFL). The thesis has benefited from the comprehensive guidance of the company supervisor

Dr. Gabriel Autès, providing industry-specific insights and oversight. Academic supervision was provided by Prof. Dr. Sabine Süsstrunk, who leads the Image and Visual Representation Lab at the School of Computer and Communication Sciences (IC). The evaluation of this thesis will be conducted by Prof. Dr. Sabine Süsstrunk, serving as the examiner, and by Dr. Gabriel Autès, serving as the external examiner. Play Suisse's editorial content team and product team are stakeholders responsible for utilizing and maintaining the tools implemented in this research.

## 1.8   Delimitations

In this thesis, the scope is specifically tailored to the extraction of thumbnails from traditional broadcast video production content, where substantial investment is made in the thumbnail creation process. The project applies automation selectively, primarily focusing on the extraction aspect. It does not extend to other elements such as title placement, color grading, or selecting a single best image, as tests revealed that automation in these areas didn't meet the required consistency or quality standards for professional use.

The research explicitly excludes the creation of thumbnails for platforms like YouTube, which have distinct criteria for attractiveness and user engagement. The styles and strategies suitable for YouTube thumbnails vary greatly and are therefore not addressed in this study.

Logo placement, while recognized as an important aspect of thumbnail design, is not a part of the automated process covered in this thesis. On the other hand, it is taken into consideration in the selection process.

The evaluation of the automated thumbnail generation tool was carried out with a limited number of participants. This limitation means that the findings might not fully capture the diverse preferences of a wider user base, a factor that should be considered when interpreting the results.

## 1.9   Outline

- Chapter 2: "Background" provides a comprehensive overview of video summarization and thumbnail selection, discussing the evolution of these concepts from heuristic and domain knowledge-based approaches to modern AI techniques. It concludes with an analysis of related works and how they compare to our proposed method.

- Chapter 3: "Methodology" details the research approach, including description of the available data. It describes the pipeline first with a general overview, then in detail for each of its steps, like downsampling, redundancy reduction, automatic cropping, aesthetic estimation, semantic consistency, shot scale classification, faces and emotions recognition, scoring system, image enhancement and generation, and GUI tool design, emphasizing the chosen methodologies for each task.

- Chapter 4: "Implementation" discusses the practical application of the methodologies described in Chapter 3. It covers the deployment of the system, including technical aspects. It addresses the challenges faced during implementation and the solutions developed to overcome them.

- Chapter 5: "Results and Discussion" evaluates the effectiveness of the implemented system. It includes an evaluation design, the method's selection performance, user preference evaluation, and feedback from professional thumbnail designers. This chapter aims to validate the thesis objectives, demonstrating the proposed method's viability and efficiency in improving thumbnail selection processes.

- Chapter 6: "Conclusion" summarizes the thesis's findings, reflecting on the research questions and discussing the implications of the work. It also outlines future research directions, highlighting potential areas for further investigation and development.

# 2 Background

*This chapter offers a detailed overview of the background upon which this thesis is built, beginning with the definition of video summarization and thumbnail selection in Section 2.1. It delves into the evolution of video summarization techniques, highlighting heuristic, domain knowledge-based, and machine learning approaches in Section 2.1.1. The process and progress in thumbnail selection, from early clustering algorithms to advanced AI models, are examined in Section 2.1.2. Additionally, a review of related work in Section 2.1.3 provides insights into various methodologies, their contributions to our research, and how they compare to our proposed methodology.*

## 2.1 Overview

The realm of multimedia understanding, particularly in the context of video content, presents a multifaceted research problem that has evolved significantly over the years. At the heart of this domain lies the task of thumbnail selection, a seemingly narrow yet intricate challenge that has been explored for many years, even before the advent of advanced AI techniques. This task is not isolated; it is intrinsically linked to the broader objective of condensing and summarizing video content into more digestible forms, be it through a single image, a GIF, a shorter video clip, or a set of images. The literature reveals a diverse array of use cases, each with its unique subproblems and tailored solutions. However, a common thread runs through these varied approaches, suggesting that certain methodologies may have broader applicability across different subdomains.

In the following sections, we comprehensively examine these domains, and we conclude by distinguishing our research from the most pertinent literature.

### 2.1.1 Video summarization

Video summarization has undergone significant transformations in its approach to condensing long-duration videos into compact, information-rich summaries. The progression of these methodologies reflects advancements in understanding both the content of videos and the underlying technologies used to analyze them.

### Heuristic approaches

The early phase of video summarization was characterized by heuristic approaches, which relied predominantly on low-level features to identify keyframes or segments within a video. Smith and Kanade's proposal [104] of selecting keyframes based on professional video editing patterns, such as extracting frames containing both text and faces in scenes, is a seminal example. This approach sought to mimic the intuition of human editors, though it was limited by its reliance on predefined rules and simple visual cues. Ma et al. [72] extended this concept by incorporating attention models based on heuristics, which accounted for a combination of visual and audio saliency, camera motion, and the presence of faces. This marked a shift towards a more holistic understanding of what constitutes informative content in a video.

Laganière et al.'s research [58] introduced an approach focusing on changes in pixel values to identify salient objects or actions. This method represented a move towards more dynamic content analysis, although it still hinged on relatively basic visual information. Arev et al. [5] furthered this evolution by utilizing multiple cameras to create a summarized event video, applying cinematographic guidelines to inform the selection process. These heuristic methods, while foundational, were somewhat limited by their rule-based nature and reliance on low-level features.

### Domain knowledge-based approaches

Domain knowledge-based approaches, emerging alongside heuristic methods, sought to tailor video summarization techniques to specific video types or genres. This customization was based on the premise that different video domains possess unique characteristics that standard summarization methods might overlook. Babaguchi et al.'s work [9] exemplified this approach by using game statistics to summarize American football videos, thereby aligning the summarization process with the specific content and structure of sports videos. Similarly, Sang and Xu [96] leveraged movie scripts for content analysis in film, using the inherent narrative structure of scripts to guide the summarization. Aizawa et al. [2] took an even more personalized approach by suggesting the use of brain wave monitoring to detect scenes of interest in egocentric videos. These domain-specific methods represented a significant advancement in contextual understanding, although their applicability was limited to particular types of video content and additional input data.

### Unsupervised machine learning methods

The advent of machine learning brought a paradigm shift in video summarization, introducing methods that could learn from data rather than relying solely on predefined rules or domain-specific knowledge. This shift began with unsupervised methods, such as the ones developed by Gong and Liu [35] and Uchihashi et al. [112], which focused on clustering visually similar frames to reduce redundancy. DeMenthon et al. [25] approached frame clustering as a curve simplification problem, an innovative perspective that linked video summarization with broader concepts in data simplification and representation. Behrooz et al. [74] and Rochan and Wang [90] introduced encoder-decoder models with adversarial learning, which represented a further advancement in leveraging complex machine learning architectures for frame selection.

**Supervised machine learning methods**

Weakly supervised methods, which utilize coarse annotation data for training, emerged as a compromise between the data-hungry requirements of fully supervised methods and the less precise nature of unsupervised approaches. Potapov et al. [82], for instance, used video topics as a proxy for inferring segment importance, while Xiong et al. [119] hypothesized that video duration could be an indicator of importance. Khosla et al. [50] and Chu and Jaimes [20] exploited web data for summarization, capitalizing on online user behavior and preferences to inform their models. This approach reflected a growing trend in machine learning to leverage large-scale, diverse data sources for training models.

Supervised machine learning methods have significantly advanced the field of video summarization, as highlighted by Gygli et al. [39]. These methods utilize benchmark datasets like SumMe and TVSum to assess frame importance, focusing on reducing estimation errors during training. The incorporation of deep neural networks, along with LSTM layers and attention mechanisms, enables these models to effectively navigate complex temporal relationships within videos. This capability is essential for accurately understanding and summarizing video content.

Rochan et al. [89] introduced a novel framework for personalized video highlight detection. This system is composed of two distinct sub-networks: a fully temporal convolutional network dedicated to highlight prediction, and a history encoder network for analyzing user viewing history. Such an approach allows for tailored highlight detection based on individual user preferences.

Expanding on the use of attention mechanisms, Yen Ting Liu et al. [68] employed self-attention across both temporal and conceptual video features. This method moves beyond mere frame-to-frame correspondence, embracing a broader context encompassing both spatial and temporal diversity for more effective video summarization. By focusing on transformers, their model identifies key regions across these features, facilitating a more nuanced understanding and summarization of video content.

The field then witnessed a resurgence of multimodal methods with Narasimhan et al. [78], who proposed a framework for both general and query-focused video summarization. Utilizing a language-guided multimodal transformer, their approach evaluates video frames for their relevance and connection to either a user-defined query or an auto-generated dense video caption. This allows for the tailoring of video summaries to align with specific user interests. Figure 2.1 serves as a visual representation of the video summarization task and it is taken from the paper of this method.

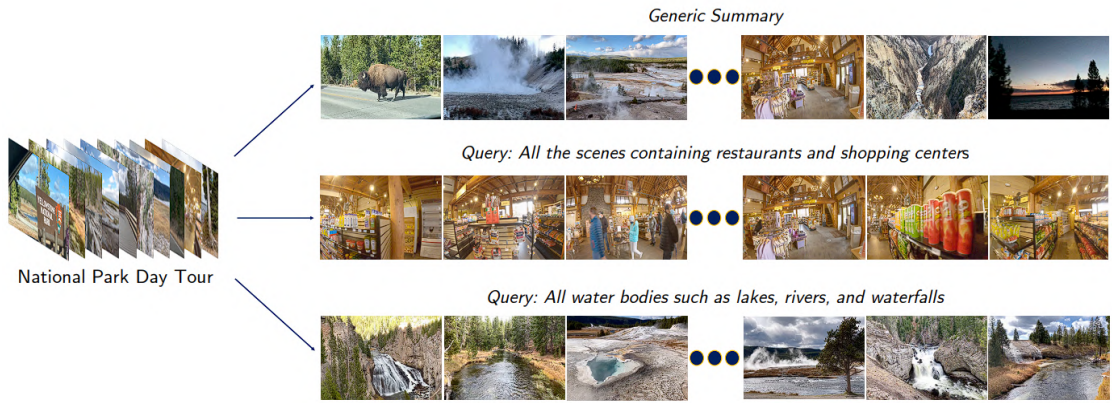

Figure 2.1: Example of a visual representation of the video summarization task taken from CLIP-It! paper [78]. It showcases a video represented by a sequence of frames on the left side, and on the right, multiple possible collections of frames that represent the video's summarization output. This specific method can generate summaries based on user queries given as input.

In a more recent advancement, Haopeng et al. [62] harnessed the power of multimodal capabilities alongside transformer architectures. They integrated BERT for processing textual information and CNNs for analyzing image data. The framework employs specialized loss functions to ensure Cross-modality Semantic Correspondence and to capture Temporal Dependencies in videos. This learning process unfolds in a multi-stage manner, progressively refining the selection of representative frames through each stage.

In summary, the field of video summarization has evolved from relying on simple heuristic rules and domain-specific knowledge to embracing sophisticated machine learning techniques. This evolution reflects a broader trend in technology and data science, where advancements in computational power and data availability enable more nuanced and effective analysis of complex media like videos.

### 2.1.2 Thumbnail selection

Thumbnail selection, akin to video summarization, has been pivotal in enhancing user engagement with video content. Historically, this task was addressed even before the proliferation of artificial intelligence (AI) technologies. Thumbnail selection refers to the process of choosing a frame directly from the video without altering its original content. It distinguishes from thumbnail generation, which involves modifying or creating new images, sometimes partially or wholly different from the original frame.

**Initial approaches**

Initial approaches to thumbnail selection predominantly utilized clustering algorithms, focusing on grouping video frames based on various low-level features. These features included aspects like blur, lightness, colors, scene composition, and the presence of objects, which could be quantified through statistical analysis of the images [128].

Another approach was the PME model, which focused on motion patterns to select keyframes. This method, documented by Sujatha and Mudenagudi [106] employed visual feature descriptors like color histograms, wavelet statistics, and edge direction, enabling the analysis of complex events for keyframe selection.

Unlike previous approaches, Gao et al. [32] emphasized identifying central themes for each video using keywords extracted from audio, tags, or titles. An algorithmic model then detects similarities among images sharing the same theme. This thematic model relies on common features found in relevant visual samples obtained from keyword-based searches in a visual database. The method proceeds comparing candidate thumbnails with visual samples from the database using keyword-based searches.

**Machine learning approaches**

In 2015, Liu et al. [67] introduced a multi-task deep visual-semantic embedding model. This model operates on candidate thumbnails selected offline based on criteria like color entropy, motion blur, and edge sharpness. It computes relevance scores for each candidate thumbnail with respect to a query and selects the one with the highest score. This model marked a shift towards integrating deep AI in thumbnail selection.

Vasudevan's thesis, issued by EPFL, further developed this concept by employing a deep Visual Semantic Embedding Model combined with submodular mixtures [113]. This approach aimed at selecting relevant and visually diverse thumbnails based on text queries, mapping textual and visual information into a common space, training a joint embedding network.

The domain of thumbnail selection was further enriched by the contribution of Song et al. [105]. Their approach, though not based on advanced AI techniques, relied on classical unsupervised machine learning methods. Initially, they filtered out low-quality and transitioning frames using metrics like luminance, sharpness, and uniformity.

Following the filtering process, their process identifies subshots within the video, aiming to discard nearly identical frames and retain those with the highest stillness score. The final step involved clustering the remaining frames using the K-means algorithm and selecting the frame with the highest aesthetic score from each cluster. This methodology, straightforward yet limited by the empirical nature of the metrics used, often served as a benchmark for future studies.

As AI technologies advanced, so did the methods for thumbnail selection. Arthurs et al. in 2017 proposed an intuitive CNN classifier to categorize thumbnails as "good" or "bad" [6]. Their model was based on the assumption that a video's number of views is positively correlated with the quality of its thumbnail. However, this approach did not consider the relevance of the thumbnail to the video content, focusing primarily on visual quality.

In 2018, Gu and Swaminathan developed a more comprehensive system that employed LSTM architecture [37]. This model assigned importance scores to frames for relevance selection, choosing not to rely on semantic information like textual summaries due to the uncertainty of its quality in real-world scenarios.

In 2019, Tsao et al. proposed a five-step pipeline for thumbnail selection [111]. Their method included downsampling, filtering, feature extraction, ranking, and enhancement. They included face detection and similarity elimination in the downsampling process.

In the year 2020, Pretorious and Pillay conducted a review [84] in which they undertook a comparative analysis of various thumbnail classifiers. Their primary focus was on binary classifiers that had been trained on datasets sourced from YouTube videos. These classifiers were specifically designed to assign a likeliness score to images for their potential use as thumbnails. Google [117] applied the same approach in 2015 to automatically suggest suitable thumbnails on YouTube.

In 2020, Yu et al. [123] introduced a method that begins by sampling video frames, assessing their aesthetic quality using a specialized neural network, and then selecting the top frames. Features from these frames, along with video text and audio data, are extracted using specialized models (VGG16, ELECTRA, and TRILL). Features are merged by transformer encoder layers and a context-gating mechanism. The final thumbnail is chosen by matching these multimodal features to a vector in a latent space, ensuring both visual appeal and relevance to the video's content.

The evolution of thumbnail selection methods from early clustering algorithms to sophisticated AI-based models reflects the rapid advancements in video processing and machine learning noticed previously in the video summarization task, and it underscores the increasing complexity and efficacy of these methods.

### 2.1.3 Related work

In the pursuit of developing a comprehensive solution for thumbnail generation and selection, it is imperative to examine previous works, each addressing similar problems from unique angles. While these methods offer innovative solutions tailored to specific use cases, they fall short in directly aligning with our requirements or serving as a baseline for improve-ment. However, elements of these approaches have significantly inspired and informed our methodology.

The primary criteria for our use case encompass a need for multiple thumbnail proposals that are not only diverse and aesthetically pleasing but also focus on human emotions and effectively represent the video content. Additional considerations include the incorporation of title placement and the adaptation to a vertical aspect ratio. Our aim extends beyond simple selection to the modification and generation of thumbnails, thereby amplifying the challenge.

A closer examination of recent works reveals a range of methodologies, each with its unique merits and limitations.

Evlampios Apostolidis et al. [4] introduce a Reinforcement Learning-based Thumbnail Selector, which evaluates the aesthetic quality and importance of each frame using a GoogleNet [107] architecture and a bidirectional LSTM, respectively. Although effective in generating multiple and diverse proposals, the reliance on small, annotated datasets for short-length videos limits its applicability to our context of longer movies. The method's scalability and integration of additional criteria also pose significant challenges, leading us to discard reinforcement learning as a viable solution. The possibility of creating a novel dataset of annotated frames has been taken into consideration, but careful analysis revealed that it would have been unfeasible or at least not preferable for our use case, so similar approaches that required custom training data have been also discarded as possibilities for our methodology.

Jinyu Li et al. [64] method focuses specifically on news videos, altering original frames to include additional imagery and textual elements. While the use of face detection and saliency maps to strategically place these elements, avoiding to draw over important areas of the frame, inspired our pipeline, their outdated keyword extraction method [94] and lack of aesthetic consideration rendered the approach less relevant to our needs.

Andreas Husa et al. [46] proposed a system that incorporates elements like brands' logo and face detection. It also considers aesthetic using a CNN-based method [52].Their methodology is designed for quick, real-time thumbnail proposals, limiting the processing of extracted data to a more simple rule-based decision tree. On the other hand, their usage of an image classifier trained to identify close-up shots has inspired our methodology.

As mentioned in the previous section, Song et al. (2016) [105] has been an influential work that, while dated, provides a comprehensive approach to thumbnail selection. Utilizing a blend of rule-based techniques and classic machine learning methods, it offers multiple, diverse, and aesthetic proposals. However, its reliance on empirical experience and older AI architectures presents limitations in terms of scalability and adaptability to modern criteria. Our method builds upon this foundation, enhancing the approach with the latest AI breakthroughs to better address aesthetic, diversity, and additional criteria like title placement and aspect ratio. In particular, while more recent method use a naive uniform downsampling technique, we apply their strategy of filtering based on empirically set thresholds for luminance, sharpness, uniformity, and stillness.

Zhifeng Yu et al. (2020) [123] developed a multi-modal deep learning model, focusing on aesthetic and representativeness aspects. Their methodology began with filtering frames based on the aesthetic evaluation obtained with a CNN model. Then they extracted features from these frames using another deep CNN. Additionally, it employed the Electra model to analyze textual content and a similar approach for audio features. The final selection process involved using transformers to understand the context within and across these modalities. The model's limitation, however, was its focus on selecting only a single best frame from the shortlisted ones, based on mean squared error (MSE) comparison to the training target, not taking multiple and diverse proposals into account. In addition, this approach is tailored for short videos and would not work on long videos, requiring too big context windows, and it requires training data that we could not provide, as previously mentioned.

Akari Shimono et al. (2020) [102] explores automatic YouTube-thumbnail generation with an emphasis on emotion recognition. While its approach aligns with our interest in capturing visible emotions, the method's simplicity and reliance on basic off-the-shelf components limit its application to our high-quality standards.

In their 2020 study, Baoquan Zhao et al. [127] developed a method for generating informative video thumbnails with a focus on human-centric shots. Their approach combined a portrait attractiveness ranking model and a deep CNN to evaluate frame quality and aesthetics. They selected visually appealing frames, then used clustering for representativeness, based on shot length. Additionally, saliency mapping was employed to arrange multiple images within one thumbnail. Although aspects like saliency and facial aesthetics were pertinent, the overall method of frame merging was deemed too simplistic and unaesthetic for our high-quality standards.

Chun-Ning Tsao et al. (2019) [111] developed a method tailored for selecting thumbnail images for VOD services, particularly focusing on TV series. Their process involved initially discarding the first and last 10% of frames and then sampling at 1fps to perform shot detection, prioritizing longer shots for stability and clarity. They also incorporated criteria such as the presence of characters, image sharpness, and color dynamics in their selection process. While their method and use case shared strong similarities with our approach, it did not extensively address aesthetic appeal or representativeness, making it only partially applicable to our needs. We took strong inspiration from their work for the implementation of our face position score.

Finally, Mahmut Çakar et al. [129] introduced a method for thumbnail selection that involved clustering images and employing various models for face detection, emotion analysis, and object recognition. A notable aspect of their method was the use of the K-means algorithm for dominant color calculation, aiding in avoiding color overlap with film logos. Their reliance on non-state-of-the-art models indicated potential areas for improvement, but the ideas and use cases of their method are close to our work.

| Method | Diverse | Many | Aesthetic | Face | Rep. | Title | Portrait |
|--------|:-------:|:----:|:---------:|:----:|:----:|:-----:|:--------:|
| [4] (2023) | ✓ | ✓ | ✓ | | ✓ | | |
| [64] (2022) | | | | | ✓ | | |
| [46] (2021) | | | ✓ | ✓ | | | |
| [105] (2016) | ✓ | ✓ | ✓ | | | | |
| [123] (2020) | | | ✓ | | ✓ | | |
| [102] (2020) | | ✓ | | ✓ | | ✓ | |
| [127] (2020) | | | ✓ | ✓ | ✓ | | |
| [111] (2019) | ✓ | ✓ | | ✓ | | | |
| [129] (2021) | ✓ | ✓ | | ✓ | | ✓ | |
| Our method | ✓ | ✓ | ✓ | ✓ | ✓ | ✓ | ✓ |

Table 2.1: A comparative overview of the methodologies from the related works section, showing if they take into consideration the key criteria relevant to our thesis: image diversity, proposal multiplicity, aesthetic quality, faces and their emotions, title placement, and vertical portraits.

In summary, while the studies reviewed provide valuable insights and methodologies, none of them fully align with the specific requirements of our project. Notably, none of the existing models address the crucial aspect of generating vertical aspect ratio thumbnails, which is a distinctive aspect of our work. To provide a clearer overview of these models and their coverage of our criteria, Table 2.1 presents a comprehensive summary. Through the integration of cutting-edge AI techniques and customized adaptations, we have effectively addressed the limitations of existing approaches and successfully met all the criteria specified for our project.

# 3 Methodology

*This chapter meticulously delineates the methodology underpinning this thesis, commencing with an exploration of the literature research methodology in Section 3.1, where the foundational approach and rationale behind the pipeline's development are discussed. The data that informs our research is comprehensively outlined in Section 3.2, underscoring its significance and the inherent limitations. A panoramic overview of the proposed pipeline, including its stages and the logic behind its structure, is presented in Section 3.3. The process begins with downsampling techniques in Section 3.4, crucial for efficiency and effectiveness, followed by redundancy reduction strategies in Section 3.5 to ensure diversity among the thumbnails. The automatic cropping method to meet specific aspect ratios is elaborated in Section 3.6, while Section 3.7 explores the methodology behind aesthetic estimation. Ensuring semantic consistency is discussed in Section 3.8, highlighting the importance of relevance to video content. The classification of shot scales is examined in Section 3.9, and the significance of facial and emotional recognition in thumbnails is detailed in Section 3.10. The scoring system to rank potential thumbnails is explained in Section 3.11, while Section 3.12 addresses enhancing the diversity of proposed thumbnails. The chapter further explores image enhancement and generation techniques in Section 3.13 and concludes with the design of a graphical user interface tool in Section 3.14, aimed at assisting professional designers in selecting optimal thumbnails efficiently.*

## 3.1 Literature research methodology

In this section of the thesis, we delve into a comprehensive analysis of our methodology, emphasizing the rationale behind each component. This discussion builds upon the foundation laid in the Introduction, where the primary problem and its specific criteria were identified, and the Background section, which provided a thorough literature review. Our approach involved a meticulous examination of existing literature to discern how others have approached similar challenges. This process was instrumental in distinguishing elements that were beneficial for our context from those that were not suitable for our specific use case.

Guided by a robust understanding of the broader problem domain, we proceeded to formulate an initial proposal for a pipeline designed to meet our established criteria. This pipeline is a culmination of extensive research into each subproblem we encountered. For every segment of the pipeline, a detailed investigation was undertaken to balance various factors: the trade-offs between performance and state-of-the-art (SOTA) solutions, the decision between open-source and proprietary options, and the consideration of custom-developed solutions versus readily available alternatives. These decisions were informed by the relative importance of each aspect of the pipeline, along with the potential for customization of pre-existing methods.

Throughout the development process, the pipeline underwent several iterations. Each modification was guided by feedback from a Play Suisse's professional thumbnail designer, whose insights were crucial in discerning the essential and non-essential features and functionalities. This iterative process of adding and subtracting elements led to the refinement of the pipeline, resulting in the version presented in this section. The collaborative approach not only ensured the practicality and relevance of our methodology but also enriched it with professional expertise, culminating in a solution that is both innovative and grounded in real-world applicability.

## 3.2 Data

In the development of our proposed pipeline, a critical initial step is the evaluation and understanding of the input data available. This research, conducted in collaboration with Play Suisse, leverages their extensive metadata and database information on content.

A limitation in leveraging this data is that most of it becomes accessible only after a video's public release, whereas our pipeline needs to function in the pre-release phase. Furthermore, the quality and extent of the available data vary significantly across different content types. For instance, higher-budget productions typically come equipped with high-quality videos and metadata, including professionally created thumbnails provided by the production companies. In contrast, lower-budget contents often lack such resources. These productions either do not have pre-made thumbnails or have ones of inferior quality, underscoring the importance of our method for these cases, while still being relevant to the creation of additional thumbnails for higher-quality content.

Our method is designed to be versatile, applicable not only to movies but also to TV series. However, the requirement from Play Suisse is to concentrate primarily on single-episode content, such as movies, documentaries, concerts, and so on. While main thumbnails for TV series are usually supplied by the production companies, our method remains relevant for selecting keyframes for individual episodes.

An additional consideration is the length of the videos. While shorter videos, such as trailers, offer more straightforward processing, the reality is that most content from Play Suisse does not include trailers. Therefore, our primary data source is the full-length videos themselves.

A review of existing methods reveals a reliance on complex AI architectures, including transformers and LSTM, to learn intricate temporal dependencies within videos. However, learning these dependencies in longer videos is considerably more challenging. Our research benefits from the finding that temporal dependency is not a critical criterion for our purpose. Unlike video summarization, which requires understanding the sequence and significance of images, selecting a thumbnail can focus on any part of the movie, provided it is contextually relevant.

Besides the video content, accompanying metadata plays a crucial role. This includes tags, categories, production company information, and notably, a brief summary or description of the video content. This summary, even though it can vary in languages, reflecting the language of the production company, is consistently present across all videos, and it is vital for attaining a semantic understanding of the video, crucial for the representativeness of the selected thumbnail. Notably, this kind of descriptive metadata is not commonly available in other use cases, where similar information could be obtained by smart processing of the captions' text.

In summary, the input data for our pipeline comprises the complete video content and its associated metadata. The quality and quantity of this data vary, with the video summary emerging as a particularly crucial element for our thumbnail selection process.

## 3.3 Pipeline Overview

In addressing the multifaceted challenge presented in the introductory chapter, this thesis delineates a structured pipeline, architected to systematically dissect and resolve the various criteria stipulated. The pipeline, a confluence of sophisticated AI models and algorithms, is designed to efficiently process video frames, transforming them into potential thumbnail candidates that adhere to a set of predefined standards. This section provides a panoramic view of the entire pipeline, elucidating its constituent parts and their respective roles in achieving the overarching objective.

- **Downsampling**: The inaugural step in the pipeline is the reduction of frames to be processed. This downsampling is crucial as it circumvents protracted processing times, thus rendering the methodology feasible not only in theoretical constructs but also in real-world applications. By strategically discarding bad and similar frames, this stage ensures that subsequent processes are both efficient and effective.

- **Redundancy Reduction**: Despite the initial downsampling, a degree of redundancy in frames is inevitable. Recognizing that similar frames are likely to yield similar thumbnail suitability, this stage aims to cluster such frames, treating them collectively. This approach not only streamlines the process but also upholds the diversity criterion, ensuring a broad spectrum of thumbnail options.

- **Automatic cropping**: Adhering to the specific horizontal and vertical aspect ratios

required by the Play Suisse platform, this phase automatically adjusts the variable aspect ratios of frames by cropping them. This automated process should maintain the aesthetic and composition of the original frames while conforming to the necessary aspect ratios.

- **Aesthetic Estimation**: At this stage, each frame undergoes an aesthetic evaluation, assigning it a quantifiable score that reflects its visual appeal, with the challenge to turn subjective aspects like beauty into an objective metric.

- **Semantic Consistency**: A thumbnail must encapsulate the essence of the video content. This step involves extracting keywords from content summaries given as input with the rest of the metadata and ensuring that the selected frames resonate with these keywords. Such semantic alignment guarantees that the thumbnails are not only visually captivating but also contextually relevant.

- **Shot scale classification**: Differentiating between various shot scales, this part of the pipeline categorizes images based on their framing. This classification allows for tailored processing, with different scales like long shots and close-ups being used differently to enhance the diversity and applicability of the thumbnails. For instance, long shots are often used as background scenes, and close-up shots are typically suitable for foreground or direct thumbnail use. The ultimate goal is to diversify the range of thumbnail options generated.

- **Faces and Emotions**: In this phase, our focus is on facial features, aligning with the criteria that assign greater significance to individuals and their emotional expressions. The initial step involves face detection, followed by an assessment of whether these detected faces have open or closed eyes. Subsequently, we delve into the identification of emotions conveyed by these faces. Furthermore, we perform face identification and analyze the frequency of appearance of the identified faces, with the presumption that faces appearing more frequently are likely the main actors or characters within the video, making them prime candidates for inclusion in the thumbnail.

- **Scoring System**: In order to rank the potential thumbnail candidates effectively, it is essential to consolidate the information obtained from the preceding steps into a single numerical score. This score is derived from the conversion of various extracted data into weighted scores, encompassing factors such as aesthetic appeal, semantic relevance to keywords, the faces' positions and their importance in the whole image (on face focus), and the available space for logo placement.

- **Variety enhancement**: The scoring system enables the numerical assessment of images for thumbnail selection. However, the current process lacks diversity, often resulting in a pool dominated by similar options. In this step, we first select the highest-scoring image from similar ones. Yet, achieving true diversity remains challenging, as images, while meeting various criteria, may lack specific relevance to individual keywords. This is accomplished with another step through the strategic adjustment of scoring criteria to align with distinct objectives.

- **Image enhancement and generation**: This step addresses the generation of thumbnails not originally present in the video, emphasizing the need for content fidelity. The approach involves extracting optimal portraits of main characters from the video and integrating them with appropriate backgrounds, either from the video itself or generated through text-to-image models. For the first option, we initially identify optimal foreground and background candidates, then the foreground elements are isolated from the initial images through a combination of segmentation and matting methodologies. Subsequently, the background is substituted, and a harmonization technique is employed to enhance the coherence and visual integration of the composite image. For text-to-image model-generated backgrounds, we employ methodologies aimed at exercising control over the generation process, thereby ensuring the resultant content remains contextually pertinent and faithfully representative of the video.

The final goal of the project is to create a graphical interface that allows professional designers to efficiently choose optimal thumbnails from a video and make final adjustments such as color grading and logo placement. The interface aims to be intuitive, practical, and user-friendly, enhancing the user's efficiency in finding necessary resources. It's designed to streamline the workflow and empower designers to make swift decisions. This part is detailed in the final section of methodology.

To provide a comprehensive understanding of our approach, Figure 3.1 presents a high-level overview of the pipeline, closely mirroring the sequence of steps in the implemented system. We will now proceed to conduct an in-depth analysis of each component within the pipeline.

## 3.4 Downsampling

In the domain of digital video processing, particularly in the context of thumbnail generation for movies, the concept of downsampling plays a crucial role. Downsampling, fundamentally, refers to the reduction of samples or data points in a digital signal or dataset. In the context of video processing, it pertains to decreasing the number of frames in a video for efficient processing and analysis. This reduction is essential given the high frame rate at which videos are typically recorded and displayed.

Standard practice in the film and television industry is to record and display videos at 24 frames per second (fps), which has been established as the minimum frame rate to preserve the illusion of continuous motion and realism. This standardization, however, presents a challenge when selecting a single frame to serve as a video's thumbnail, especially in the case of a two-hour movie, which would result in a staggering 172,800 frames to choose from. Processing this volume of data is not only resource-intensive but also largely redundant, as successive frames in a video often exhibit minimal changes except during transitions between shots, where a shot stands as a term for the range of frames from the moment that the camera starts rolling until the moment it stops.

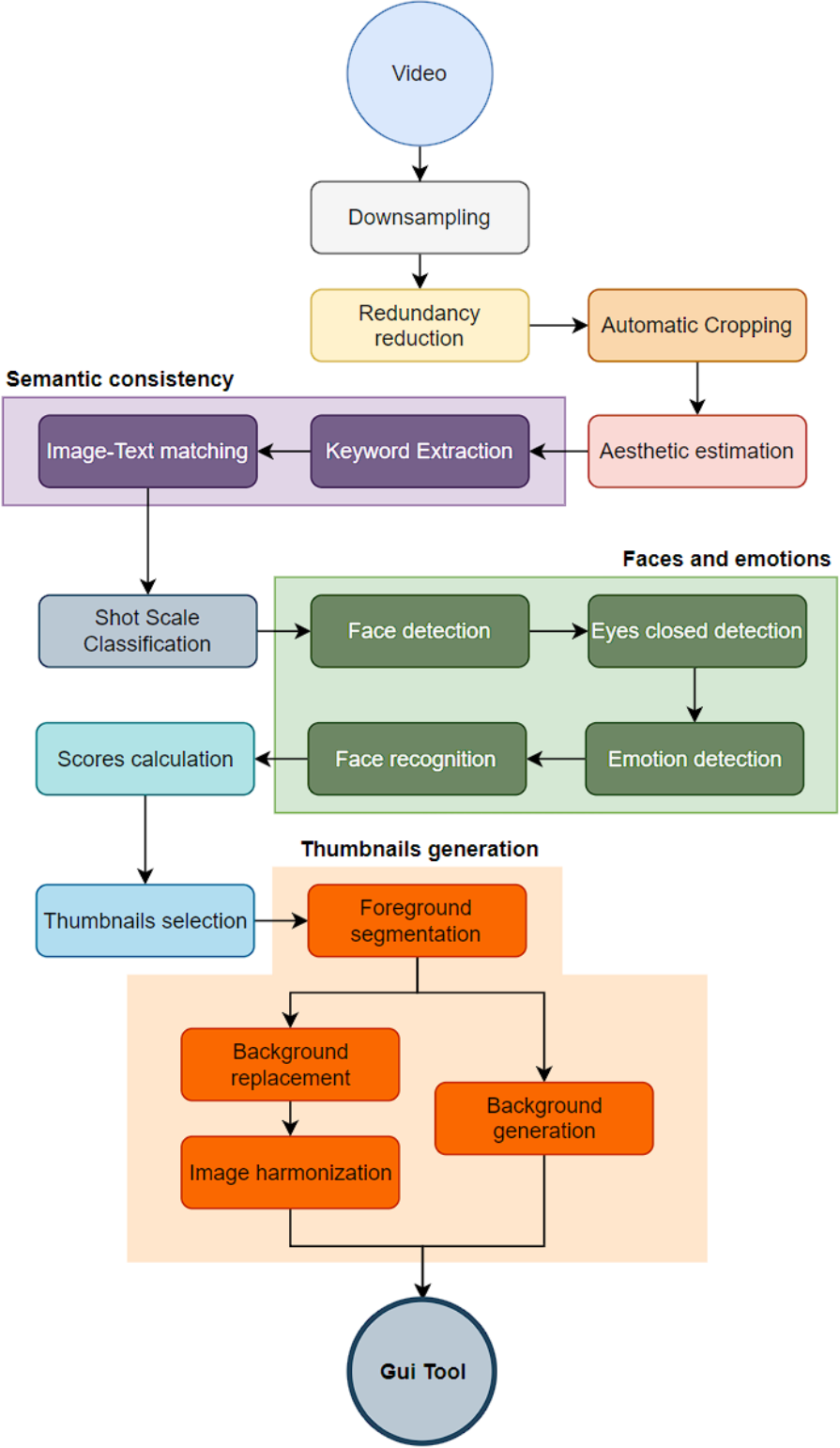

Figure 3.1: Schematic representation of the whole pipeline process

### 3.4.1 Literature research

In response to this challenge, various methods have been developed and applied in the field, each with its unique approach to downsampling. A common initial step in many of these methods is the uniform subsampling of frames, typically at a rate of one frame per second. This approach significantly reduces the volume of frames to be processed. Tsao et al. [111], for example, employed a multi-step process beginning with the discarding of the initial and final 10% of the video, followed by subsampling at one frame per second, identification of shots, and retention of the 100 longest shots. This method operates under the assumption that longer shots tend to be more stable and sharper, while also being more representative. However, we contend that among the 24 frames captured each second, the best frame may not necessarily follow a regular interval pattern. Additionally, limiting the selection to just 100 shots may not be sufficient, especially for movies with thousands of them, with shorter shots that could be relevant as well. Finally, the decision to exclude ending frames with potential spoilers should be left to professional designers during the selection process.

Alternatively, Mahmut Çakar et al. [129] proposed a technique that retains only frames containing a detected face, utilizing the Haar Cascade algorithm for face detection. This method, while effective in certain contexts, may not be ideal for genres like documentaries where significant frames may not contain faces. Additionally, the Haar Cascade, being somewhat outdated, may miss several faces, leading to the exclusion of potentially relevant frames.

Another approach involves the use of Convolutional Neural Networks (CNNs) to classify frames, as suggested by Pretorious et al. [84]. In this method, frames with the highest classification scores are selected as thumbnails. This method does not inherently involve downsampling, but it could be adapted for such. The downside of this approach is its lack of interpretability and control over the selection process, as well as the computational expense of applying CNNs to every frame. Moreover, CNNs do not incorporate temporal information, which can lead to the omission of entire sections of the video. Recognizing these limitations, Evlampios Apostolidis et al. [4] added a bidirectional LSTM layer to include the temporal dimension in the thumbnail selection process, aknowleding the importance of a frame's temporal context in determining its representativeness.

### 3.4.2 Chosen methodology: Hecate

After thorough consideration of these and other methods, the most effective downsampling technique appears to be that proposed by Song et al. [105], called Hecate. They argue against uniform subsampling as it could discard ideal frames, advocating for a more nuanced approach. Their method begins with filtering out low-quality frames based on criteria such as

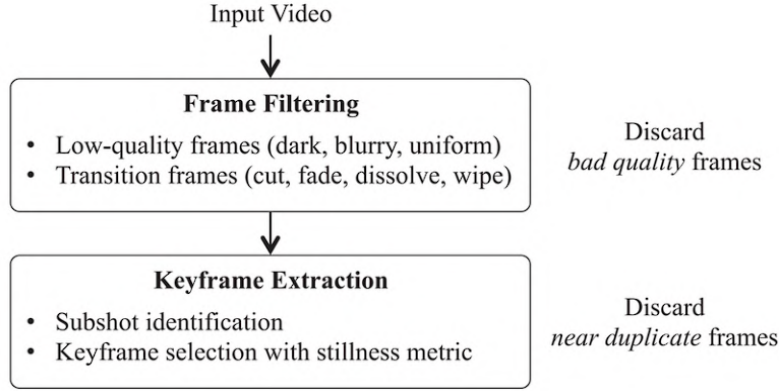

Figure 3.2: Schematic representation of the downsampling process of Hecate, taken from the original paper [105]

darkness, blurriness, and uniformity. Such criteria are evaluated by the following equations:

$$Luminance(I_{rgb}) = 0.2126I_r + 0.7152I_g + 0.0722I_b \qquad (3.1)$$

$$Sharpness(I_{gray}) = \sqrt{(\Delta_x I_{gray})^2 + (\Delta_y I_{gray})^2} \qquad (3.2)$$

$$Uniformity(I_{gray}) = \int_0^{5\%} cdf(sort(hist(I_{gray}))) dp \qquad (3.3)$$

In these formulas, $I_{rgb}$ represents an RGB image, where $I_r$, $I_g$, and $I_b$ are the red, green, and blue channels of the image, respectively. The symbol $I_{gray}$ represents a grayscale image. The terms $\Delta_x$ and $\Delta_y$ represent the gradient of the image in the horizontal and vertical directions, respectively. The function $cdf$ denotes the cumulative distribution function, and $hist$ represents the histogram of the image intensities. The $sort$ function is used to sort the values in descending order. The integral from 0 to 5% represents the sum of the lowest 5% of the cumulative distribution function of the sorted histogram values.

Following this, they address the challenge of transition frames in a less extreme manner than simply selecting the longest shots. By employing the shot detection technique of Zabih et al. [124], they discard frames occurring in transitions between shots. Subsequently, they do not limit the selection to one frame per shot but instead identify subshots. This is achieved by clustering the remaining frames using the k-means algorithm. For each frame in a subshot, its stillness value is computed as an inverse of the sum-squared pixel-wise frame difference value between two time-consecutive frames. Keyframes are then extracted from each subshot based on the stillness metric, selecting the most stable frame from each. An overview of the whole method can be find in Figure 3.2.

While the method from Song et Al continues to further process the remaining frames and select a set of thumbnails, we leveraged only this first part as our downsampling technique.

By rapidly filtering out frames based on objective quality criteria and motion dynamics, this method avoids the pitfalls of random uniform sampling and ensures that the retained frames are both representative and of high quality. Thus, it offers a balanced and fast solution.

## 3.5 Redundancy reduction

In the preceding discussion, we elaborated on our downsampling methodology, which segments each shot into subshots, retaining a single frame from each. This approach is crucial for capturing the dynamic range of content within shots. As an example, in scenarios involving motion, such as a car speeding on a highway in an action film, the camera's perspective may transition from the front of the car to its rear as it overtakes the camera, showcasing dramatically different visuals at the beginning and end of the shot. This variability underscores the inadequacy of methods that select a single frame per shot, as they risk omitting significant content variations within shots.

However, this method introduces a challenge in handling shots where the initial, final, and intermediary frames exhibit minimal variance. Despite the efforts to mitigate this redundancy through k-means clustering in the downsampling process, a substantial number of similar images persist. For our automated thumbnail selection system this could be a problem, since we analyze and score images based on extracted features, aggregating these scores into a composite measure. Given the minor differences among closely related images—sometimes as negligible as slight pixel shifts or color alterations—the resulting scores for these images are nearly identical. Consequently, this can lead to the recommendation of nearly duplicate images as top thumbnail candidates, contravening our diversity criterion.

To circumvent this issue, while maintaining the strategy of selecting one frame per subshot—thus allowing for multiple frames per shot in the downsampling phase—we ensure that subsequent stages of our pipeline are equipped to handle this diversity effectively. Ultimately, when presenting thumbnail candidates to the user, we group similar frames, showcasing only the highest-scoring frame within each group. This allows us to maintain a balance between diversity and relevance in our recommendations. Should the user desire, they have the option to view all other images within a group, facilitating a more tailored selection process.

To address the aforementioned challenge, our approach primarily utilizes unsupervised clustering techniques, given their capacity to discern intrinsic groupings within the data without predetermined labels. A pivotal obstacle in this context is the indeterminacy regarding the number of clusters, which precludes the application of algorithms necessitating a predefined cluster count.

### 3.5.1 Unsupervised clustering

In navigating this challenge, we opted for the Density-Based Spatial Clustering of Applications with Noise (DBSCAN), as delineated in Ester et al. [29]. This selection was informed by its congruence with our objectives, particularly its applicability to scenarios akin to our encounter with the face identification problem, wherein it was imperative to cluster similar facial embeddings. An exhaustive exploration and comparative analysis of alternatives to DBSCAN, especially in the context of the face identification problem, will be deferred to a subsequent section dedicated to that topic.

DBSCAN groups data points based on their density distribution in the feature space, identifying clusters as regions where data points are closely packed together, separated by areas with lower point density. It is effective at finding clusters of arbitrary shapes and can distinguish noise points that do not belong to any cluster.

### 3.5.2 Feature space selection

In selecting a feature space for clustering, traditional methods often rely on histogram-based features. For instance, Song et al. [105] utilized a combination of HSV histogram pyramids and edge orientation and magnitude histograms, culminating in a feature vector of 2,220 dimensions for subshot identification. Conversely, more contemporary strategies incorporate deep learning techniques, extracting image embeddings via pre-trained neural networks. A prevalent example includes embeddings derived from a VGG16 network [103], which was initially designed for image classification tasks. Our methodology, however, opts for leveraging embeddings from the CLIP model [86], which will be described in more detail in a later section.

CLIP embeddings generate feature vectors with a dimensionality of 512, posing a significant challenge for clustering due to the inherent sparsity in such high-dimensional spaces. This sparsity can lead to an overestimation of the distinctiveness of each data point, potentially misclassifying them as noise. To counteract this, we employ Principal Component Analysis (PCA), a statistical technique aimed at dimensionality reduction while retaining the most important information. PCA accomplishes this by identifying principal components, which are linear combinations of the original variables that capture the greatest variance within the dataset. An empirical threshold for the explained variance—a metric quantifying the proportion of total variance a component accounts for—was established to determine the optimal reduced dimensionality.

### 3.5.3 Refinements

Nevertheless, clustering alone does not suffice to address all challenges. Specifically, this approach may still result in diverse clusters for highly similar images or, conversely, amalgamate similar images based solely on their visual content, disregarding their contextual separation within the narrative, being from very different parts of the video. To refine this, we incorporate insights from previously conducted shot detection, integrating it with CLIP-based clustering to enhance the coherence of our groupings. Specifically, we merge consecutive shots into a single group if CLIP clustering puts them in the same group, while we divide CLIP clusters that amalgamate frames from narratively distant shots.

This hybrid strategy, which synergizes shot detection with CLIP clustering, facilitates a more nuanced and contextually aware grouping of frames. By doing so, we ensure that our clustering not only reflects visual similarity but also respects the narrative structure and temporal distribution of shots.

## 3.6   Automatic cropping

In the literature review, we noted that existing methodologies for thumbnail generation did not account for the potential alteration of aspect ratios, typically selecting frames directly from videos without modification. This approach overlooks a critical practical consideration: thumbnails must conform to specific aspect ratios dictated by the platforms on which they are displayed, such as YouTube, Netflix, or Play Suisse. These platforms have strict requirements for thumbnail dimensions, necessitating adaptations to ensure compatibility. Films and documentaries, which may employ a variety of aspect ratios for artistic expression, are often modified through letterboxing (the addition of black bars above and below the image) to meet these platform-specific standards. However, such letterboxing is undesirable for thumbnails, as it detracts from the aesthetic appeal. Consequently, it is essential to eliminate these black bars by automatically removing rows of pixels that contain only black, thereby preserving the integrity of the image.

After the removal of letterboxing, it becomes necessary to adjust the aspect ratio of the resulting image to fit platform specifications, a process that involves cropping the image. This adjustment aims to retain as much of the image's crucial content as possible while eliminating extraneous elements. This method is particularly relevant for Play Suisse, which traditionally employs a standard horizontal aspect ratio of 16:9. However, the rising popularity of vertical content on social media platforms such as Instagram and TikTok, where videos are predominantly vertical, has led to an increased demand for vertical thumbnails. Consequently, platforms like Netflix have adapted by displaying vertical portraits on certain devices, with an always increasing focus on characters and their emotions. Recognizing this trend, Play Suisse is also embracing vertical aspect ratios, specifically a 2:3 ratio, underscoring the importance of our methodology's capability to generate vertical thumbnails from inherently horizontal content through strategic cropping. This approach distinguishes our method from others by its consideration of aspect ratio adjustments and cropping techniques, ensuring the generation of aesthetically pleasing and platform-compliant thumbnails.

Addressing the challenge of maintaining the aesthetic and compositional integrity of thumbnails during aspect ratio adjustment necessitates a nuanced approach beyond the simplistic strategy of uniformly trimming pixels from the edges. This method, while straightforward, risks omitting critical elements of the image that contribute to its selection as a thumbnail candidate. The complexity of this task is further compounded by the vast array of possible cropping dimensions and positions, rendering the identification of an optimal solution not so trivial.

### 3.6.1 Literature research

The domain of autocropping has evolved significantly, marked by a series of innovative methodologies aimed at refining the process of image cropping to preserve aesthetic value. We will rapidly go through the latest works. In 2017, Debang Li et al. [60] introduced a supervised learning approach that conceptualizes image cropping as a sequence of decision-making steps, optimized through reinforcement learning. This method, which emphasizes aesthetic considerations, diminishes the reliance on generating numerous candidate crops by employing an aesthetics-aware reward function within an actor-critic framework.

Further advancements were made in 2019 by Peng Lu et al. [71], who leveraged aesthetic insights derived from high-quality photographs to inform the cropping process. Their methodology utilizes a CNN enhanced by a Gaussian kernel to pinpoint an anchor region, ensuring the focal points of interest are preserved. The integration of a regression network further refines the cropping process, establishing a direct correlation between the identified anchor region and the ultimate crop, thereby streamlining the procedure.

Continuing this trajectory, in 2020 Debang Li et al. [61] proposed a novel framework capable of accommodating various aspect ratios through the introduction of meta-learners. These meta-learners adjust the cropping model's parameters to suit specific aspect ratio requirements, facilitating the generation of tailored cropping models without necessitating multiple iterations of the model for different ratios. This approach significantly enhances both the precision and efficiency of the cropping process.

In a more recent exploration, Chaoyi Hong et al. [44] ventured into the realm of explicitly encoding photographic composition rules within the cropping algorithm. By devising a Key Composition Map (KCM), this methodology integrates foundational principles of composition directly into the model, guiding the cropping mechanism towards regions of importance in alignment with established compositional guidelines.

Despite these advances, limitations persist, particularly in the context of our application. Some methods lack the flexibility to adjust the output's aspect ratio according to specific requirements, automatically determining the optimal ratio for the image at hand. Alternatively, models that do permit aspect ratio customization often require retraining for each new ratio, a process that is neither practical nor scalable, especially considering potential future changes in thumbnail aspect ratio preferences by the Play Suisse platform.

### 3.6.2 Chosen methodology: GAIC

In addressing the constraints identified with existing methods, our approach sought a more adaptable solution. This led to the adoption of the method proposed by Zeng et al. in 2019 [125], called Grid Anchor based Image Cropping (GAIC). This technique significantly streamlines the cropping process by overlaying a grid on the image, utilizing intersections as anchor points for candidate crops. This approach not only narrows the search space for potential crops but also ensures that the selected crops adhere to aesthetic and compositional standards, including the maintenance of desired aspect ratios and the preservation of important content.

The method employs a Convolutional Neural Network (CNN) architecture designed for efficient cropping, featuring multi-scale feature extraction to analyze images at varying scales. This allows for the meticulous preservation of essential content while simultaneously identifying and eliminating superfluous elements. The model's lightweight and rapid processing capabilities further enhance its suitability for practical application. Essentially, the anchor-based grid system suggests a variety of possible cropping boxes, and the CNN model assigns scores to these boxes, with the selection of the highest-scoring box as the final crop.

The versatility of the grid anchor-based cropping method is highlighted by its ability to accommodate any chosen aspect ratio without necessitating retraining for each new specification. This adaptability is achieved by ensuring that the crop boxes generated by the grid conform to the desired aspect ratio.

### 3.6.3 Refinements

Moreover, the method can be easily integrated with an additional step that filters the selected crop boxes before they are evaluated by the CNN model. This filtering will leverage the face detection step that is required also by subsequent phases of the pipeline. More specifically, crops that inadvertently bisect faces or disproportionately highlight smaller faces over larger ones are filtered out. Additionally, for vertical portraits, which demand a high degree of compositional balance, any crop box that features a solitary face not aligned near the center of the vertical axis is discarded. This ensures that the resulting thumbnails are not only aesthetically pleasing but also compositionally sound.

To illustrate the comprehensive nature of this methodology, a schematic representation is provided in Figure 3.3.

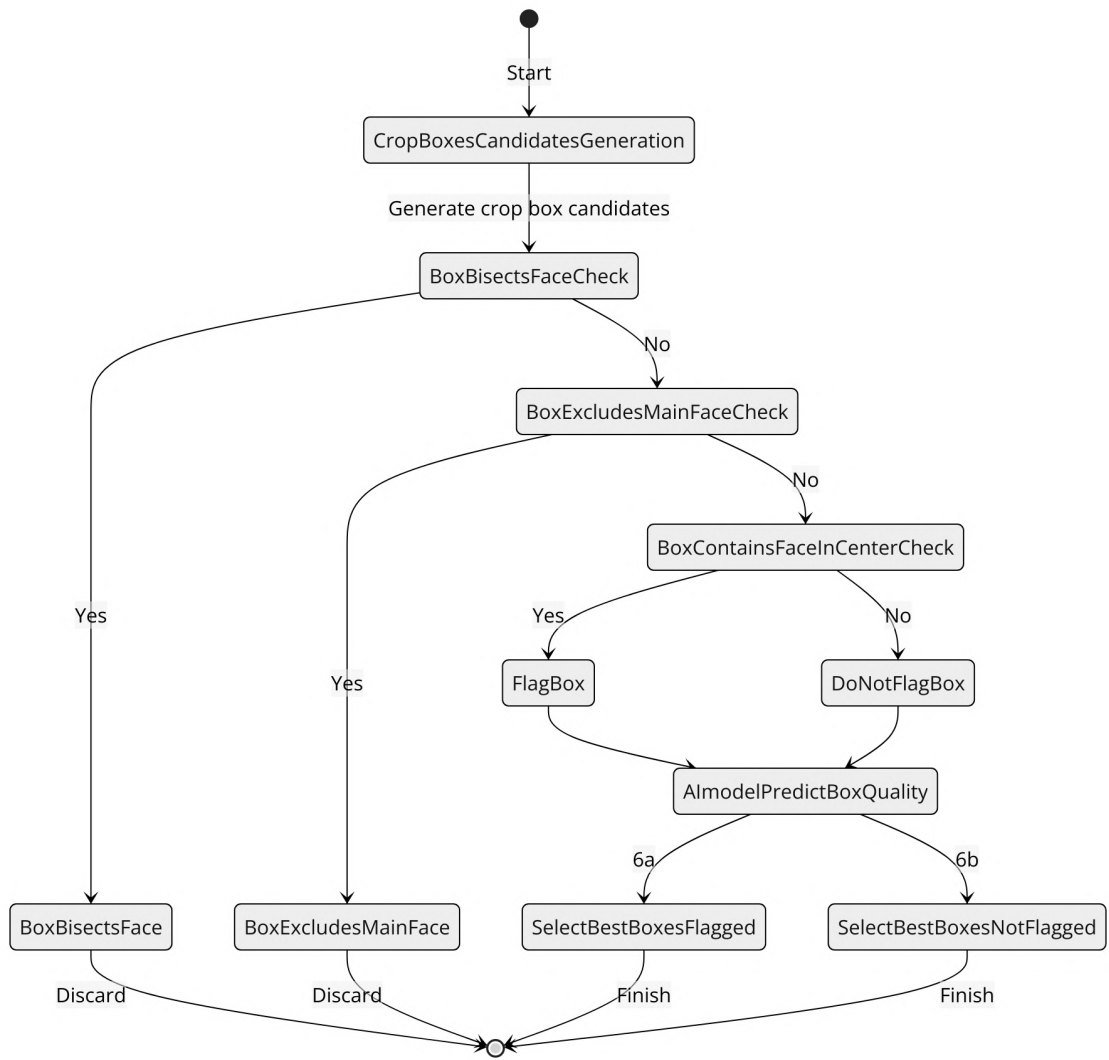

Figure 3.3: Schematic representation of the cropping process

## 3.7  Aesthetic estimation

In the introductory section, we emphasized the pivotal role of aesthetic appeal in the selection criteria for our proposed thumbnails. This emphasis stems from the understanding that an image lacking in visual appeal can inadvertently signal poor video quality, thereby deterring potential viewers from engaging with the content. To address this, we propose assigning an aesthetic score to each image, allowing us to prioritize those images that meet a high standard of visual appeal.

The endeavor to automate the aesthetic evaluation of images presents a complex challenge, as it necessitates the development of algorithms capable of mirroring human judgment in assessing aesthetic quality. This complexity is rooted in the subjective nature of aesthetic appreciation, which exhibits significant variation across individuals. Aesthetic judgment is influenced by a multitude of factors, including but not limited to composition, color harmony, and emotional resonance, all of which are challenging to quantify objectively. Consequently, automatic aesthetic estimation represents a sophisticated intersection of challenges spanning image analysis, machine learning, and the understanding of human aesthetics.

Given the scarcity of universally accepted benchmarks, pinpointing a method that can be definitively labeled as the current state of the art in automatic aesthetic estimation is challenging. In light of this, we embarked on a comprehensive literature review concerning the topic of aesthetic estimation. Through this review, we identified the method most suited to our specific use case. Our selection was informed by a combination of the method's alignment with our requirements and our subjective evaluation based on experimental outcomes, which will be shown in the Implementation Section 4.5.

### 3.7.1 Literature research

Noteworthy is the work by Hossein Talebi and Peyman Milanfar [108], who introduced Nima, a convolutional neural network (CNN) based approach that assesses image quality and aesthetics by predicting the distribution of human ratings, rather than just the mean score, employing a squared earth mover's distance (EMD) loss to refine the assessment process. This methodology not only enhances automatic image evaluation but also aids in improving image enhancement techniques.

Parallel to this, Shuai He et al. [42] focused on color aesthetics through the Delegate Transformer, which employs a unique attention mechanism to allocate interest points for dominant colors and learns to segment color spaces by simulating human behavior. On the other hand, their narrowed focus on colors might overlook other aesthetic dimensions such as composition. In a complementary vein, the same group's earlier work [43] tackled the influence of image themes on visual perception with TANet, a model that adapts evaluation criteria based on thematic content, addressing the issue of uniform criteria applied across varied themes in existing image aesthetic assessment (IAA) methodologies.

Junjie Ke [49] ventured into leveraging transformer models for IQA with the multi-scale image quality Transformer (MUSIQ), which processes images at their native resolution to circumvent the limitations of CNNs. It incorporates an hash-based 2D spatial embedding and a scale embedding to support positional embedding within this multi-scale framework. This approach, while innovative, provides limited advantage from these features in contexts where image resolution is uniform, such as content coming from the same video as in our use case.

Further, Chaofeng Chen et al. [16] introduced TOPIQ, employing a heuristic coarse-to-fine network (CFANet) inspired by the human visual process. This model, utilizing a cross-scale attention mechanism and ResNet50 as its backbone, uses high-level semantics to guide the IQA network towards focusing on semantically important local distortion regions.

### 3.7.2   Chosen methodology: CLIP-IQA

The preferred approach in recent literature is the CLIP-IQA method proposed by Jianyi Wang et al. [116], which harnesses the visual-language knowledge embedded in CLIP models [86]. This method leverages CLIP for zero-shot image quality and abstract perception assessments by comparing images with evaluative prompts like "good picture" and "bad picture," using cosine similarity and softmax to gauge semantic similarity.

Cosine similarity is a measure used to determine how similar two vectors are in a multi-dimensional space. It is defined as the cosine of the angle between the two vectors. For two vectors $\mathbf{A}$ and $\mathbf{B}$ in an $n$-dimensional space, the cosine similarity cosine_similarity($\mathbf{A}, \mathbf{B}$) is calculated as:

$$\text{cosine\_similarity}(\mathbf{A}, \mathbf{B}) = \frac{\sum_{i=1}^{n} A_i \times B_i}{\sqrt{\sum_{i=1}^{n} A_i^2} \times \sqrt{\sum_{i=1}^{n} B_i^2}} \tag{3.4}$$

where $A_i$ and $B_i$ represent the $i$th components of vectors $\mathbf{A}$ and $\mathbf{B}$, respectively. The dot product of vectors $\mathbf{A}$ and $\mathbf{B}$, $\sum_{i=1}^{n} A_i \times B_i$, represents the similarity between the vectors in the direction of each component. $\sqrt{\sum_{i=1}^{n} A_i^2}$ and $\sqrt{\sum_{i=1}^{n} B_i^2}$ denote the magnitudes of vectors $\mathbf{A}$ and $\mathbf{B}$, respectively, calculated using the Euclidean norm.

This approach mirrors our use of CLIP for semantic similarity, as explained in the next section, with their effective prompt pairing strategy capitalizing on CLIP's pre-existing priors for broad perceptual evaluations. An extension, CLIP-IQA+, further refines this by fine-tuning prompts through backpropagation while keeping the network weights static. This fine-tuning of prompts, starting from ["Good photo.", "Bad photo."], enhances the model's ability to generalize across different assessments. A visual representation of the method is presented in Figure 3.4.

### 3.7.3   Difference between IQA and IAA

It is pertinent to acknowledge that certain methodologies [49, 16, 116] address the domain of image quality assessment (IQA). Notably, IQA represents a broader scope than image aesthetic assessment (IAA); however, these methodologies demonstrate remarkable efficacy when employed for aesthetic evaluation. Within the realm of IQA, there are two primary tasks: the full reference task (FR), which primarily involves assessing discrepancies between distorted images and their corresponding reference images, and the no reference task (NR), which necessitates the evaluation of overall aesthetic quality without access to reference images. Consequently, IQA methods that are adept at addressing NR tasks possess the inherent potential to also excel in the context of IAA.

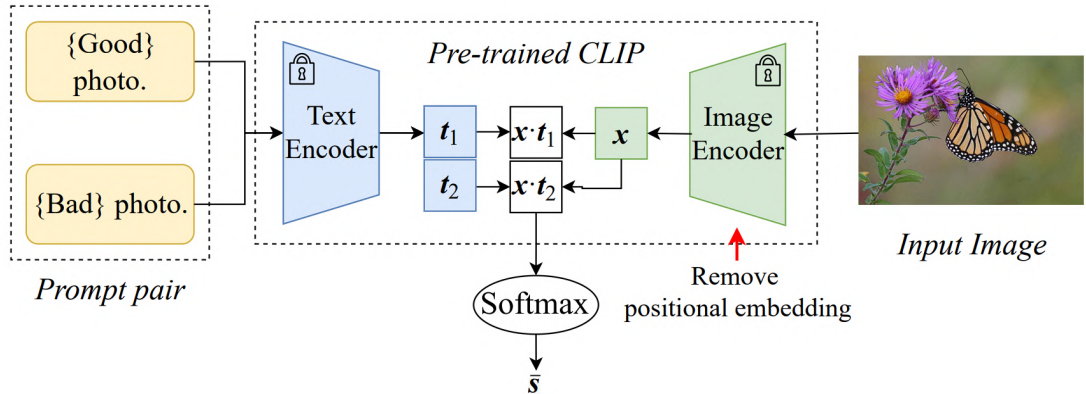

Figure 3.4: Visual representation of the approach used by CLIP-IQA [116], taken from the original paper. The image embedding is compared to the embeddings of the prompt pair using cosine similarity, and a softmax function is then applied to the resulting pair of cosine similarities to obtain a final score.

## 3.8 Semantic consistency

In previous discussions, we emphasized the critical role of ensuring that a video's thumbnail semantically aligns with its content, a challenge particularly significant for thumbnail designers at Play Suisse. Consider the example of a documentary on eagles, enriched with captivating scenery and extensive shots of mountains. In such cases, it is imperative to feature an eagle prominently in the foreground of the thumbnail. However, not just any depiction of an eagle suffices. A more fitting choice would be an image that not only highlights the eagle but also incorporates a scenic backdrop, possibly including mountains, to provide a more comprehensive representation of the video's content. The complexity of automating this nuanced process of semantic consistency is notable.

Contrastingly, existing methodologies for thumbnail selection primarily address the issue of representativeness from divergent angles, often sidelining the semantic perspective. This is partly due to the availability of content descriptions and other metadata, which are not always present in various scenarios [37]. For instance, the approach by Zhifeng Yu et al. (2020) [123] attempts to deduce semantic content from multiple modalities like text and sound. We posit that a similar semantic processing could be achieved by synthesizing a summary from intelligently processed captions, which could be auto-generated when not available. However, this avenue remains unexplored in our context, where descriptions are consistently provided by Play Suisse metadata. Conversely, several strategies [127, 105] do not primarily focus on semantic representativeness but instead infer the significance of a frame based on its context within the shot sequence. They employ clustering to gauge representativeness, hypothesizing that longer shots correlate more closely with content relevance. Yet, this assumption may not always hold. Revisiting the eagle documentary example, even if the majority of the footage comprises close-ups of experts discussing eagles, with only a brief portion showcasing the eagles themselves, a thumbnail featuring an eagle would arguably convey a more accurate glimpse into the video's essence than one focusing on the leading expert. It is this rationale that guides our method's reliance on semantic considerations for thumbnail selection.

In the exploration of leveraging semantics for the purpose at hand, our work builds upon previous studies in the domain. Prior research, notably by Liu et al. [67], introduced a multi-task deep visual-semantic embedding model based on CNN architecture. Similarly, Vasudevan's work [113] integrated the CNN Semantic Model with submodular mixtures to compute relevance scores for candidate thumbnails in relation to a given query, selecting the highest-scoring thumbnails. Our methodology, while bearing resemblance to these approaches, diverges by leveraging transformers architecture to model the multimodality of semantics, leveraging the power of SOTA pre-trained models. This shift allows us to bypass the necessity for extensive training data and the associated costs of training processes.

### 3.8.1 Keywords extraction

**Literature research**

The above-mentioned methods relied on user-provided queries to extract keywords. In contrast, our approach introduces automation in this process, enabling the inference of keywords from potentially lengthy descriptions. This task, a classic challenge within the realm of Natural Language Processing (NLP), has historically employed various techniques. Among these, Term Frequency-Inverse Document Frequency (TF-IDF) assesses the significance of a term within a document relative to a corpus. On the other hand, Latent Dirichlet Allocation (LDA) [11], a generative statistical model, posits that each document is an amalgamation of a limited number of topics, with the presence of each word attributed to a specific topic within the document. Similarly, Latent Semantic Analysis (LSA) [24] employs singular value decomposition (SVD) on the term-document matrix, uncovering patterns in the interconnections between terms and documents across the corpus. Rapid Automatic Keyword Extraction (RAKE) [94] represents a more contemporary approach, identifying key phrases by examining the frequency of word appearances and their co-occurrences with other words. The advent of the transformer architecture marked a significant leap forward in performance for these tasks. Pre-trained language models, such as BERT [27], have revolutionized keyword extraction by embedding, identifying, and ranking terms based on their contextual relevance within the document.

**Chosen methodology: Large Language Models**

In our investigation, we initially explored KeyBERT [36], which leverages BERT for extracting document embeddings, thereby achieving a document-level representation. Subsequently, word embeddings are derived for N-gram words, enabling a nuanced understanding of keyword significance within the document.

Not being satisfied with the results, which will be shown in detail in the Implementation Sec-tion 4.7, our research pivoted towards the exploitation of large language models (LLMs). These models, distinguished by their billions of parameters, excel in discerning complex patterns within textual data. Such capability is derived from their extensive pre-training across broad corpora, followed by fine-tuning for specialized tasks. Our exploratory phase commenced with the evaluation of open-source models, specifically Mistral 7B [48] and Llama 2 7B [110], subsequently extending our investigation to OpenAI's proprietary GPT-3.5 model [12]. The decision to harness these advanced models was primarily motivated by their proficiency in Few-shot prompting—a technique wherein LLMs are guided to execute specific tasks or enhance their performance through a minimal set of examples. This method capitalizes on the intrinsic knowledge and capabilities of LLMs, obviating the need for extensive dataset retraining or fine-tuning. The thorough analysis by Silviu Pitis et al. [80] delves into this phenomenon, providing a comprehensive understanding.

The strategic application of Few-shot prompting with LLMs enabled the development of a keyword extraction process finely tuned to our specific requirements. Contrary to general methods such as KeyBERT, which indiscriminately extracts keywords, our approach necessitates the selection of keywords optimized for image searches. For example, we want to minimize abstract keywords. The rationale behind this is the inherent challenge associated with querying images based on abstract concepts, although, as we will demonstrate in the Implementation Section 4.7, such queries are feasible and can yield meaningful results. Moreover, personal nouns, including names of cities or individuals, are deemed less pertinent for image searches, thereby warranting their exclusion. Leveraging Large Language Models (LLMs), we can intricately define these specifications within the prompt, alongside providing illustrative examples of both input and expected output. This approach, after meticulous prompt engineering, significantly enhances the quality of results.

### 3.8.2 Matching text and images

After extracting keywords, our objective is to evaluate the relevance of each frame-keyword pair and assign a corresponding score to assess how well the image aligns with the given keyword. To do so, we decided to utilize advanced vision-language models, specifically BLIP-2 [65] and CLIP [86], to assign a relevance score. BLIP-2, an acronym for Bootstrapping Language-Image Pre-training, represents a significant advancement in unified vision-language understanding and generation. It excels in various vision-language tasks, including image captioning, visual question answering, and image-text retrieval. The model undergoes a pre-training phase where it optimizes three objectives. One of these is the Image-Text Contrastive Loss (ITC) objective, which aims to align the feature spaces of the visual and textual transformers, enhancing the similarity of representations between positive image-text pairs in contrast to negative ones. Meanwhile, the Image-Text Matching Loss (ITM) objective involves a binary classification task to determine the congruency of image-text pairs, utilizing a dedicated ITM head for prediction.

Conversely, CLIP's pre-training methodology leverages a text encoder and an image encoder, based on a Vision Transformer (ViT) architecture, to generate semantic embeddings for text and images, respectively. The primary objective of CLIP's pre-training is to maximize the similarity between the embeddings of images and their corresponding text descriptions through a Contrastive Function, which employs Cosine Similarity 3.4 to adjust the model parameters for increased matching accuracy between image and text embeddings. During inference, CLIP evaluates the compatibility of text inputs with a set of images by comparing their embeddings through the same Contrastive Function used in pre-training, selecting the text with the highest similarity to an image embedding as the most suitable match.

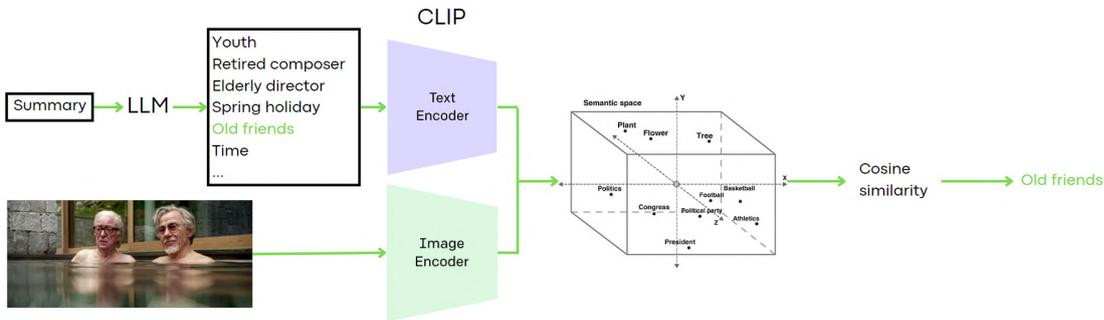

Figure 3.5: Visual representation of the semantic consistency step. The process involves extracting keywords from the video's text summary using an LLM, and pairing them with each image. These pairs are projected into the same semantic space to assess their semantic consistency using cosine similarity.

**Practical comparison of CLIP and BLIP-2**

Both CLIP and BLIP-2's methodologies incorporate the use of cosine similarity between text and image embedding vectors to gauge their relevance. However, BLIP-2 introduces the additional ITM loss, potentially offering a more nuanced approach for matching image-text pairs, which appears particularly suited for our application. Despite BLIP-2's more recent development and the sophisticated nature of its ITM score, which leverages an attention mechanism for a potentially more refined performance, our empirical investigations revealed that CLIP outperforms BLIP-2 in our specific context (more details provided in the Implementation Section 4.7). This unexpected outcome led to the selection of CLIP for our application, also due to the efficiency of its cosine similarity-based approach over the more computationally intensive ITM score.

In Figure 3.6 we show how clip works while a comprehensive visual representation of the process is shown in Figure 3.5.

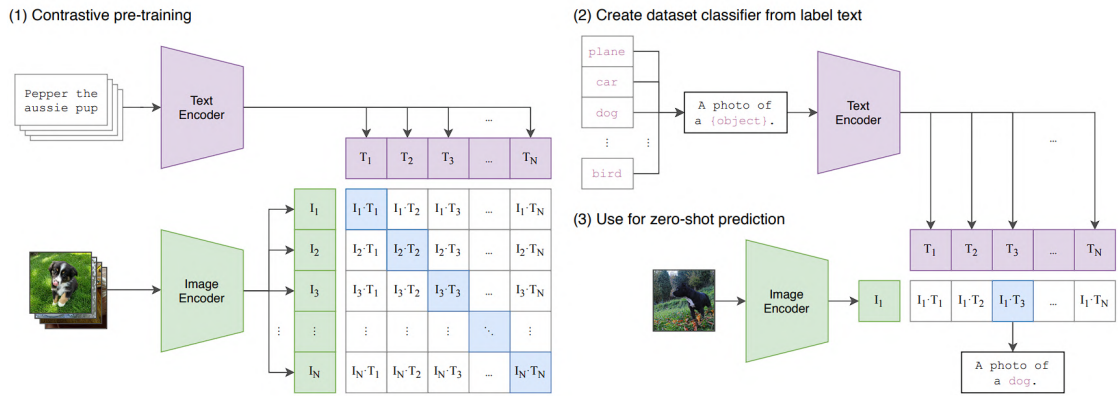

Figure 3.6: CLIP is an approach that involves training both an image encoder and a text encoder simultaneously. During training, they learn to predict the correct pairings of images and corresponding text descriptions. At test time, the trained text encoder can generate embeddings for the names or descriptions of classes in the target dataset. These embeddings effectively serve as a zero-shot linear classifier, allowing the model to make predictions for classes it hasn't seen during training. Image taken from the original paper [86].

In summary, our approach offers a customized keyword extraction technique tailored for image search, thanks to the remarkable few-shot prompting capability of Language Model (LLM), then it uses such keywords to find the most relevant thumbnails. This is done solely harnessing pretrained models, eliminating the need for any additional training.

**Refinements**

It's important to note that we also considered enhancing the embedding of each image frame through semantic segmentation using models such as SAM [53]. The idea was to extract the words of the objects within the image, use CLIP to compute their embeddings, and merge them with the thumbnail's embedding, obtaining a more detailed representation of the image in the latent semantic space. However, due to the already excellent performance we achieved, we ultimately opted to omit this step.

## 3.9 Shot scale classification

In the proposed pipeline, an important step involves the automatic detection of the shot scale within images. Shot scale—a fundamental concept in film and photography—denotes the spatial relationship between the camera and the subject. This relationship significantly influences the composition of the frame, dictating the extent to which the subject and its surrounding context are visible. The choice of shot scale is pivotal in storytelling, shaping the viewer's perception and emotional engagement with the scene. Commonly distinguished shot scales include Extreme Long Shot (ELS), Long Shot (LS), Medium Long Shot (MLS), Medium

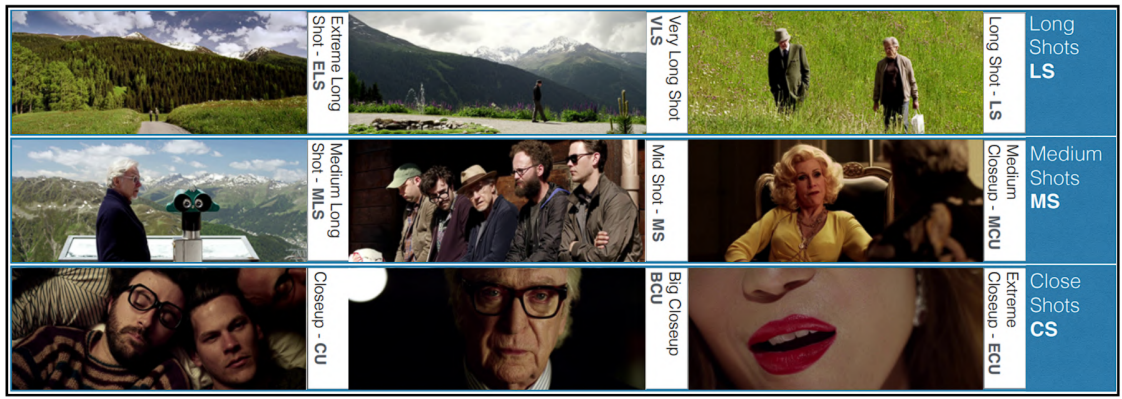

Figure 3.7: Shot scales examples.

Shot (MS), Medium Close-Up (MCU), Close-Up (CU), and Extreme Close-Up (ECU). For the purposes of this research, however, we categorize shot scales into three broader categories: Long, Medium, and Close-Up shots. Examples for each shot scale are provided in Figure 3.7.

A Long Shot (LS) encapsulates the subject entirely within its environment, striking a balance between the subject and its setting. This scale is instrumental in depicting the relationship between the subject and its locale. Conversely, a Medium Shot (MS) typically frames the subject from the waist up, making it particularly effective for showcasing expressions and body language within a contextual backdrop. A Close-Up (CU) shot, on the other hand, tightly captures a subject's face or a specific detail, emphasizing emotions or pivotal details.

The ability to recognize the shot scale of an image within our pipeline facilitates tailored processing and filtering of images based on their scale. This capability is especially valuable in contexts where the focus is on human subjects and their emotions, such as in cinematic productions, where Medium to Close-Up shots are often prioritized. Alternatively, in documentary filmmaking, which may concentrate on a broader array of subjects including objects, animals, places, and sports, the preferred shot scale may vary. Long shots are typically suited for documentaries emphasizing locations, whereas Close-Up shots are preferable for those focusing on animals or objects. The applicability of shot scale detection thus varies on a case-by-case basis, underscoring its utility in content-specific filtering and processing to propose the most suitable thumbnail candidate.

Moreover, this research explores thumbnail generation through the extraction of foreground elements from images and the subsequent substitution of the background with a contextually relevant alternative. Long Shot scales are particularly beneficial for sourcing suitable backgrounds, while Medium to Close-Up shots are more appropriate for foreground extraction.

### 3.9.1 Literature research

In addressing the challenge of shot classification, the decision to utilize deep convolutional neural networks (CNNs) was made with a clear rationale. These models have established themselves as exceptionally capable in the realm of image classification for a variety of applications, a fact well-illustrated by the notorious work on AlexNet by Krizhevsky et al. [55]. Their efficiency in processing and potential for being lightweight—depending on the architectural design—make them particularly suitable for tasks such as shot scale classification. This can be achieved either through training models specifically for this purpose or by applying fine-tuning techniques to pre-existing models to adapt them to the task at hand.

The literature and practical implementations explored on this topic, however, presented a mixed landscape. Many efforts were identified primarily through GitHub repositories, which often suffered from a lack of comprehensive documentation or detailed exposition of the training methodologies employed. These projects typically relied on fine-tuning pre-trained CNNs using relatively small datasets, which raises questions about the robustness and generalizability of the resulting models. Contrastingly, two notable exceptions were identified in the research. The first, by Anyi Rao et al. [88], introduced the Subject Guidance Network (SGNet), a novel framework for classifying shot types in video content. SGNet innovatively distinguishes between the subject and background within a frame, utilizing dual streams to serve as guidance for classifying both the scale and movement types of video shots. This model was trained on a significantly large dataset, consisting of 46,000 shots annotated for scale and movement, showcasing a comprehensive approach to understanding video content.

The second distinguished effort was conducted by Mattia Savardi et al. [98], which focused on training a CNN specifically for shot scale classification. This project amassed an extensive dataset of 792,000 shot frames from 124 movies, with each frame being extracted at a rate of one frame per second and annotated according to its shot scale. Initially leveraging the VGG16 architecture [103] and later incorporating DenseNet [45], this approach aimed at identifying visual features critical for distinguishing between different shot scales. Through the use of transfer learning from models pre-trained on general image recognition tasks, this research demonstrated the potential of CNNs to capture and categorize complex visual information relevant to the cinematographic analysis.

### 3.9.2 Chosen methodology

In selecting between these two methodologies, our decision was inclined towards the latter for various compelling reasons. Firstly, this approach employs a significantly larger dataset and incorporates state-of-the-art (SOTA) pre-trained CNNs, a combination which is anticipated to yield superior performance. Secondly, the classification of shot scale into three distinct categories—Long, Medium, and Close-up shots—aligns precisely with the granularity required for our application. In contrast, the alternative method categorizes shots into five divisions and extends its ambition to predicting shot movement. This additional complexity is deemed unnecessary for our objectives. The requirement by Anyi Rao et al. to process the entirety of a video for shot movement prediction poses integration challenges within our pipeline, as we operate primarily with filtered frames at this stage. Conversely, the method proposed by Mattia Savardi et al. is capable of classifying each image independently, which suits our workflow more effectively. A decisive factor in our selection was the availability of code and pre-trained weights from Mattia Savardi et al., which circumvents the need for from-scratch implementation and training. This is particularly advantageous for a component of the pipeline that, while important, is not central to our methodology.

However, it is important to acknowledge a limitation in opting for image-based classification over whole video analysis. Processing videos in their entirety could potentially provide the classifying network with additional context, thereby enhancing the accuracy of classifications. This advantage is forfeited when classifying images independently, as evidenced by the occasional inconsistent classification of similar images. To mitigate this issue, we adopted a strategy where all frames within the same shot are classified according to the predominant label determined by the model. This approach aims to enhance consistency across classifications and minimize the occurrence of isolated errors.

A visual overview of the model is shown in Figure 3.8.

## 3.10   Faces and Emotions

In the process of selecting thumbnails, a pivotal consideration is the emphasis on human subjects and their emotional expressions. This approach is underpinned by the recognition that thumbnails featuring faces, particularly those conveying strong emotions, are significantly more likely to capture viewer attention [118]. Major video streaming platforms such as YouTube and Netflix exemplify this trend, showcasing a plethora of thumbnails centered on emotive human expressions. However, it is crucial to acknowledge that this strategy does not uniformly apply to all types of content. For instance, nature documentaries may not benefit from a focus on human emotions, underscoring the need for a nuanced application of this principle.

Furthermore, while emphasizing people and their emotions is important, it is equally important to maintain diversity in thumbnail presentations. A uniform focus on human-centric thumbnails for content where human interaction plays a central role could potentially violate the principle of variety. Therefore, it is essential to ensure that, alongside thumbnails focusing on human emotions, alternative thumbnails that do not center on human subjects are also considered. This approach ensures a balanced representation that caters to the diverse preferences of viewers.

To address these considerations, our methodology incorporates several steps aimed at striking an optimal balance between emphasizing human emotions and maintaining content variety in thumbnail selection. The subsequent sections will delve into the specifics of these steps.

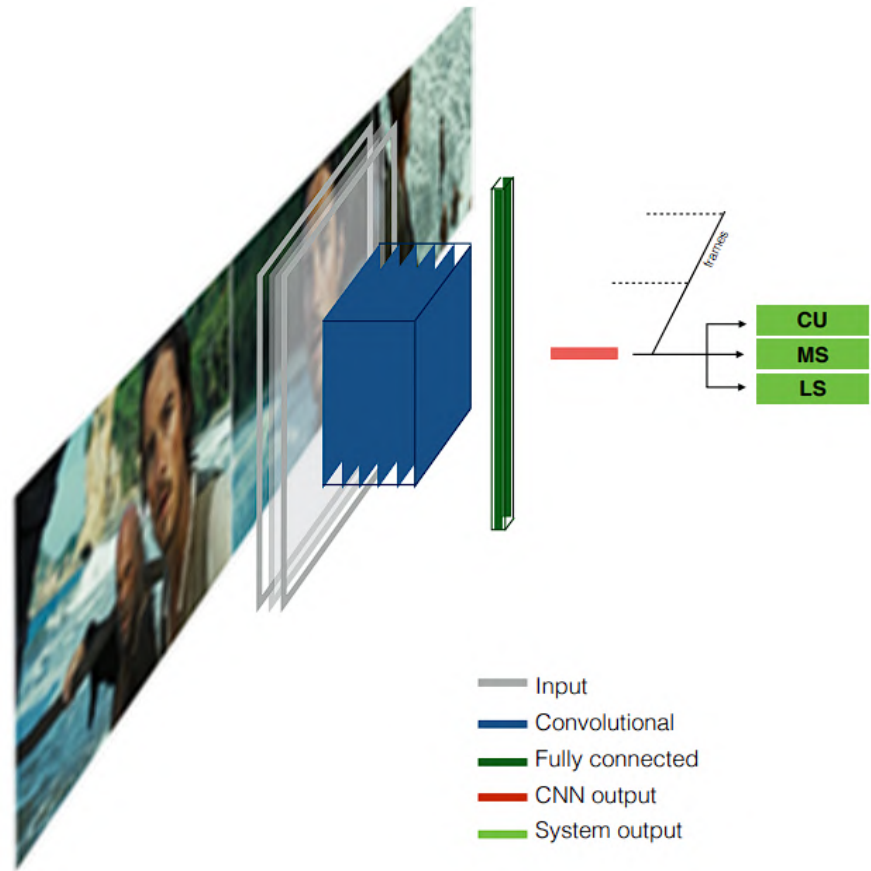

Figure 3.8: Representation of the CNN network for the shot classifier, taken from the original paper [98].

### 3.10.1  Face detection

The initial step in prioritizing images that emphasize human subjects involves the detection of faces within an image. This necessity underscores the importance of face detection in distinguishing images centered on people by identifying the presence of at least one face. Face detection, a significant domain within computer vision, encompasses methodologies designed to recognize and delineate faces in images through the demarcation of bounding boxes. These boxes, defined by four pixel coordinates, encapsulate the facial pixels, thereby facilitating further image processing and analysis.

**Literature research**

It is beyond the scope of this thesis to provide an exhaustive exploration of all available methodologies in the expansive domain of face detection. Notably, classical methods like the Histogram of Oriented Gradients (HOG) [23], introduced in 2005, are prevalently employed for face detection [46]. This technique partitions the image into small, interconnected regions termed cells, computing within each a histogram of gradient directions or edge orientations. The amalgamation of these histograms constitutes the feature descriptor, adept at capturing the edge or gradient structure emblematic of local shape. This characteristic renders it particularly efficacious for detecting faces across a spectrum of poses and lighting conditions. Conversely, Mahmut Çakar et al. [129] utilize the Haar Cascade [115], another conventional method which leverages Haar-like features to pinpoint facial characteristics and employs a cascade function to swiftly eliminate regions unlikely to contain faces, thereby markedly diminishing computational demands.

Despite the speed and lightweight nature of these classical methods, they are increasingly considered outdated, especially in their application to complex detection scenarios. These include challenges such as detecting small faces, faces viewed from oblique angles, the back, and instances where the risk of false positives—erroneously identifying faces where there are none—is heightened. In light of these limitations, there is a compelling argument for exploring more contemporary face detection methodologies that can offer enhanced accuracy and reliability in a broader range of contexts.

### Comparison of RetinaFace and DSFD

In our first experiments we adopted the 2019 RetinaFace framework [26], which for many years has remained the state of the art on different benchmarks. RetinaFace is renowned for its robust, single-stage face detection mechanism, adept at accurately and efficiently identifying faces of diverse scales in challenging environments. This proficiency is achieved through an innovative amalgamation of extra-supervised and self-supervised multi-task learning strategies. The method notably enhances its detection capabilities by incorporating manual annotations of five facial landmarks, which serve as additional supervisory signals. In parallel, it introduces a self-supervised mesh decoder branch aimed at predicting pixel-wise 3D facial shape information. This branch works in tandem with the existing supervised branches, significantly enriching the model's comprehension and interpretation of intricate facial structures. The combined efficacy of these components renders RetinaFace an exemplary model for our requirements, as illustrated in the Figure 3.9, showcasing the method's remarkable performance.

Despite the initial adoption of RetinaFace, subsequent experimental evaluations, which are elaborated in the Implementation Section 4.8, revealed that it did not entirely meet our expectations in performance. This led us to investigate alternative methodologies, bringing us to the 2018 Dual Shot Face Detector (DSFD) method proposed by Jian Li et al. [63]. DSFD revolutionizes face detection by introducing a Feature Enhance Module (FEM) that augments the original feature maps, effectively transforming the single-shot detection framework into a more formidable dual-shot system. Additionally, it employs the Progressive Anchor Loss (PAL), utilizing two distinct sets of anchors to optimize feature utilization and elevate detection accuracy. Moreover, DSFD integrates an Improved Anchor Matching (IAM) technique, which involves a novel anchor assignment strategy within the data augmentation process, fostering better initial conditions for the regression models. Although benchmarks suggest that DSFD might slightly underperform in comparison to RetinaFace, our practical assessments convinced us of its superior suitability for our specific needs. A visual representation of this method is provided in Figure 3.10, offering a comprehensive overview of its operational framework.

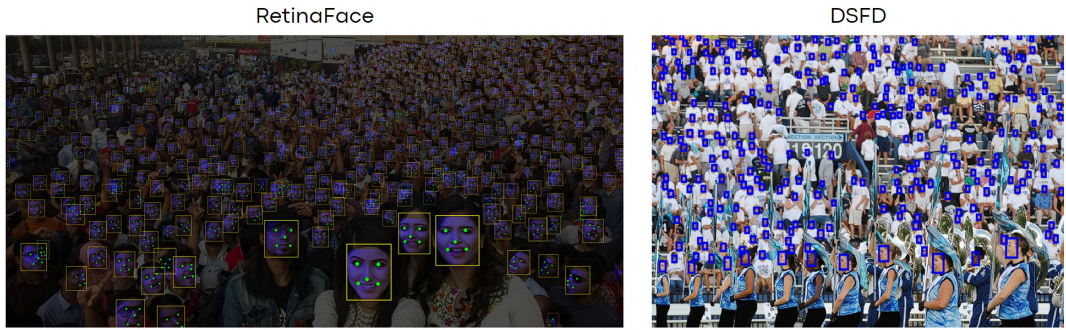

Figure 3.9: This image from the original papers shows two examples of the great performance of both the chosen method, initially RetinaFace [26] and then DSDF [63].

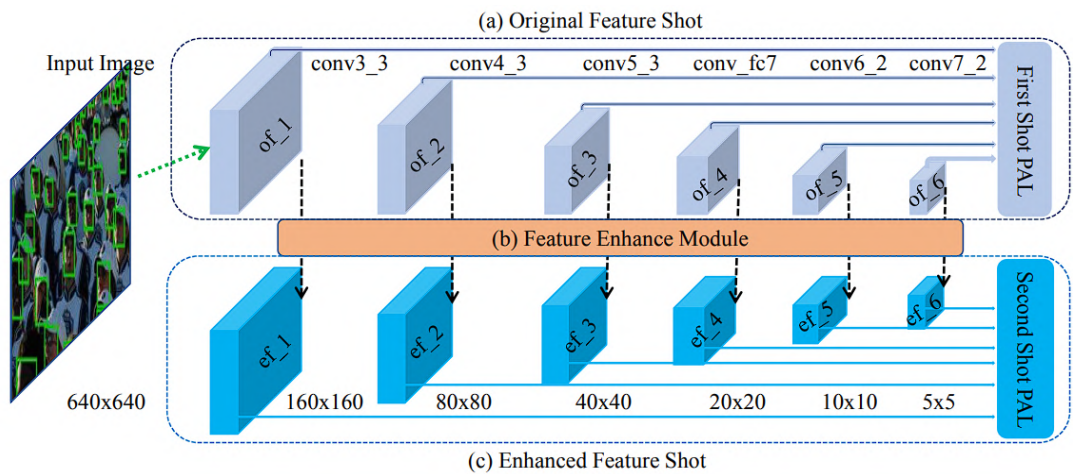

Figure 3.10: This image from the original paper [63] shows the DSFD framework. The figure illustrates a neural network architecture designed for face detection, which processes an input image through a series of convolutional layers to produce original feature maps (a), then enhances these features using a Feature Enhance Module (FEM) (b) for improved detection. The enhanced feature maps are evaluated using a Progressive Anchor Loss (PAL) mechanism (c), which assists in accurately detecting faces across various scales and conditions. This dual-shot approach, utilizing both original and enhanced features, aims to optimize the network's performance in challenging scenarios, such as varying face sizes, poses, and occlusions.

### 3.10.2   Eyes closed detection

Upon identifying the pixels within images that correspond to faces, subsequent processing can be done. A pivotal step in this process involves the detection of closed eyes, as images capturing individuals with their eyes shut are generally not preferred for thumbnail usage. The exclusion of such frames is instrumental in refining the selection of thumbnails proposed for use.

**Eye Aspect Ratio**

A prevalent method for detecting closed eyes within the realm of computer vision and facial recognition is the utilization of the Eye Aspect Ratio (EAR). The EAR serves as a quantitative metric to gauge the degree of eye openness or closure. It is derived by calculating the ratio of the sum of distances between specific vertical landmarks on the eyelids to the distance between horizontal landmarks across the eye. For a standard analysis, six facial landmarks are identified around each eye: two on the vertical edges and four along the horizontal edge. Mathematically, the EAR is expressed as follows:

$$EAR = \frac{||p_2 - p_6|| + ||p_3 - p_5||}{2 \cdot ||p_1 - p_4||}$$

In this formula, $p_1, ..., p_6$ represent the 2D coordinates of the six designated facial landmarks surrounding the eye. This calculation effectively averages the vertical distances (between points $p_2$ and $p_6$, and $p_3$ and $p_5$) and normalizes this value by the horizontal distance (between points $p_1$ and $p_4$). The Euclidean distance between two points is denoted by the norm $|| \cdot ||$. Figure 3.11 gives a visual representation of this formula.

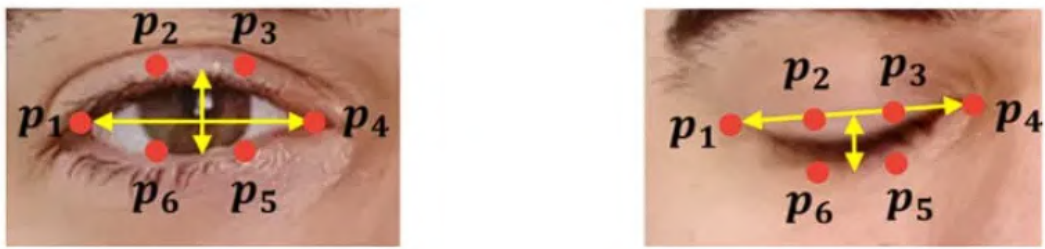

Figure 3.11: EAR calculation examples. Open eye will have higher EAR values (left), while closed eyes will have lower EAR values (right).

The EAR metric is indicative of the eye's state: a lower EAR value suggests a more closed eye, whereas a higher value indicates openness. By establishing a threshold for the EAR, it becomes feasible to detect instances of eye closure, such as during blinking or drowsiness. This simple yet effective metric has become a standard in eye state analysis in various applications, such as driver monitoring systems [51].

**Facial landmarks detection**

To be able to apply EAR on the other hand, we need to detect facial landmarks. In the realm of computer vision, the task of facial landmark detection is a common problem, involving the prediction of pivotal points on a human face that delineate critical regions such as the eyes, nose, lips, and other facial features. Our extensive survey of the domain to find a method tailored to our requirements led us to the innovative approach proposed by Xiaojie Guo et al. in 2019 [38]. This method, known as the Practical Facial Landmark Detector (PFLD), integrates a dual-subnet architecture comprising a backbone network and an auxiliary network. The former is responsible for the prediction of landmark coordinates, utilizing MobileNet blocks to facilitate a reduction in computational demands and model dimensions through the employment of depthwise separable convolutions, linear bottlenecks, and inverted residuals. The latter, an auxiliary network, aims to enhance the precision of landmark localization by incorporating a penalty for discrepancies between predicted and actual geometric angles (yaw, pitch, and roll).

Historically, the PFLD framework represented the state-of-the-art across various benchmarks, accurately identifying six points per eye to demarcate their boundaries. However, our empirical investigations revealed significant inconsistencies in its performance, highlighting a deficiency in robustness. This was particularly evident in its frequent inability to accurately detect eyes and ascertain the overall facial orientation in challenging facial poses, such as those observed from side angles—poses that are common in cinematic content. A thorough analysis of these outcomes will be elucidated in the Implementation Section 4.9.

**SPIGA**

Consequently, we selected a more computation-intensive but cutting-edge alternative, the 2022 methodology developed by Andrés Prados-Torreblanca et al. [83], which stands as the contemporary state-of-the-art in landmark detection across numerous benchmarks. This approach is encapsulated in the SPIGA model, an innovative integration of a Convolutional Neural Network (CNN) backbone with a cascade of Graph Attention Network (GAT) regressors. The CNN backbone is pivotal for providing a detailed local appearance representation of each landmark. It accomplishes this by extracting features within a square window centered on the landmark's location, progressively reducing the window's size through each cascade stage, adhering to a coarse-to-fine strategy.

The GAT regressors advance this model by learning the intricate geometrical relationships among landmarks. They achieve this through encoding the relative positions of landmarks into high-dimensional vectors, and subsequently applying an attention mechanism. This mechanism is designed to dynamically adjust the weight of information from each landmark based on its assessed reliability. Initialization of the model is conducted by projecting a generic 3D rigid face mesh, which is oriented according to the head pose estimation derived from the CNN backbone.

This method identifies nine points for each eye—eight marking the contour and one denoting the pupil. Our empirical analyses have demonstrated this method's superior consistency and robustness in results, compelling us to adopt it for our research.

### 3.10.3 Emotion detection

The extraction of faces from images enables us to identify the emotions on them as the next step. This process not only ensures the inclusion of human-centric images but also enhances the selection by identifying specific emotions, thereby aligning with the diversity criteria of our study. The ability to discern the emotional state of individuals in images—ranging from neutral to more expressive emotions such as anger, fear, or happiness—facilitates the tailoring of content to meet specific requirements and diversify the proposal. We can for example steer clear of images displaying neutral emotions and, instead, prioritize those depicting strong emotions like anger, fear, or happiness, which typically feature more prominent and discernible facial expressions.

**Literature research**

With this purpose, we conducted literature research on methods for emotion detection. Advancements in emotion detection from 2D images of faces have predominantly harnessed Convolutional Neural Networks (CNNs) and deep learning strategies, showcasing significant progress in both individual [8] and group [54] emotion analysis. Techniques range from real-time emotion detection models capable of processing video or photo inputs efficiently [75] to the application of transfer learning [34], where pre-trained models like ResNet [41], trained on extensive datasets, significantly boost accuracy. Additionally, integrating electroencephalography data with CNNs [116] has opened avenues for achieving remarkable classification accuracies, demonstrating the power of multimodal data fusion. Lightweight models like Mini-Xception [30] further exemplify the balance between computational efficiency and high accuracy in real-time applications, underscoring the diverse and innovative approaches.

**Chosen methodology: Multi-task EfficientNet-B2**

Given the constraints of relying solely on image data, our research led us to adopt a method developed by Andrey V. Savchenko et al. in 2022 [99], which integrates face detection, emotional feature extraction, and engagement prediction. This method employs a neural network architecture based on EfficientNet [109], trained on the AffectNet dataset [77], for the analysis of facial features. It is designed to aggregate facial features across frames to predict engagement and emotions in real-time, including on mobile devices. This choice was informed by the method's exemplary performance across various benchmarks, achieving a fourth position in the Facial Expression Recognition (FER) on AffectNet benchmark [126]. Despite its fourth-place ranking, the marginal difference from the top performer—which employs a computationally intensive attention mechanism—highlights the chosen method's efficiency. Its real-time processing capability, although not a necessity for our application, coupled with its implementation in a well-documented Python library adhering to the ONNX standard, an open format built to represent machine learning models, underscores its suitability for our requirements. Examples of results with the chosen method are shown in Figure 3.12.

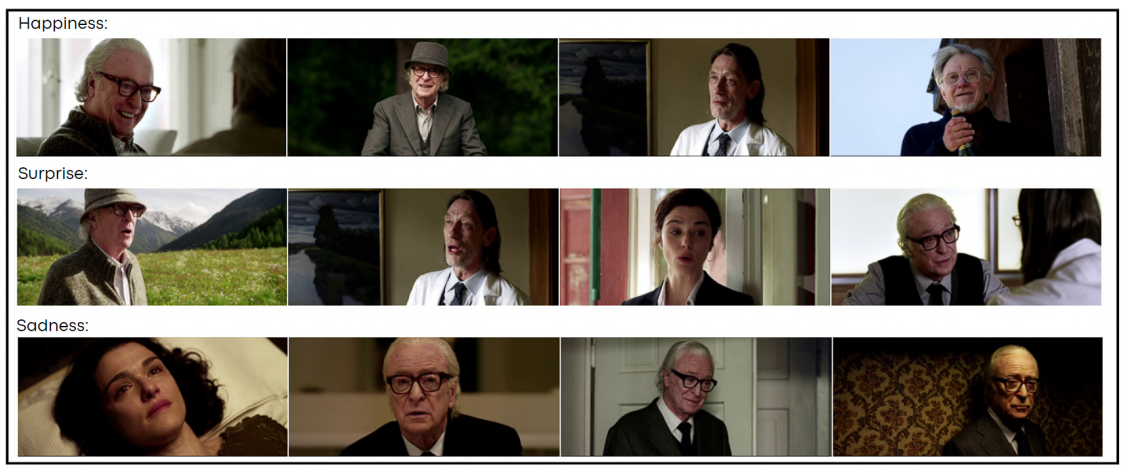

Figure 3.12: Examples of emotion recognition results obtained using Multi-task EfficientNet-B2 [99]. Three emotions are showcased here: happiness, surprise, and sadness.

### 3.10.4   Face recognition

One of the crucial steps for the face processing part involves the identification and consistent recognition of actors' faces across multiple frames. The objective is not merely to recognize the presence of a face but to accurately identify it and maintain awareness of its identity across different scenes. This capability is crucial for filtering and selecting images that feature specific actors, thereby ensuring that thumbnails represent main characters rather than peripheral ones. For instance, a thumbnail featuring a character who only appears briefly, despite its potential suitability based on other criteria, would not serve our purpose effectively. It is essential to highlight primary or secondary characters, such as a protagonist, antagonist, or a close associate of the protagonist, rather than incidental characters like a waitress in a transient scene.

An additional layer of complexity in this challenge stems from the initial unknowns regarding the number of faces that will appear in a video. While existing methodologies can identify faces by comparing them to a predefined dataset, thereby determining the closest match, such approaches presume the presence of all relevant faces within the dataset. This assumption would yield high performance under those conditions.

However, this assumption is untenable in our scenario. Even if we were to supplement the metadata from Play Suisse with cast information sourced through automated web crawling, this would not encompass all individuals appearing in the video, as comprehensive cast listings rarely include background actors. Moreover, obtaining cast information is particularly challenging for videos with limited budgets, which may have sparse online information.

Consequently, our challenge necessitates a method for recognizing all faces within the video content, subsequently grouping similar faces that are likely to represent the same individual. By identifying and analyzing the largest groups, we infer these faces to belong to the main actors, deduced from their prevalence across the footage. This approach, despite its complexities, fulfills our requirement for selecting the most appropriate thumbnails featuring main characters, thereby complementing our diverse selection process.

### Face embedding

In the domain of unsupervised clustering, the main option when dealing with unlabeled data as in our case, it is paramount to judiciously select the data upon which to execute clustering algorithms. This is especially critical in the context of clustering facial images, as clustering face images directly as they are would not work properly due to the extremely high dimensionality of images data points.

To address this, a variety of methodologies have been advanced within the realm of computer vision, aimed at extracting salient features from facial images—a process known as face recognition. With time the evolution of face recognition technologies has gravitated towards methods that project images into a latent space. In this space, representations of faces belonging to the same individual are engineered to be proximal, thus transforming images into vectorial embeddings. Within this spectrum of methodologies, we have selected the FaceNet method proposed by Florian Schroff et al. [100] in 2015, which is an highly influential work in the field of face recognition. This method from Google has remained at the forefront of face recognition technology for several years and has only recently been surpassed by newer approaches in certain benchmarks.

FaceNet's core principle involves mapping facial images into a compact Euclidean space to facilitate face recognition, verification, and clustering tasks. This is achieved through the use of a deep convolutional neural network, which is optimized using a triplet loss function. The triplet loss function plays a critical role in ensuring that images depicting the same individual are closely grouped together in the embedding space, while images of different individuals are separated by a significant margin. Additionally, FaceNet employs a strategy known as "Triplet Selection", which combines both offline and online techniques to identify "hard" triplets that effectively contribute to the learning process.

In our implementation, we have chosen to use an updated version of the architecture originally proposed in the Inception paper [107], pretrained on VGGFace2 dataset [14], as it strikes a balance between performance and computational resource requirements. While more advanced methods do exist, the differences in performance are not significant, especially when considering the impact of the clustering step, which we will elaborate on in the Implementation Section 4.10. Thus, our decision to adopt this method was based on its widespread implementation, documentation, and convenience for our specific use case.

### Face clustering

In the process of facial recognition, the critical step lies in grouping facial embeddings corresponding to the same individual under a unique identifier. This task mirrors the challenge addressed in the preceding Section 3.5, where the aim was to cluster CLIP embeddings of frames to minimize redundancy. The principal difference in this scenario is the nature of the embeddings and the vector space in which clustering is performed.

### Unsupervised clustering

When selecting an appropriate clustering algorithm for this application, it is essential to exclude any method that necessitates a predetermined number of groups, as we don't know the number of actors in the video. Our examination of prevalent clustering techniques identified several algorithms that do not require the specification of cluster quantity: Agglomerative Hierarchical Clustering [91], Mean Shift [21], OPTICS [3], DBSCAN [29], and Chinese Whispers [10]. Each algorithm embodies a distinct approach to data grouping. Agglomerative Hierarchical Clustering builds clusters hierarchically in a bottom-up fashion, ideal for visualizing data structures but computationally heavy for large datasets. Mean Shift locates the densest areas of data points without assuming cluster shapes, making it versatile for irregular clusters but potentially less effective in heterogeneous density settings. OPTICS extends DBSCAN's density-based clustering to better handle varying densities, offering a more flexible approach to spatial data analysis. DBSCAN excels in identifying clusters of similar density and separating noise, suitable for datasets with clear density distinctions. Chinese Whispers, a graph clustering algorithm, operates efficiently on large networks by iteratively updating node classes, though it may not directly address noise.

While these methodologies obviate the need for predefining cluster numbers, they nonetheless require the adjustment and fine-tuning of certain parameters to suit specific cases. Moreover, not all algorithms are equipped to effectively manage outlier data—a common occurrence in our context, where faces of background actors or main characters in atypical poses or sizes introduce significant noise. After conducting a series of experiments, we concluded that DBSCAN, following meticulous hyperparameter optimization, emerged as the most suitable choice due to its proficient handling of outliers. Although OPTICS is purported to enhance DBSCAN's functionality, it did not demonstrate superior performance in our application.

To reduce noise, we also excluded faces below a predefined resolution threshold from the clustering procedure. This decision was predicated on the understanding that the embedding process's efficacy diminishes significantly for low-resolution images. Moreover, such images, due to their minimal size compared to the whole image, do not contribute meaningful information for character and emotion representation, especially when scaled down to thumbnail dimensions.

### Dimensionality reduction

Prior to the implementation of clustering algorithms, our approach incorporates Principal Component Analysis (PCA) as a preliminary step to condense the dimensionality of the face embedding vectors. The determination of the final dimensionality is guided by an assessment of the explained variance retained by the principal components, adhering to the same principles applied in the Redundancy Reduction 3.5 phase of our research.

In the domain of dimensionality reduction, our exploration extended to alternative techniques, notably Uniform Manifold Approximation and Projection (UMAP) [76], t-Distributed Stochastic Neighbor Embedding (t-SNE) [73], and Non-negative Matrix Factorization (NMF) [59]. Each method offers distinct advantages for the simplification of high-dimensional datasets. UMAP stands out for its efficiency, scalability, and ability to maintain both global and local structural integrity in lower-dimensional spaces. t-SNE is particularly adept at revealing intricate data clusters through the minimization of divergence between dimensional probability distributions, proving invaluable for in-depth data analysis. Conversely, NMF offers a linear decomposition approach, facilitating the interpretation of datasets into constituent components and coefficients, which can be especially beneficial in applications such as topic modeling or signal analysis. While t-SNE is optimal for visualizing high-dimensional data, its purpose is different from clustering. Our empirical investigations led us to favor PCA for its overall utility in our specific context.

### Supervised clustering

We also explored the potential of supervised clustering techniques for grouping embeddings, as demonstrated in the method introduced by Yang et al. [120]. These methods have shown superior performance compared to unsupervised approaches. However, we identified some noteworthy concerns. These supervised methods are trained on a dataset that includes both the embedding model and the clustering network, that learn the data distribution during training. Afterwards, they undergo testing using the same dataset. Although the test data varies from the training data, it is drawn from an identical distribution.

This introduces a potential issue, as we would need to change the embedding model tot he same they used, and this change could result in a significant performance drop. Furthermore, the dataset used by Yang et al. to train their model comprises web images of celebrities, which is quite distinct from our movie-centric dataset. While some of our Play Suisse content features celebrities, many videos come from lower-budget production companies. Using a supervised method trained on celebrity web images could lead to reduced performance, given the differences in data distribution and the distinct nature of facial images between web photos and movie frames.

This concern is further validated by recent research conducted by Scott et al. at Google [101]. They found that deep learning methods tend to be vulnerable when dealing with embeddings characterized by higher uncertainty. In such scenarios, they may perform similarly or even worse than shallow heuristic-based methods.

Considering these factors and the limited time available for this stage of the pipeline, we made the decision to refrain from delving deeper into the integration of supervised methods with solutions such as creating our own training dataset. This choice is motivated by the already satisfactory results achieved with DBSCAN for our specific use case and needs. An example of such results is shown in Figure 3.13.

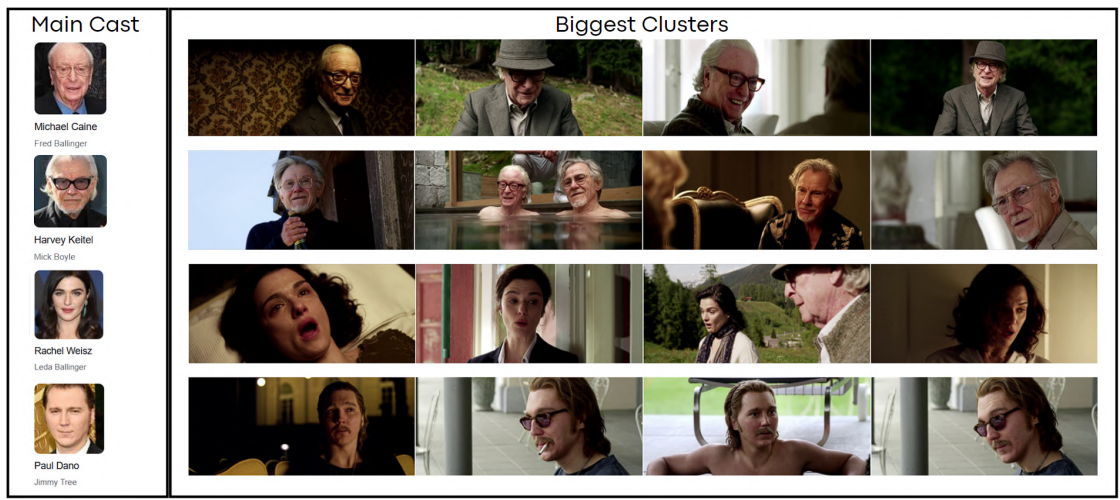

Figure 3.13: The results from face recognition are showcased with a comparative display. On the left, we present the original cast of the movie 'Youth', while on the right, clusters representing the corresponding actors, derived from the face recognition process, are depicted.

## 3.11 Scoring system

After the completion of all the steps previously mentioned in this chapter , a critical part is to understand how to leverage all the extracted data to propose the best images as thumbnails. To facilitate this, a scoring system has been devised, functioning as a quantitative framework to evaluate and compare images. This system employs a weighted average approach, integrating multiple scores that each underscore distinct attributes and characteristics of the images. Consequently, the images that obtain the highest overall score are adjudged to possess an optimal balance of these attributes, rendering them the most suitable candidates for a thumbnail. Figure 3.14 shows the scoring system visually.

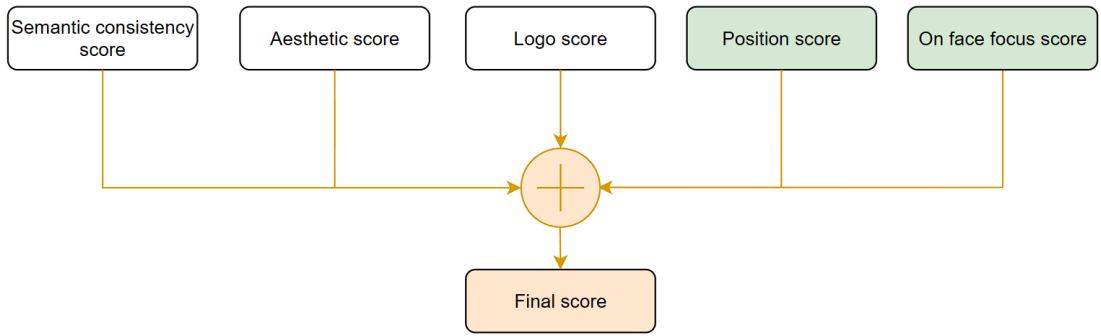

Figure 3.14: This image shows the scoring system. All the scores are summed together after being balanced by different weights and the final score is obtained.

### 3.11.1 Aesthetic score

This score is designed to give precedence to visually appealing images. The determination of this score is relatively straightforward, employing an aesthetic estimation model (as discussed earlier in Section 4.5) to assess the aesthetic quality of an image. The resulting aesthetic scores are scaled in a range from 0 to 1, where a higher score is indicative of superior aesthetic appeal. Conversely, images scoring at the lower end of this spectrum are typically characterized by issues such as blurriness, inadequate lighting, or suboptimal photographic composition.

### 3.11.2 Semantic consistency score

This score ensures that selected images are contextually relevant to the video content. This score is derived from the cosine similarity 3.4 measurements obtained in a previous step of the pipeline, where each image was compared against keywords extracted from the video description. By scaling these cosine similarity values in a range from 0 to 1, the system is able to quantitatively assess the semantic relevance of an image to a given keyword. For instance, in the context of the keyword "kid," an image featuring children would achieve a high Semantic Consistency Score, whereas an image depicting a landscape devoid of people would score lower. This score underlines the importance of representativeness, ensuring that thumbnail candidates are not only visually appealing but also contextually appropriate.

### 3.11.3 Logo score

This score is designed to prioritize images that offer ample space for placing logos, taking logo placement into account as a crucial criterion. The logo score quantifies the available space within an image to accommodate the title or logo of a movie. To compute this score, we leveraged the dataset comprising 8958 thumbnails of Play Suisse, manually created by professional designers. Employing an off-the-shelf optical character recognition (OCR) technique, we automatically identified the logo in each thumbnail, generating a probability map indicating potential logo placement areas. By tallying the frequency with which each pixel is utilized for logo placement across all thumbnails and normalizing it by the total number of thumbnails, we derived a matrix delineating the likelihood of logo placement in different regions of the image.

However, the mere identification of probable logo placement areas does not suffice for optimal thumbnail design. The composition of the image itself plays a crucial role in determining suitable locations for logo insertion. Specifically, images often contain salient areas that immediately draw the viewer's attention, such as a brightly colored object in a monochromatic scene. Placing a logo over these salient areas or over faces, which are often focal points of viewer attention, can detract from the overall impact and readability of the thumbnail. To address this, it was imperative to incorporate saliency prediction into our methodology.

### Literature research for saliency prediction

Our literature review in the field of saliency prediction revealed several advanced techniques for identifying areas within an image that are likely to capture immediate viewer attention. The method by Alexander Kroner et al. [56] introduces a novel convolutional neural network (CNN) approach for predicting salient regions in natural images, leveraging a pre-trained encoder-decoder structure. This methodology uniquely incorporates a module with convolutional layers at different dilation rates for capturing multi-scale features, alongside global scene information to enhance prediction accuracy. The method by Tanveer Hussain et al. [47], called Pyramidal Attention for Saliency Detection, uses only RGB images to estimate depth and leverage intermediate depth features, overcoming the limitations of RGB-based and RGB-D models, to predict saliency. The methodology employs a pyramidal attention structure that combines convolutional and transformer features to process and enhance stage representations, with a backbone transformer model producing global receptive fields for fine-grained predictions refined by a residual convolutional attention decoder.

### Chosen methodology: TranSalNet

Our preference was for the TranSalNet model by Jianxun Lou et al. [70], which is the current state-of-the-art according to different benchmarks. Their method combines convolutional neural networks (CNNs) with transformer encoders to address CNNs' limitations in capturing long-range contextual information. The methodology involves using a CNN encoder to extract multi-scale feature maps from input images, leveraging architectures like ResNet-50 [41] and DenseNet-161 [45], which are then enhanced by transformer [114] encoders for long-range context. This process aims to mimic the human visual system's ability to perceive saliency by integrating both local and long-range visual cues. A CNN decoder fuses these enhanced feature maps to generate the final saliency map. The model employs a loss function that combines four different metrics to optimize predictions towards human visual perception.

To compute the logo score, we first established a base probability map for logo placement using OCR-detected logo locations. We then adjusted this map by subtracting saliency values obtained from TranSalNet's predictions, effectively reducing the probability scores of areas identified as salient. Additionally, any regions identified as containing faces were also set to a zero probability to avoid logo placement over facial features. This comprehensive approach ensures that the final logo score reflects an image's suitability for logo placement, considering both traditional design preferences and the intrinsic visual focus points within the image. High logo scores are thus indicative of images that offer ample and strategically viable space for logo insertion, while low scores suggest a cluttered or focus-dense image landscape, limiting viable logo placement options. A visual representation of the process is shown in Figure 3.15.

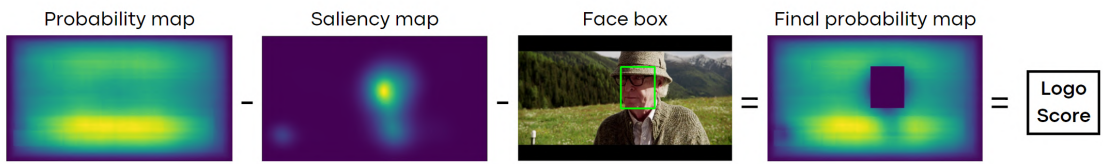

Figure 3.15: This diagram illustrates the logo score computation process. Initially, we begin with the default logo placement probability map. We then proceed to eliminate the predicted saliency map and adjust the probabilities to zero within the bounding box encompassing the detected faces. Next, we sum the remaining probabilities from the map to calculate the final score.

### 3.11.4 On-face focus score

In the pursuit of optimizing image selection criteria with a focus on human subjects and their emotive expressions, we initially implemented a metric termed the "On-Face Focus Score." This metric was designed to prioritize images where individuals and their facial expressions are prominently featured, initially correlating the score with the proportion of the image's area occupied by detected faces. Consequently, images featuring a multitude of faces or close-up views of faces occupying a substantial portion of the frame received elevated scores. However, this approach inherently favored images with disproportionately large facial representations, which, from a compositional perspective, are not always aesthetically pleasing or suitable for thumbnail usage.

Advancing beyond this preliminary method, we explored the utilization of predicted saliency maps—initially computed for assessing the logo score—as a refined means of calculating the On-Face Focus Score. In this enhanced methodology, the score is derived by assessing the percentage of attentional focus within the boundaries of detected facial regions. This adjustment enables the metric to assign high scores to images featuring faces of conventional size that nevertheless attract significant viewer attention, thereby aligning more closely with our objective to highlight images that effectively capture human faces and their emotional expressions, leading to the abandonment of the original method predicated on facial area size. A visual representation of the process is shown in Figure 3.16.

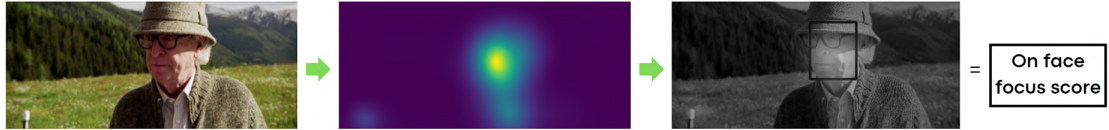

Figure 3.16: This diagram illustrates the on-face focus score computation process. Initially, we predict the saliency map for the image, and then we compute the percentage of the saliency map inside the detected face area.

### 3.11.5 Face position score

The face position score is designed to prioritize images where faces are centrally located, aligning with the criteria that emphasize the significance of human presence and emotional expression. This scoring mechanism is underpinned by the hypothesis that faces positioned centrally within an image are more prominent, thereby enhancing the visibility of their emotions. This concept draws inspiration from the study conducted by Tsao et al. [111], which proposes a scoring system based on the facial positioning within an image. According to this system, faces that are positioned either too high, too low, or excessively towards the edges of an image are assigned lower scores. This approach was further validated through consultations with Play Suisse's professional thumbnail designer, leading to the adoption of a scoring table represented in Figure 3.17. The methodology employed for determining the center of the face involves the midpoint of the line connecting the two pupils, as identified by the landmark detection model in preceding steps.

| 0 | $\frac{1}{5}$X | | $\frac{4}{5}$X | X |
|---|---|---|---|---|
| $\frac{1}{6}$Y | | 0.5 | | |
| $\frac{2}{6}$Y | | 0.75 | | |
| $\frac{3}{6}$Y | 0.1 | 1 | | 0.1 |
| $\frac{4}{6}$Y | | 0.75 | | |
| $\frac{5}{6}$Y | | 0.5 | | |
| Y | | 0.25 | | |

Figure 3.17: Face position weights from the original paper by Tsao et al. [111]

It is pertinent to note that this face-related scoring system assigns individual scores to each face in an image, thereby necessitating a mechanism to aggregate these scores for images containing multiple faces. To this end, two options were considered: calculating the average score of all faces or selecting the maximum score among them. The latter option was chosen, premised on the rationale that an image featuring a centrally positioned face of the main character would still be considered suitable for a thumbnail, even if other, less significant faces were located towards the periphery.

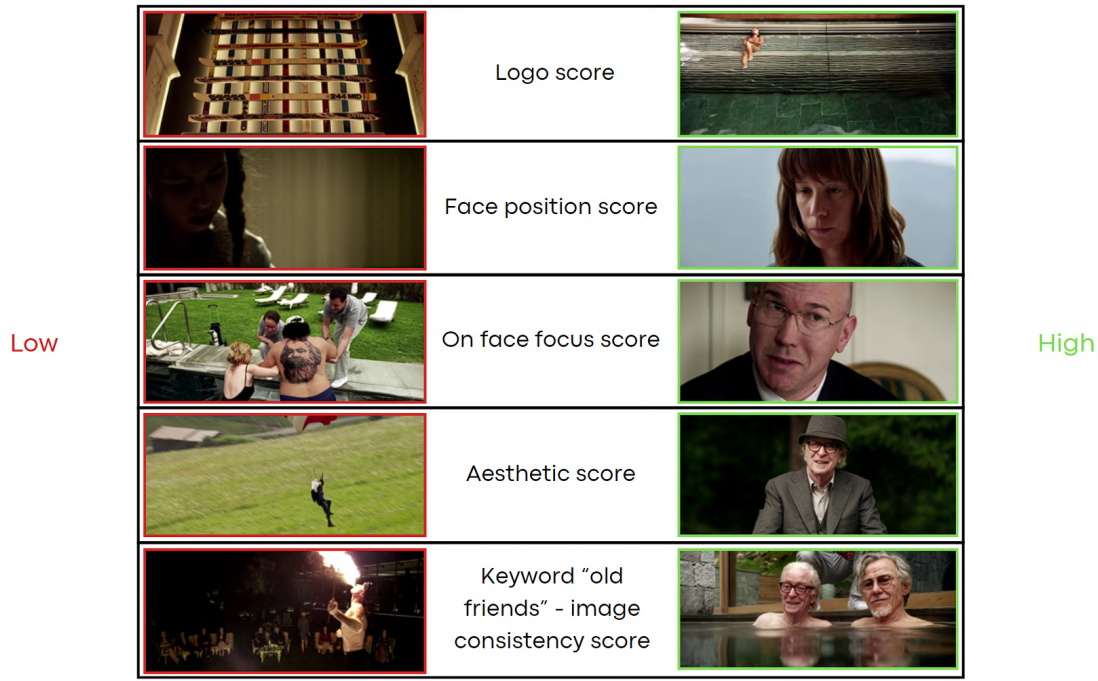

Figure 3.18: Examples of images with low and high scores for each one of the discussed scores.

Furthermore, it is essential to clarify that images devoid of faces are assigned a null score in the categories of face position and focus. Such images are not penalized in the overall scoring schema, as the final score is computed as an average of the applicable scores. This methodology ensures that images lacking faces are not at a disadvantage when compared to those with faces, even if the latter may have high scores in aesthetic, logo, and semantic categories but null face-related scores. The balancing of scores across different categories is meticulously designed to address our specific requirements, as detailed in the subsequent section.

We provide Figure 3.18 to represent the prioritization effect of each score.

## 3.12   Variety enhancement

In light of the scoring system elucidated previously, we are now equipped to allocate a numerical score to each image, facilitating the selection of images with the highest scores as prime candidates for thumbnail selection. However, the current pipeline exhibits notable deficiencies, particularly concerning the criterion of diversity. A primary issue is the homogeneity of scores among similar images, leading to a selection pool dominated by nearly identical images. This problem is addressed in the previous section on redundancy reduction 3.5, which proposes a solution by treating similar images as a collective entity, from which only the highest-scoring image is chosen for the final thumbnail proposal list.

Despite the implementation of this solution, the objective of achieving diversity remains incompletely fulfilled. The proposed images, while excelling in aesthetics, ample logo space, semantic relevance to all extracted keywords, and focus on well-positioned faces (when present), do not achieve true diversity. This approach, predicated on semantic relevance to an aggregate of diverse keywords, may result in images that, while superficially related to all keywords, lack significant relevance to any specific keyword, thereby failing to meet the representativeness criterion.

To address these shortcomings, we have embarked on the development of a more varied selection of proposed images by integrating diverse filters and scoring weight policies. Our goal is to curate the optimal portraits representing the main characters, the most accurate portrayals of each identified emotion, and the images that best encapsulate each extracted keyword. Considering that most scores vary across different aspect ratios, this process is conducted separately for both horizontal and vertical formats, ensuring a unique proposed list of thumbnails for each ratio.

### 3.12.1   Best main characters portraits

To derive optimal portraits of principal characters, our methodology initially excludes images devoid of faces, those featuring closed eyes, and long shots where faces are not the primary subject, given that a portrait inherently necessitates a closer focus on the subject's face. Subsequent to this filtration, we employ a clustering algorithm to identify major groups of faces, from which we select the highest-rated images based on all metrics except the semantic consistency score, which we deem irrelevant for this specific analysis. This process is repeated to ensure inclusion of secondary characters by evaluating images of faces not categorized within the main clusters. This approach ensures a balanced representation of both primary and secondary characters in the final selection. An example of selected thumbnails using this preset is shown in Figure 3.19.

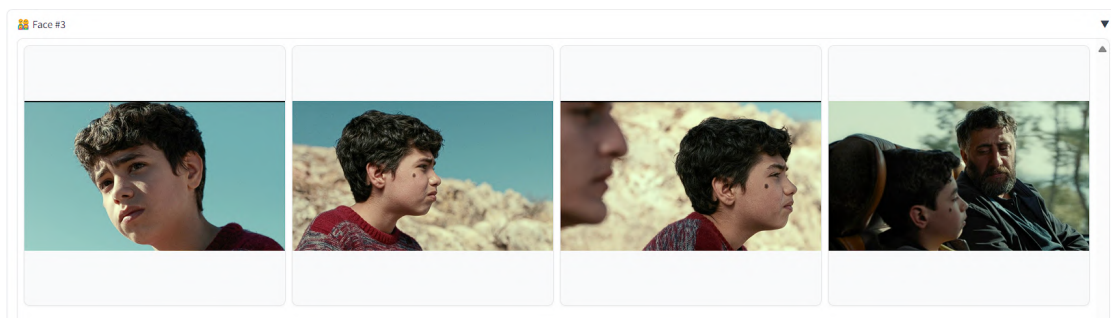

Figure 3.19: Example of four selected thumbnails using the "Best main characters portraits" preset. In this example the character is identified by the id "3".

### 3.12.2 Best portraits for each emotion

For the categorization of portraits by emotion, the procedure mirrors previously explained preset, excluding images based on the same criteria. However, the selection process pivots to focus on emotional expression, disregarding specific identities. For each identified emotion within the dataset, the best representations are chosen based on all evaluative criteria excluding the semantic consistency score. This method allows for a comprehensive representation of emotional range within the dataset. An example of selected thumbnails using this preset is shown in Figure 3.20.

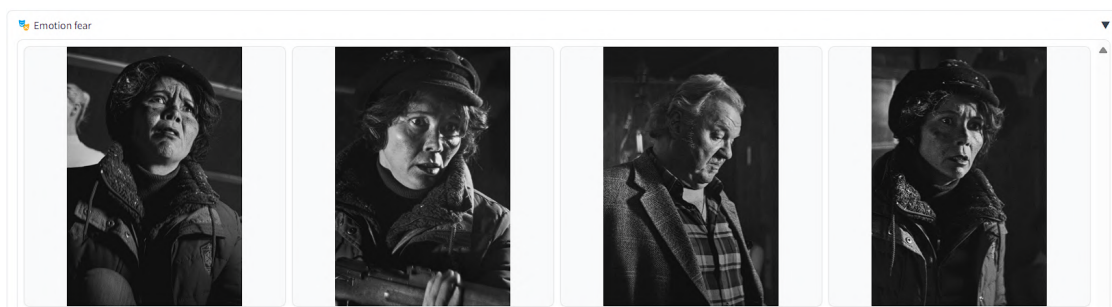

Figure 3.20: Example of four selected thumbnails using the "Best portraits for each emotion" preset. In this example the emotion displayed is "fear".

### 3.12.3 Best images for each keyword

Lastly, in the identification of optimal images corresponding to selected keywords, the process is inclusive of all images without the prior constraints of face focus and positioning. Here, the emphasis shifts towards semantic consistency, alongside aesthetic and logo scores, to ascertain images that most accurately embody the designated keywords. This approach is designed to yield images that are not only visually appealing but also semantically aligned with the keywords of interest. An example of selected thumbnails using this preset is shown in Figure 3.21.

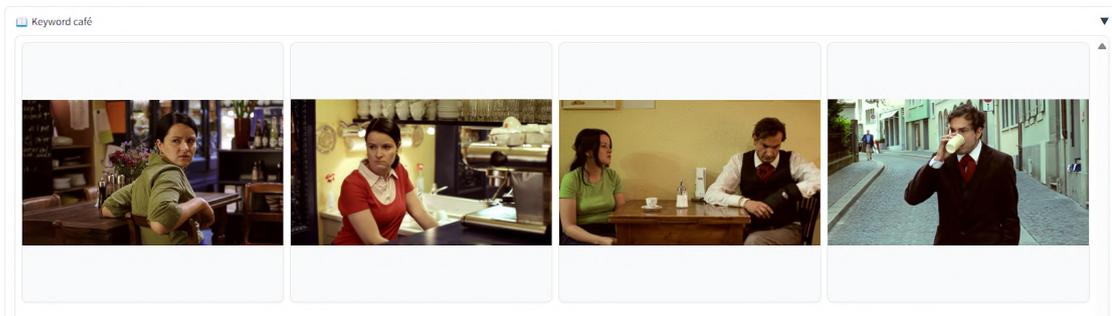

Figure 3.21: Example of four selected thumbnails using the "Best images for each keyword" preset. In this example the keyword displayed is "cafè".

In this methodology, we leveraged the comprehensive dataset derived from preceding phases, ensuring that every piece of data plays a pivotal role in curating a selection that is both optimal and varied. This strategy ensures that the curated images not only adhere to aesthetic standards and accommodate the placement of logos but also encapsulate a range of emotions and characters, thereby enhancing relevance. Specifically, while certain images are designed to evoke emotional responses by featuring key characters, others are selected for their ability to encapsulate the core themes of the video content, as delineated in the description, thereby satisfying the criterion of representativeness.

The emphasis on diversification is critical. For instance, in a film that intertwines elements of war with a romantic subplot, thumbnails emphasizing the "war" aspect are likely to appeal to aficionados of action genres, whereas those highlighting the "love story" are poised to attract a romantic film audience. This strategic segmentation facilitates Play Suisse's ability to target specific user groups based on their cinematic preferences, thereby optimizing viewer engagement.

## 3.13   Image enhancement and generation

As discussed in the introduction section, the primary objective of this project is to develop a methodology for selecting thumbnails from a vast collection of video frames, numbering in the hundreds of thousands, to identify approximately 50 images that accurately represent the video content. Concurrently, a subsidiary aspect of the project addresses the generation of thumbnails not originally present within the video. This process mandates that the generated image maintains a degree of content fidelity to the source material. The alternative of sourcing copyright-free images from the internet or employing a diffusion model for text-to-image synthesis [92], does not align with our criteria for content representativeness. Such synthe­sized images may not accurately depict the specific individuals, characters, objects, or settings present in the video, posing significant challenges for content such as movies where accurate representation of the main actors is crucial, and to a lesser extent, documentaries where the depiction of natural entities or elements might be less restrictive. However, even in documen-taries, a synthesized thumbnail might fail to convey nuances such as cinematographic quality or production value.

Notably, while AI-generated images for thumbnails are becoming increasingly prevalent on platforms like YouTube, and their usage is often criticized as clickbait—a term denoting the practice of employing sensationalized or misleading headlines or thumbnails to increase viewership—such practices are explicitly eschewed in contexts such as Play Suisse, where misleading representations are undesirable.

In response to these considerations, our approach involves the strategic composition of images extracted from the video itself. This method entails selecting optimal portraits of principal characters, extracting them from their original backgrounds to isolate the foreground, and subsequently integrating these foregrounds with backgrounds that are either derived from other frames of the video or generated through text-to-image diffusion models [92]. While opting for the latter choice may heighten the risk of generating clickbait, contradicting our earlier stance, we can mitigate this by employing advanced image generation techniques conditioned on video content. This approach restricts generated backgrounds to content resembling the videos, thus ensuring they align with our criteria for representational accuracy.

### 3.13.1 Foreground segmentation and matting

To obtain the goal described above, the first challenge is to segment the foreground from the background. The field of image segmentation has seen significant advancements through key contributions such as the efficient graph-based method introduced by Pedro F. Felzenszwalb et al. [31] (2004), which segments images based on pixel similarity. The concept of Fully Convolutional Networks (FCNs) for semantic segmentation was revolutionized by Jonathan Long et al. [69], enabling dense prediction over arbitrary-sized inputs. Olaf Ronneberger et al. [93] further specialized segmentation for biomedical applications with the U-Net architecture, known for its precision and efficiency. Liang-Chieh Chen et al. [18] enhanced semantic segmentation through DeepLab, combining deep neural networks with atrous convolution and CRFs for improved boundary accuracy. Additionally, Mask R-CNN by Kaiming He et al. [40] introduced a flexible approach for instance segmentation by adding a branch for predicting segmentation masks to the Faster R-CNN framework.

Beyond these foundational works, the literature encompasses a broad spectrum of segmentation methodologies tailored to diverse applications, including the specialized domain of foreground segmentation in portrait images, which primarily focuses on human subjects. One of the inherent challenges in segmentation is achieving precise edge definition, as standard segmentation techniques often yield masks with pixelated boundaries, particularly in complex regions such as hair and beards. This limitation underscores the distinction between conventional segmentation and the nuanced requirements of image matting. Image matting transcends binary segmentation by conceptualizing pixels as composites of foreground and background elements in varying proportions, represented by a soft mask or alpha matte. This advanced approach facilitates the refinement of segmentation masks through the generation of trimaps and the application of matting models. These models intricately assess bordering pixels to accurately differentiate between foreground, background, and mixed pixels, thereby addressing the precision requisite in applications such as thumbnail generation, where manual interventions have traditionally set the benchmark for quality. An example of the difference between utilizing the rough pixelated mask output of the segmentation process versus employing a matting model to refine and improve the final mask quality is included in Figure 3.22.

**Chosen methodology: Matte Anything**

To address these multifaceted challenges, our strategy incorporated a novel technique that synergizes the forefront methodologies for each constituent task, namely segmentation and matting. The pioneering approach by Jingfeng Yao et al. [121] heralds the Matte Anything (MatAny) model, an innovative solution for natural image matting. This model adeptly automates the production of pseudo trimaps, obviating the conventional need for manual trimap creation and thus streamlining the matting process. MatAny ingeniously integrates three distinct components to achieve its objectives: the Segment Anything Model (SAM) [53], which excels in crafting high-fidelity masks from minimal user inputs such as points and scribbles; the Open Vocabulary (OV) Detector (GroundingDINO) [66], which facilitates the identification of objects within images via textual prompts, thereby eliminating the necessity for manual input and automating the segmentation process; and the VitMatte model [121], which employs the aforementioned pseudo trimaps in conjunction with RGB images to accurately predict the final alpha-matte. This holistic integration not only simplifies the creation of high-quality mattes but also significantly elevates the output quality, showcasing the efficacy of task-specific vision models in enhancing process efficiency.

The Matte-Anything framework thus enables precise object segmentation through simple textual prompts, adeptly handling complex subjects such as hair and transparent objects (e.g., glass), where it accurately detects and eliminates the background visible through transparent sections from the final mask. This methodology stands out for its excellence in segmenting human subjects, while its versatility holds promise for future applications across diverse domains. Utilizing the text prompt "person," we successfully identified and segmented individuals within images, though the approach also supports more generic prompts such as "foreground" for broader applications. For instance, in the context of natural documentaries, it is conceivable to extract and automatically segment prominent images of specific animals using relevant keywords, without the dependency on specialized animal segmentation models. While the current implementation is focused on human subjects, the potential for future expansion and application, including manual prompt entry for tailored use cases by thumbnail designers, is evident.

### 3.13.2   Background replacement and Image harmonization

Following the segmentation of individuals from the original footage, we proceed to enhance the composition by integrating superior background imagery derived from the same video source. This process entails the selection of background images through a methodology analogous to the one employed for generating proposal images pertinent to designated keywords. The selection criteria are further refined by incorporating keywords associated with the quality attributes of photographs, specifically targeting long shots to eliminate unsuitable options. The optimal images thus identified are employed to substitute the original backgrounds.

Despite these efforts, the juxtaposition of elements from disparate sources often results in a discernible lack of cohesion, manifesting as suboptimal composite imagery. This outcome falls short of the high-quality standards envisaged, necessitating further refinement to achieve a seamless integration. The concept of image harmonization emerges as a critical intervention in this context, serving to enhance visual consistency between the foreground and background elements by adjusting the appearance of the foreground to align with the new backdrop.

#### 3D Relighting

In pursuit of advanced harmonization techniques, we examined the work of Rohit Pandey et al. [79], which introduces a groundbreaking approach titled "Total Relighting: Learning to Relight Portraits for Background Replacement". This method excels by not only adjusting coloration but also by implementing comprehensive relighting techniques. It necessitates an HDR lighting map, or environment map, which delineates the light sources surrounding the subject. The relighting module incorporates a geometry network to deduce per-pixel surface normals and albedo, alongside a novel lighting representation designed to simulate reflections accurately through advanced light transport mechanisms such as ray-tracing. Despite its sophistication, the requirement for an HDR lighting map presents a challenge, given the unavailability of such data for our backgrounds. While methods to infer these maps exist [33], their limitations and the added complexity render them impractical for the scope of our project.

#### 2D Image Harmonization

Consequently, our focus shifts to image harmonization techniques that operate within a two-dimensional framework, bypassing the need for 3D property considerations. Among the various methods explored, we have adopted the HINet technique proposed by Jianqi Chen et al. [17], which represents a novel approach to high-resolution image harmonization. HINet's architecture, an encoder-decoder model with distinct Multilayer Perceptrons (MLPs) for content and appearance, effectively addresses the challenges of achieving pixel-to-pixel harmonization in high-resolution images without excessive memory demands. By separating content extraction from appearance rendering and employing a Low-Resolution Image Prior (LRIP) strategy, the method minimizes boundary inconsistencies. Furthermore, the integration of an optional 3D Look-Up Table (3DLUT) enhances both control and comprehensibility, rendering this approach particularly appealing. The 3DLUT component, compatible with professional image editing software, facilitates optional fine-tuning by our thumbnail designers during the final color grading process, thereby ensuring optimal harmonization outcomes. Figure 3.22 provides a visual representation of the whole process.

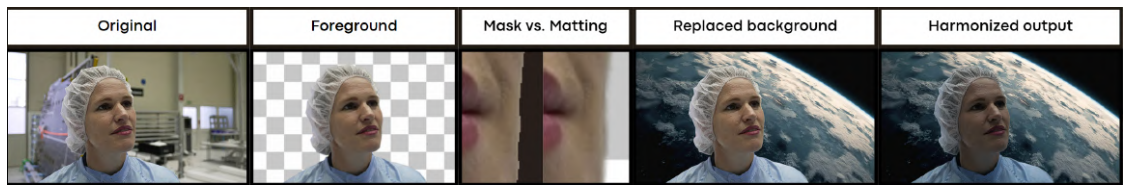

Figure 3.22: This image demonstrates the process of background replacement and image harmonization. The section labeled "Mask vs Matting" illustrates the distinction between solely utilizing the rough pixelated mask output of the segmentation process versus employing a matting model to refine and improve the final mask quality.

### 3.13.3 Background generation

To address the challenge of modifying image backgrounds while preserving the foreground, recent advancements in text-to-image diffusion models, such as the open-source Stable Diffusion proposed by Robin Rombach et al. [92], offer a viable solution. These models generate images from textual descriptions, aligning the output closely with the content described in the prompt. However, this approach encounters several obstacles when applied to our context. Firstly, maintaining the original foreground necessitates a method that allows selective alteration of the image parts, which is not inherently supported by these models. Secondly, the requirement for automated text prompt generation complicates the process, as it demands the creation of accurate descriptions without manual intervention. Lastly, the potential for generating misleading thumbnails that do not accurately represent video content must be mitigated.

To overcome the first of the mentioned problems, inpainting techniques for diffusion models, as discussed by Ramesh et al. [87], are employed. These techniques enable the specification of masks to designate areas within the image for generation, thus facilitating the creation of backgrounds that seamlessly integrate with the preserved foreground. The subsequent issues of automated prompt generation and ensuring content accuracy are addressed through the adoption of ControlNet, a novel architecture proposed by Lvmin Zhang et al. [126]. ControlNet incorporates spatial conditioning controls into pre-trained text-to-image diffusion models by utilizing zero convolutions—1×1 convolution layers with weights and biases initialized to zero. This ensures the undisturbed initial training stages and enables the application of various conditional controls through Classifier-Free Guidance Resolution Weighting during the inference phase. By allowing the conditioning of image generation on input images rather than text prompts, through various means such as poses, depth maps, segmentation masks, and object contours, ControlNet facilitates the creation of images that are both relevant to the video content and seamlessly integrated with the foreground, leveraging the capabilities of inpainting models.

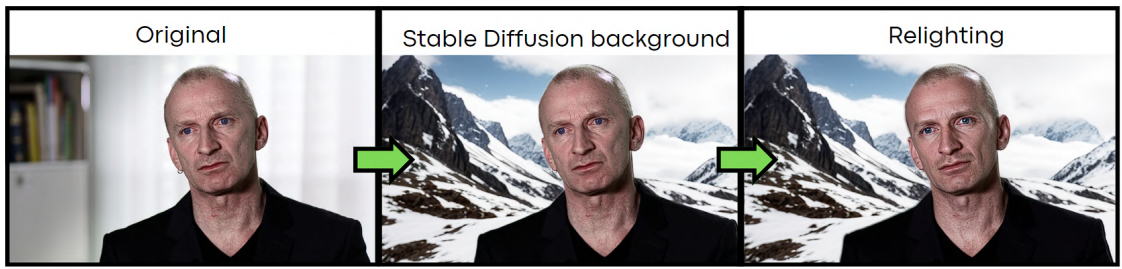

Figure 3.23: Example of background generation: the foreground elements are initially extracted from the image. Subsequently, the original background is removed and replaced with a newly generated one using Stable Diffusion [92]. Additionally, this method is employed to subtly adjust the lighting of the foreground elements to ensure coherence and visual harmony within the composite image, using a finetuned low value of InPaint conditional mask strength to avoid changing the face recognizability.

The exploration of alternative methodologies to ControlNet for enhancing background integration without compromising foreground prominence has led to the adoption of abstract or anonymous blurry backgrounds. This approach, leveraging default text prompts with minor automatic adjustments, aims to accentuate the foreground by utilizing backgrounds that are harmoniously blended by diffusion models based on foreground cues.

Although this strategy typically yields satisfactory harmonization, further refinement through inpainting—by inverting the mask to subtly adjust the foreground in response to the newly created background—has shown to improve both the foreground and background quality. This is particularly effective when conditioning the generation with specific keywords, as diffusion models are adept at generating high-quality portraits. However, preserving the recognizability of central characters poses a significant challenge, as neither ControlNet nor adjustments in the diffusion noise levels suffice to maintain their distinctiveness without altering their general appearance. A solution has been found in fine-tuning the InPaint conditional mask strength, a parameter that lets us control how strongly an image should conform to the original shape. This technique has also been tested for image harmonization in non-diffusion model-generated images, such as those with replaced backgrounds, but with limited success due to the lack of cohesive generation between the foreground and background. This concept becomes clearer when both the background and foreground are generated by the same diffusion model, allowing for more efficiency. When the foreground adapts to resemble the background generated by the same model, the process becomes more seamless and effective. An example of the final result is shown in Figure 3.23.

Further experiments have explored alternative methods for generating new images of main actors beyond video extraction. One such method is Dreambooth [95], which fine-tunes a pretrained text-to-image model to associate a unique identifier with a specific subject, enabling the synthesis of novel, photorealistic images of the subject in various contexts. Despite its potential, this technique faces challenges in automation, consistency, and the significant resources required for fine-tuning, rendering it impractical for each main character in every video. Another approach involves face swapping techniques designed for diffusion models, such as those introduced by Roop. While these methods effectively swap faces in the final image, they do not adequately address inconsistencies in hair and body representation, thus not fully resolving the issue.

By leveraging all the previous mentioned parts we are able to provide the highest level of diversity possible for our proposal.

## 3.14 GUI Tool

The final step of this project is to display the computed outcomes of the delineated pipeline in an intuitive, practical, and user-friendly graphical interface. This interface is designed to empower professional designers to swiftly identify and select optimal thumbnails from a video, subsequently facilitating the execution of final adjustments, such as color grading and logo placement. Concurrently, the interface aims to offer an adaptable platform that significantly enhances the user's efficiency in locating requisite resources. An illustrative example of the interface is presented in Figure 3.24.

### 3.14.1 Automated thumbnail proposition

The initial goal of the tool is to ensure such simplicity that even individuals without professional thumbnail design experience can effectively select an appropriate thumbnail. This objective is achieved through an initial section that displays the video's metadata, including title and textual summary, upon the selection of a desired video from a list of pre-processed videos. This feature is critical for conveying the essence of the video content without necessitating a comprehensive review of the entire video, a process typically required in manual workflows. Subsequently, a dropdown menu presents pre-selected thumbnails, curated by our method to highlight the most engaging images of principal actors, emotive expressions, and visuals pertinent to the video's central themes. This approach intuitively guides even the novice user towards making an informed thumbnail selection. An illustrative example of the panel is presented in Figure 3.25.

The ultimate objective of our research, which will be rigorously assessed in the experimental section, is to ensure that, in the majority of instances and across a broad spectrum of videos, professional thumbnail designers are consistently able to identify at least one viable option among those generated by our method.

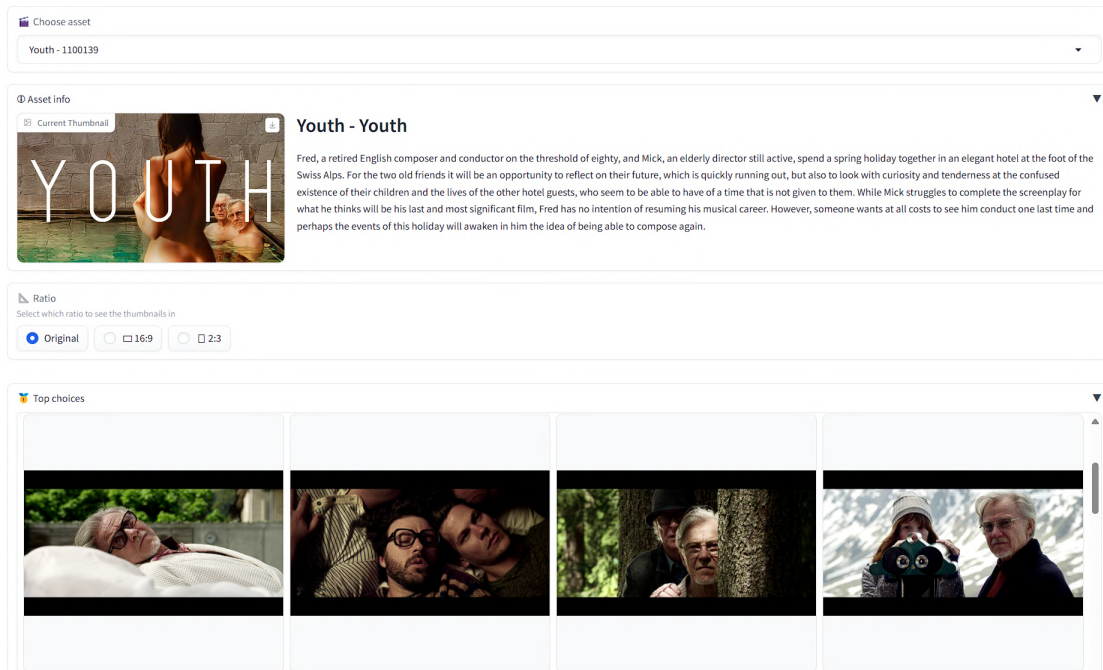

Figure 3.24: Snapshot of the upper part of the GUI Tool's interface. The user has selected "Original" aspect ratio and the best thumbnails are displayed below (only four are visible in the image).

### 3.14.2 Manual filtering panel

Furthermore, our tool is designed to serve a dual purpose by also catering to more specialized and targeted research needs. Specifically, in scenarios where a designer possesses a particular concept for a thumbnail, or in cases where none of the automatically generated thumbnails meet the user's requirements, our tool facilitates rapid and precise selection. This is achieved through the implementation of an advanced filtering panel, enabling users to refine their search based on a variety of criteria including the number of faces, expressed emotions, eye status (open or closed), actor identity, shot scale, aspect ratio, and semantic similarity to specified keywords or phrases. For users seeking further customization, the tool offers the capability to adjust the clustering algorithm to better detect faces according to user preference, and to modify the scoring system to prioritize aspects such as aesthetics, logo presence, focus on faces, facial positioning, and semantic consistency. Enhanced thumbnails' exploration is obtained allowing users to click on an image to reveal additional similar images from the same shot, available in both the original aspect ratio and alternative formats. For instance, if a user identifies a horizontal thumbnail that would be ideal in a vertical format, clicking on the image will unveil all corresponding vertical thumbnails derived from the horizontal shot and its similar. An illustrative example of the panel is presented in Figure 3.26.

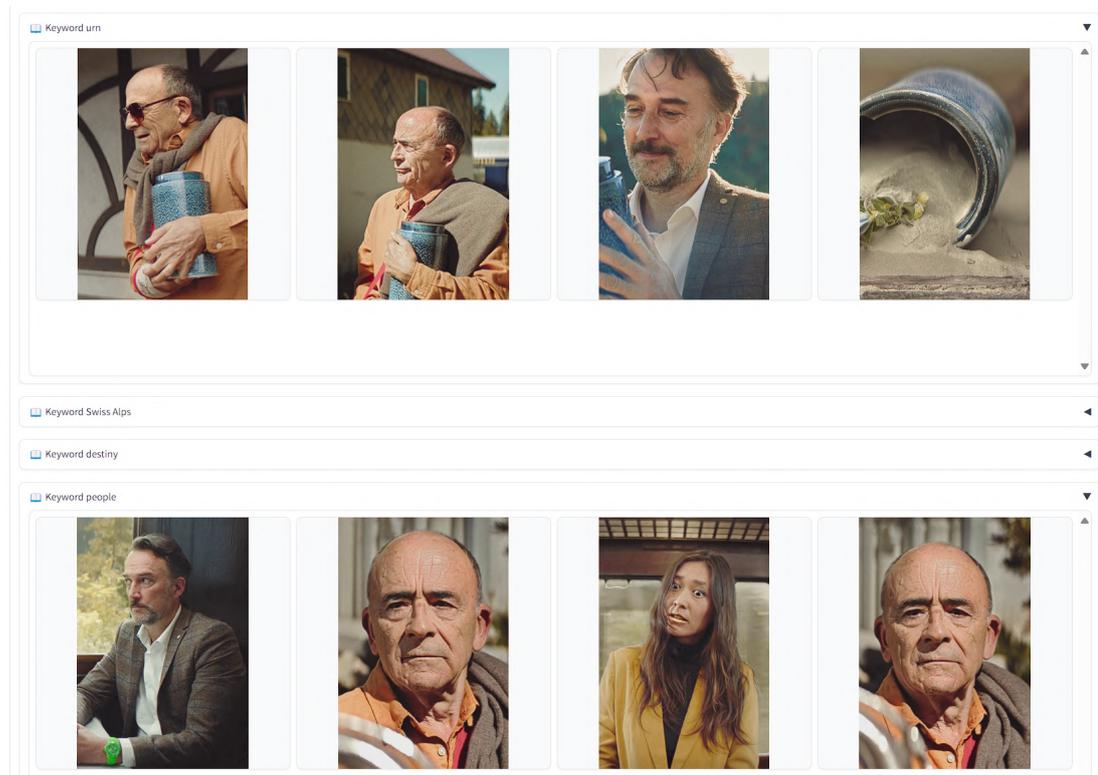

Figure 3.25: Snapshot of the automated thumbnail proposition panel interface. The user has opened the sub-panels with the keywords "urn" and "people", resulting in the presentation of the four top thumbnails. The user has specifically opted for a vertical aspect ratio, and consequently, the showcased results adhere to this preference.

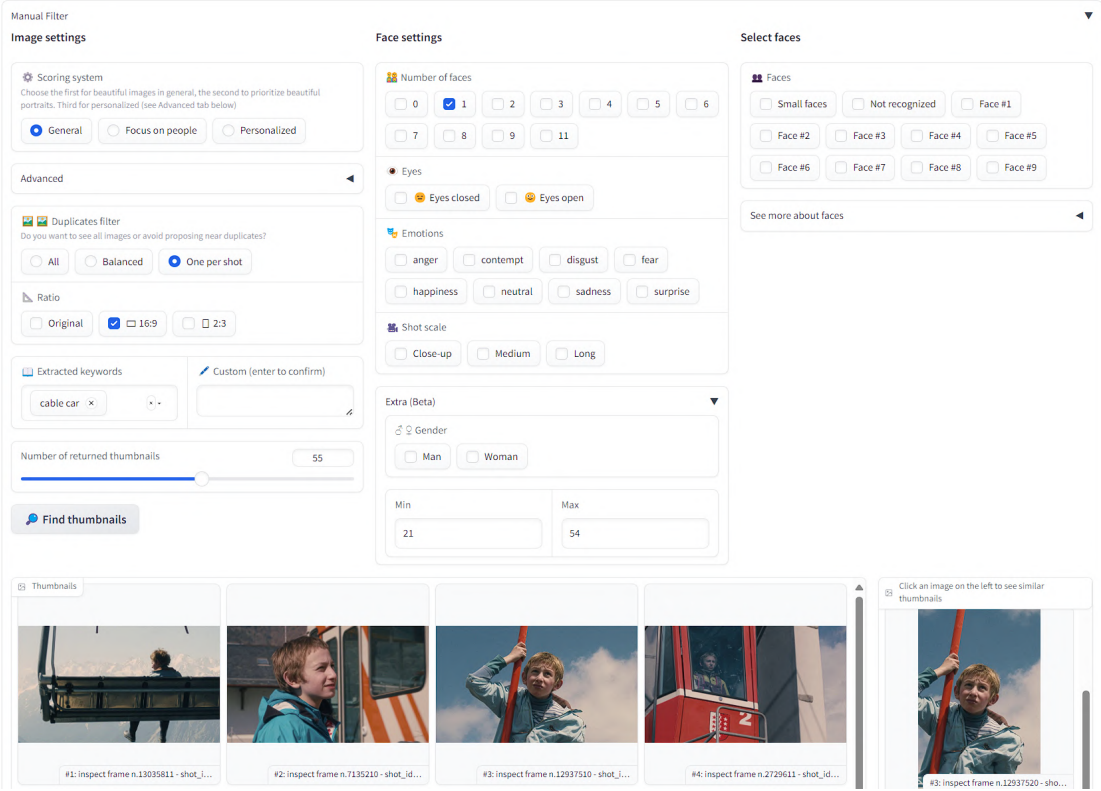

Figure 3.26: Example showcasing the interface of the manual filtering panel. The user has selected "1" for the number of faces, opted for a "16:9" aspect ratio, and activated the keyword "cable car". Displayed below the filtering options are the first four search results. On the right-hand side, the user can find a corresponding vertical image from the same shot as the third result, which was previously clicked.

# 4 Implementation

*This chapter delves into the implementation details of the thesis' project, starting with an introduction to the chapter in Section 4.1, including details on the Azure cloud platform used for hosting. It then discusses the technicalities of downsampling and Docker container utilization in Section 4.2, focusing on the adaptation challenges and solutions. The redundancy avoidance strategies and their implementation are examined in Section 4.3, highlighting the use of different feature extraction methodologies and clustering techniques. Section 4.4 addresses the cropping process, including the removal of letterboxing and code adaptations to the integrated library. The aesthetic prediction mechanisms are explored in Section 4.5, detailing the selection process and the limitations of the NR-IQA benchmarks. Tag extraction methodologies utilizing large language models are discussed in Section 4.6, including a comparative analysis of different models and the final choice. Section 4.7 elaborates on the semantic consistency assessment, comparing CLIP and BLIP 2 methodologies. The challenges and solutions related to face detection are presented in Section 4.8, followed by the advanced techniques for landmark detection and EAR in Section 4.9. Section 4.10 delves into face identification, clustering strategies, and their fine-tuning. Time and memory requirements are defined in Section 4.11. Adjustments to the pipeline based on professional thumbnail designers' feedback are discussed in Section 4.12. The scoring system and its purpose for variety in thumbnails are explained in Section 4.13. Image enhancement techniques, including foreground segmentation and matting, are detailed in Section 4.14, while Section 4.15 discusses the image generation's integration and challenges. Finally, the development and optimization of the GUI tool are covered in Section 4.16, emphasizing its design choices and user interaction features.*

## 4.1 Introduction

In the preceding chapter, this thesis scrutinized the obstacles encountered by our project, with a primary focus on the theoretical exploration of solutions to these challenges. This chapter delves into the technical difficulties encountered during the implementation, adaptation, and integration of previously mentioned methods into our operational pipeline. The culmination of this endeavor involves the deployment and hosting of the entire pipeline and graphical user interface (GUI) tool on the company's servers. These servers utilize Microsoft Azure, a cloud computing service devised by Microsoft, which facilitates the construction, testing, deployment, and management of applications and services via Microsoft-managed data centers. The analysis within this chapter will adhere to the structure established in the methodology section, examining the challenges associated with each phase of the pipeline.

## 4.2 Downsampling

### 4.2.1 Docker container

The initial phase of our research framework involves the downsampling process, leveraging an adaptation of the methodology proposed by Song et al. [105]. The entirety of our pipeline is implemented in Python; however, the referenced library was developed in C++. An initial attempt to reimplement the core components in Python resulted in significantly slower performance, attributable to C++'s superior efficiency due to its lower-level operations. Consequently, we encountered challenges in compiling the C++ code, primarily due to issues in resolving and compiling various dependencies. Our ultimate resolution involved the utilization of Docker, an open-source platform that facilitates the creation, deployment, and management of virtualized application containers on a shared operating system. This approach enabled us to encapsulate all necessary system requirements and library dependencies within an isolated container, which, once configured, allows for execution across any Docker-supported system, including Microsoft's Azure cloud servers. This solution not only circumvented the aforementioned compilation issues but also presented Docker as the optimal choice for our requirements. We therefore automated the Docker engine startup, the container creation from the built image, and the subsequent container destruction post-execution.

### 4.2.2 Code adaptations

In terms of code adaptation, our focus was on integrating the downsampling and shot detection components of the original method. Our contribution includes modifications to save images immediately following the downsampling process rather than at the conclusion of the pipeline, and to store shot detection data in a CSV file format. Notably, the original method is characterized by substantial memory consumption, primarily due to its storage of all video frames in RAM. Given a hardware limitation of 16GB of RAM, processing videos exceeding 75 minutes in duration resulted in system crashes. The original implementation provided a feature to mitigate this issue by halving the frame rate from the standard 24fps to 12fps for videos exceeding a predefined duration threshold, thus reducing memory requirements. However, a bug was identified that compromised this functionality, which we successfully rectified, thereby enabling the processing of longer videos without necessitating high-specification hardware.

Additionally, we modified several default parameters, including the retention of video seg-ments from the initial and final minutes, and the preservation of frames in their original resolution. We also implemented a feature to extract and store video metadata, such as frame number, frames per second, duration, and other relevant information. Our empirical analysis, conducted across twenty videos, revealed that the adapted method resulted in the extraction of only 2.79% ± 0.2% of total video frames on average, compared to 4.16% that would have been extracted using a uniform downsampling algorithm at 1fps. This outcome underscores the method's efficacy in filtering extraneous frames while efficiently selecting keyframes based on empirical thresholds outlined in the methodology, as opposed to the arbitrary selection inherent in uniform downsampling. Importantly, the method's scalability is non-linear, demonstrating that an increase in video length does not proportionately increase the number of extracted frames, thus avoiding data redundancy in extended videos.

Our analysis also involved comparing the current thumbnails, which were manually selected by professional designers, with frames extracted using our integrated method. The results consistently showed that either the exact frame or one with minimal temporal differences was included. Out of 69 videos tested, 13 had thumbnails sourced from images outside the video content, leaving us with 56 videos where perfect match was possible. Remarkably, in 51 instances, the closest image matched almost perfectly, affirming the method's capability to preserve critical frames while reducing the overall frame count. These findings are visually depicted in Figure 4.1, demonstrating the remarkable similarity between the obtained images. The image that does not perfectly match is attributed to an edited image.

| | Current thumbnail | Closest image after downsampling | |
|---|---|---|---|
| Cosine similarity: 0.673 | 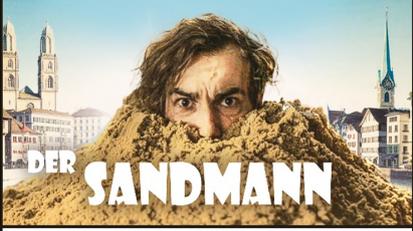 | 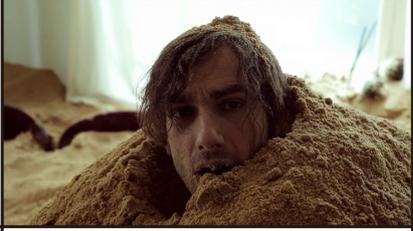 | Worst match |
| Cosine similarity: 0.8853 | 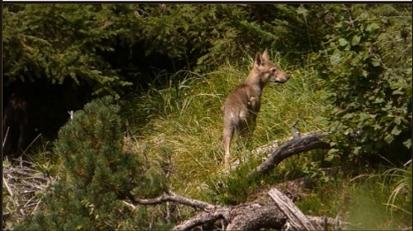 | 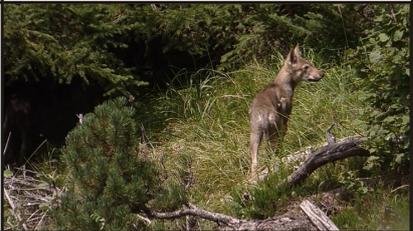 | Mean match |
| Cosine similarity: 0.9917 | 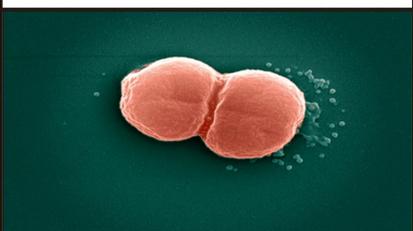 | 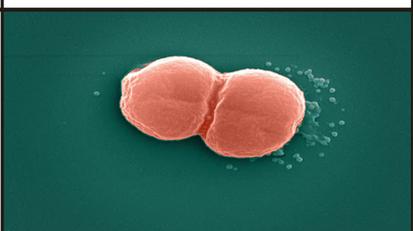 | Best match |

Figure 4.1: The image displays the current thumbnails for a selection of videos from Play Suisse's catalogue, along with the closest image with the highest cosine similarity from the extracted frames after downsampling. In the top row, we observe the poorest match out of the 69 videos tested. This mismatch can be attributed to the thumbnail being edited, making it impossible to extract the exact same frame. Moving to the middle row, we find the average match, representing the typical level of similarity across the dataset. Finally, in the bottom row, we encounter the best match achieved during the analysis.

## 4.3   Redundancy Avoidance

In the quest to optimize the clustering process for grouping similar images, our study embarked on an exploratory journey utilizing two distinct feature extraction methodologies. Initially, features were derived using a pretrained model based on the VGG architecture [103]. Subsequently, we employed embeddings obtained from the CLIP model [86], aiming to ascertain the most efficacious feature space for our clustering endeavors. The comparative analysis of the outcomes from these experiments is delineated in Figure 4.2.

The experimental framework was further enriched by fine-tuning a suite of hyperparameters, notably the number of components for Principal Component Analysis (PCA) to condense the embedding dimensionality, alongside the epsilon and minPoints parameters of the DBSCAN algorithm. Epsilon delineates the radius for density checks around each data point, transforming into a hypersphere in multidimensional spaces, while minPoints, set at one, ensures each frame is allocated to a cluster to avert the amalgamation of dissimilar images as noise. Through meticulous fine-tuning, it was observed that the CLIP model exhibited superior performance in forming larger, more coherent clusters without amalgamating disparate image groups.

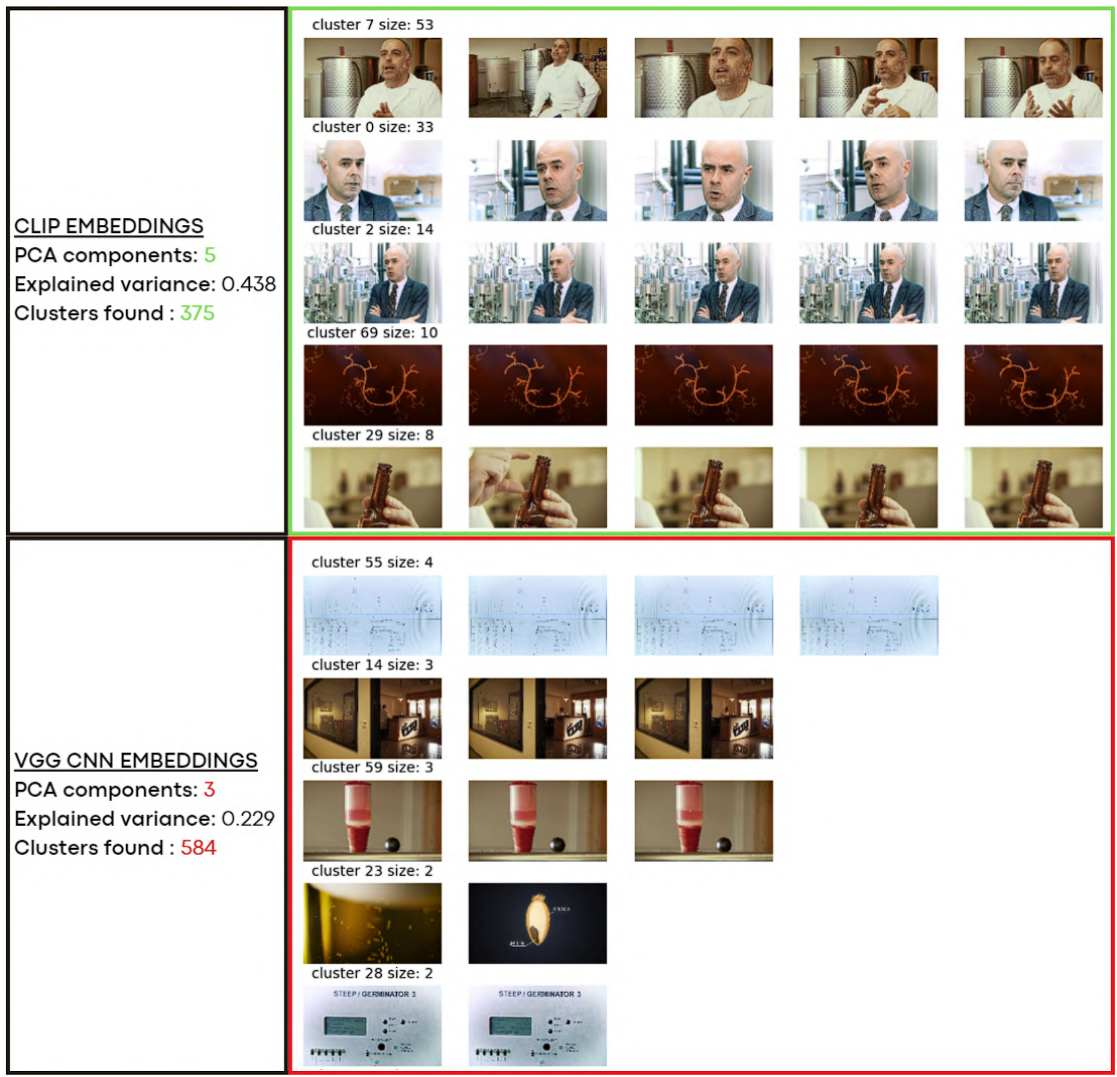

Figure 4.2: This image shows a comparison of the five biggest clusters (five random images from each) obtained using CLIP [86] and VGG embeddings [103]. Even with a very low number of components for PCA, the clusters generated by VGG are smaller in size compared to those produced by CLIP, resulting in an excessive number of clusters.

Following the selection of CLIP for feature extraction, a comparative study was conducted across three different videos to identify the threshold of explained variance retained post-PCA reduction. This threshold was critical to prevent disparate images from being erroneously clustered together. The investigation revealed a critical value of 0.43 for the explained variance, below which the likelihood of dissimilar images being grouped increased. Figure 4.3 illustrates how varying the number of components affects the resulting number and size of clusters, as well as the explained variance.

Our methodological approach also explored various granular levels of redundancy reduction techniques, testing methods that delineated scenes, shots, and clusters based on clip embeddings. Highlighting the outcomes from one video, originally comprising 4427 frames extracted over 36 minutes, we identified 924 scenes, 1973 shots, and 3398 clusters through these methods. A synthesis of shot information and CLIP clustering further refined the grouping, culminating in 1890 distinct groups.

These results underscore the limitations of scene-based grouping due to its broad categorization, and the propensity of CLIP-based embeddings to create excessive groups. Conversely, the amalgamation of shot information with CLIP clusters emerged as the most balanced approach, achieving an optimal grouping of similar images while avoiding overgeneralization or excessive fragmentation. Figure 4.4 shows the difference between using no redundancy avoidance technique, grouping by shot information, and merging shot and CLIP clustering info.

## 4.4 Cropping

In the methodology section, we delineate our approach to creating a set of potential cropping possibilities tailored to the required aspect ratio. This process involves initially generating a wide array of cropping options, followed by a meticulous filtering process that excludes unsuitable choices, such as those that bisect faces, subsequently ranking the remaining options from most to least favorable.

### 4.4.1 Letterboxing removal

A critical preliminary step in this methodology is the removal of letterboxing, which refers to the black bars that are added to the top and bottom of frames to adjust the original aspect ratio of the video to the desired one. This adjustment is necessary because all videos on Play Suisse are presented in the standard horizontal format of 16:9, the common aspect ratio for most screens and televisions. However, each video retains its unique production aspect ratio, necessitating the use of letterboxing. Our processing, on the contrary, does not require letterboxing; failing to eliminate it prior to cropping could result in selecting a crop that includes undesirable black pixels at the top or bottom as the optimal candidate, as illustrated in Figure 4.5.

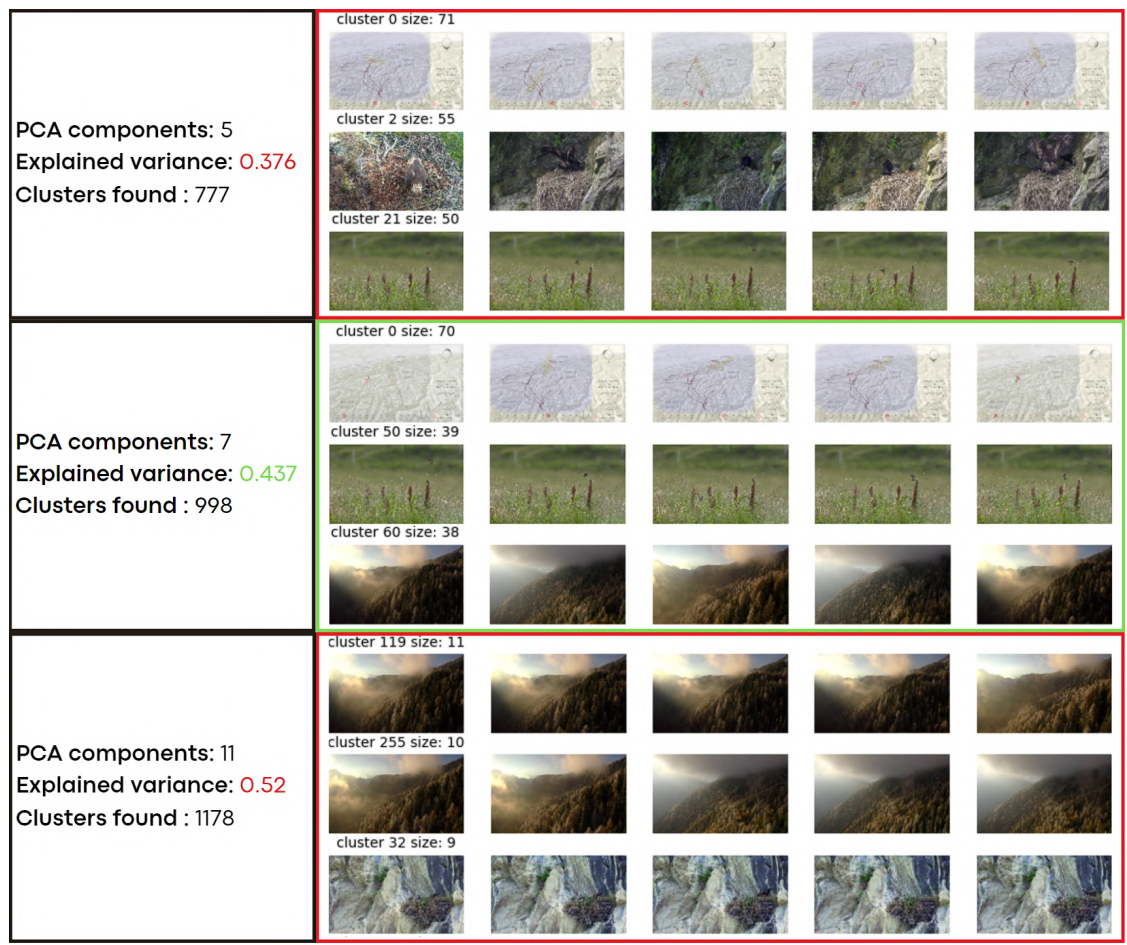

Figure 4.3: The provided image displays the outcomes of clustering analysis using different numbers of components in PCA. It illustrates how varying the number of components affects the resulting number and size of clusters, as well as the explained variance. From each output we display the three biggest clusters obtained, showing five random images from each. As the number of components increases, so does the explained variance, leading to a higher number of clusters, albeit with smaller sizes. However, this can result in similar images being fragmented across multiple clusters, which is not desirable. Conversely, using a lower number of components, such as "5", reduces both the explained variance and the number of clusters, yielding larger clusters. However, this may also lead to dissimilar images being grouped together within a single cluster. The optimal balance appears to be achieved with a value of "7" for the number of components. At this point, there is a satisfactory trade-off resulting in a more coherent clustering solution where similar images are appropriately grouped together without overly fragmenting the dataset.

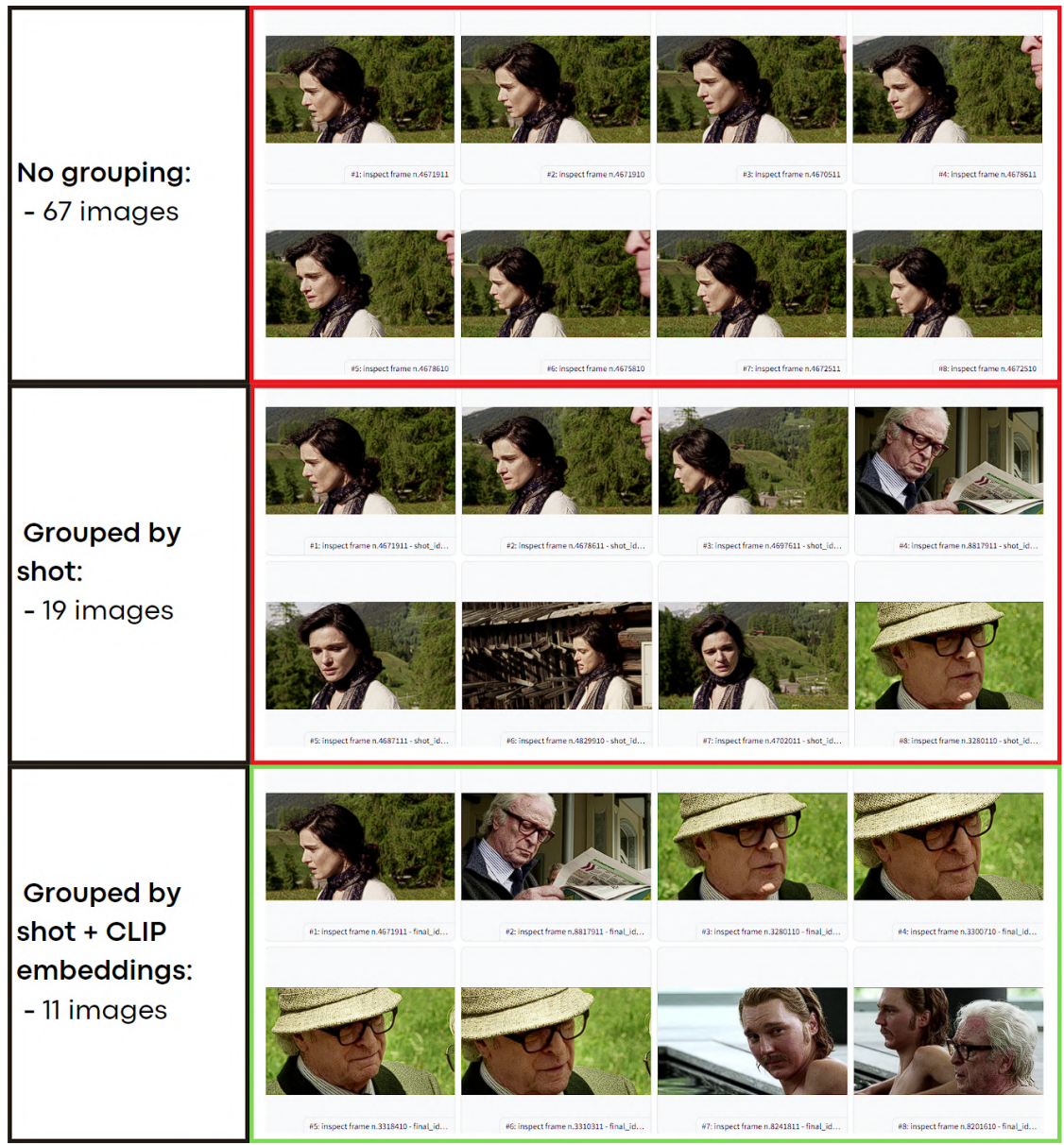

Figure 4.4: This image displays the top 8 results obtained from a query search. Initially, the search yielded 67 images, with the first 8 being nearly identical. Employing shot information to group similar images from the same shot and selecting the finest representation from each group improved the results. However, the most effective enhancement in improving the diversity of the top 8 proposals was achieved by integrating both shot and CLIP clustering information.

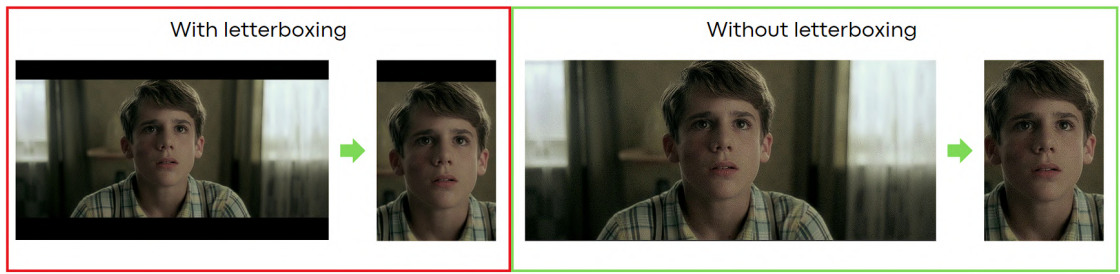

Figure 4.5: When the image is left with letterboxing, the cropped result could contain part of the black pixels. Removing the letterboxing solves this problem.

To effectively remove letterboxing, it is sufficient to crop the frame back to its original aspect ratio, thereby eliminating the black bars. The challenge lies in not knowing the original aspect ratio and the complexity of inferring it. Our initial strategy involved detecting the original aspect ratio by analyzing the first few frames to identify the first top and bottom rows that were not entirely black. This approach, however, proved to be less straightforward for some documentaries that feature short scenes with varying aspect ratios, often shifting to 4:3, and for many movies that commence with frames that are all or predominantly black. Our solution entailed selecting a random sample of 200 frames and calculating the median of the first rows from the top and bottom with at least 30% of pixels being non-black.

### 4.4.2 Results

Following this preparatory step, we apply the cropping method. This implementation was adapted from an existing GitHub repository that had already integrated the cropping method [125], along with a feature for filtering cropping proposals based on face detection, utilizing DSFD [63]. In the subsequent Figure 4.6, we illustrate the disparity between utilizing solely the cropping method [125] and enhanced results, which incorporate filtering based on face detection and the central positioning of faces. The resulting cropped images are then stored in memory (actually, the coordinates of the cropping bounding boxes are stored to minimize memory usage), and each subsequent pipeline step is applied to each image and its corresponding crop as if they were independent images. However, certain steps can be performed just once. For instance, semantic consistency must be evaluated for each image independently, as a cropped image may not include elements present in the full image, thereby affecting the relevance of specific CLIP embeddings. While this principle applies to other steps like aesthetic estimation, shot scale detection is conducted only once per image, as shot scale is determined by the camera's distance to the subject, a factor that remains unchanged by cropping. Similarly, face processing is executed just once for each face, regardless of the number of images or crops containing that face, to conserve processing time.

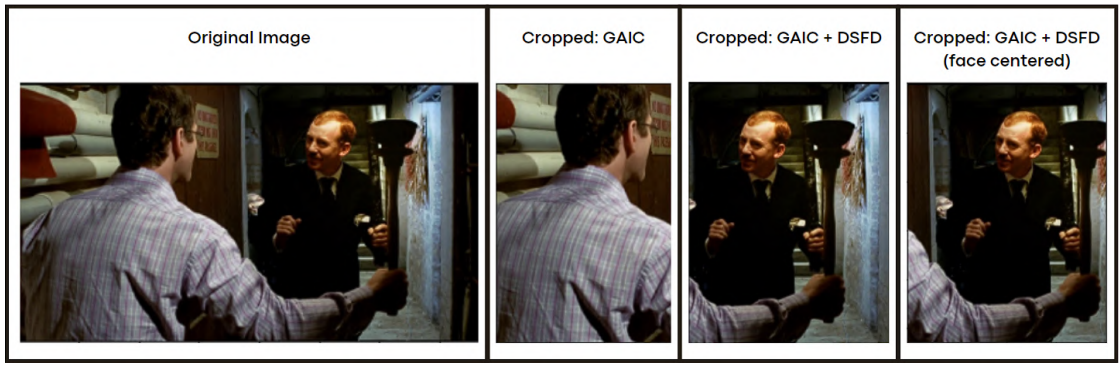

Figure 4.6: The automatic cropping process depicted in the image demonstrates that relying solely on the GAIC method [125] for cropping can potentially omit crucial areas, such as faces. Hence, employing the DSFD face detector [63] is preferable to ensure that crops containing faces are not overlooked. Opting for a crop centered around the face frequently enhances the overall composition for vertical thumbnails, as illustrated in the example on the right.

## 4.5 Aesthetic prediction

In the course of our investigation, we explored various methodologies and implementations to derive an aesthetic score from images. The methodology that was ultimately adopted for integration into our framework was based on the work of Wang et al. [116], utilizing a Python implementation available in the IQA-PyTorch library [15]. This exploration involved conducting a series of experiments to ascertain which method deemed images, extracted from video frames, to possess the highest aesthetic value.

In their corresponding papers, each methodology was compared against their predecessors employing established no-reference image quality assessment (NR-IQA) benchmarks, which usually compare the predicted aesthetic scores of a multitude of images against scores manually assigned by various human judges. However, these benchmarks have limitations. Among these are the subjective nature of quality assessment, which can significantly vary among different observers, and often focus on a limited set of distortions or degradations. Further, the potential for dataset bias exists, which could result in models that fail to generalize well across varied datasets or conditions. Additionally, the metrics employed for evaluation might not fully encapsulate the extent of perceptual quality differences. Finally, an additional consideration to be done for some models is their computational requirements, limiting the applicability of some of them for scenarios requiring real-time processing or in resource-constrained environments.

Given these considerations, our reliance on these benchmarks was only partial. Our primary guide was our personal preference and taste, leading us to favor the CLIP-IQA method, as demonstrated in Figure 4.7. Our experimentation also included calculating the final aesthetic score as an average of two scores derived from different methods. This approach emphasized the importance of varying types of images but did not significantly enhance the final outcomes, while necessitating twice the computational effort, therefore it was discarded.

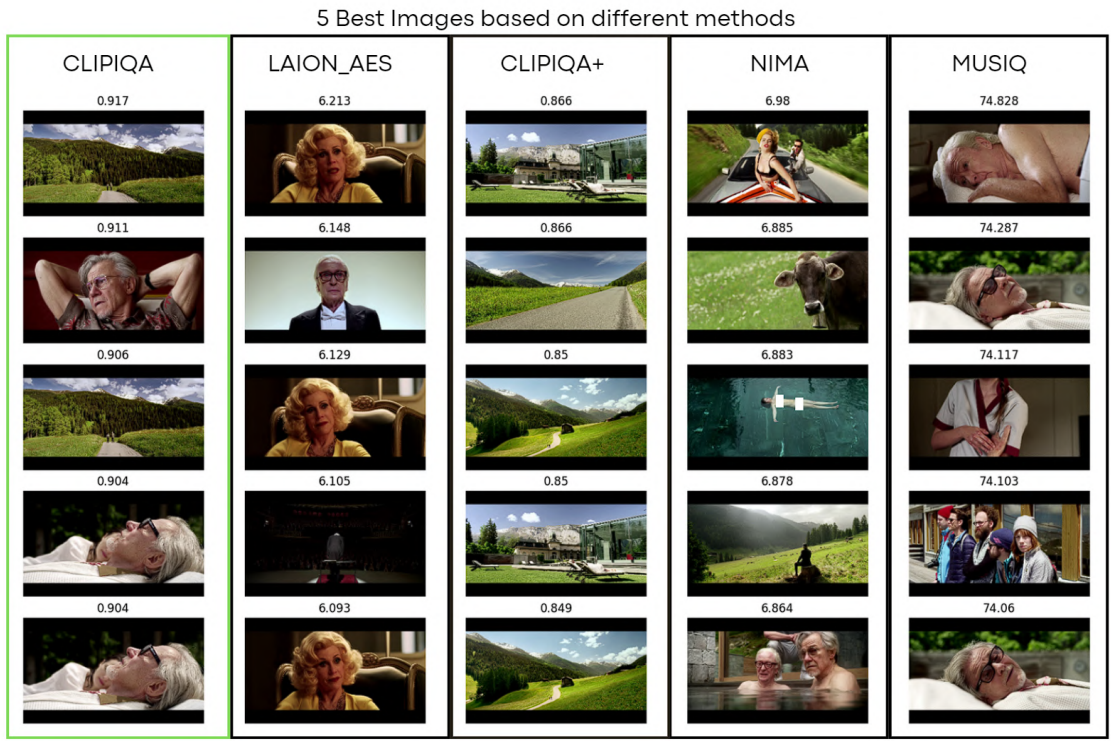

Figure 4.7: In this image, each column showcases five images from the same movie, selected based on their aesthetic appeal as predicted by five distinct methods. The numbers above each image represent the predicted scores, varying in scale according to the respective method. The leftmost column, highlighted in green, aligns closely with our subjective preference.

### 4.5.1 Results

An attempt was made to utilize an IQA method trained on a dataset of face images, with the intention of applying these scores to images featuring faces. However, by integrating the more generalized CLIP-IQA aesthetic scores together with scores related to facial positioning and focus, we achieved comparable or superior results. This approach mitigated the drawbacks of using two distinct scoring distributions, which could potentially introduce bias into the scoring system towards one of the two groups.

The adaptability afforded by the IQA-PyTorch implementation on GitHub facilitates the easy interchange of different methods. Should preferences evolve in the future, adopting a new methodology can be seamlessly accomplished by modifying the configuration file. Despite the diversity of methods, the primary objective is to prioritize visually appealing images over those that are unattractive. This holds true across all methods, where poorly graded images typically exhibit characteristics such as blurriness, distortion, or darkness (refer to Figure 4.8). As long as high-quality images receive high scores, it's not crucial for the top-ranked proposal to always be the best thumbnail option. Other scores play a role in ensuring that the most suitable thumbnail takes precedence over others.

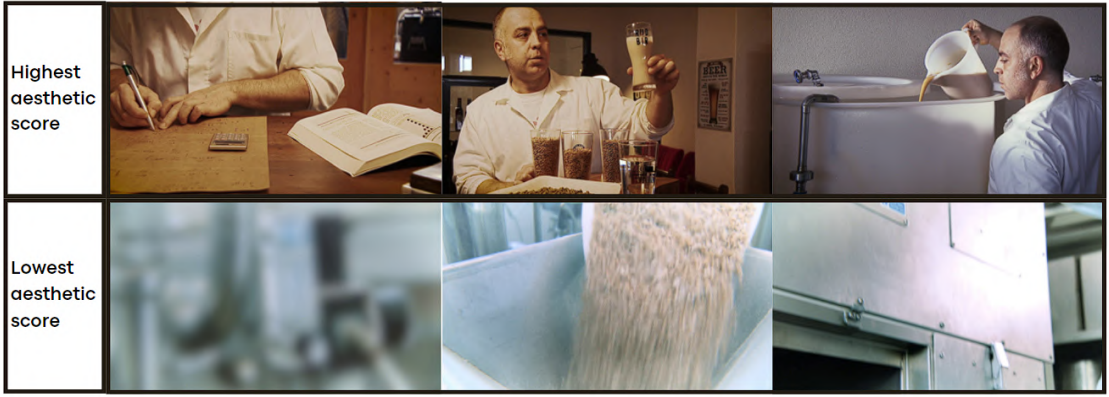

Figure 4.8: The images displayed in the top box exhibit the highest aesthetic scores from the video, while those in the bottom box depict the lowest scores. This demonstrates the model's capability to effectively de-prioritize blurry and visually unappealing images.

## 4.6 Tag extraction

In the preliminary phase of our research, we directed our focus towards utilizing open-source Large Language Models (LLMs), specifically Mistral 7B [48] and Llama 2 [110]. These models were accessed through the Hugging Face library, a renowned open-source platform offering cutting-edge natural language processing (NLP) models and tools for both researchers and developers. Given the necessity for these models to operate on consumer-grade hardware, such as laptops equipped with GPUs, we opted for the smaller variant of Llama 2, which comprises approximately 7 billion parameters, mirroring the scale of Mistral's model. To facilitate the local operation of these models on a GPU with 8GB of VRAM, we employed 4-bit quantization.

### 4.6.1 Quantization

Quantization, within the realm of neural networks like LLMs, entails the conversion of model weights and activations from floating-point representations to integers of lower precision. Our application of 4-bit quantization meant that each weight or activation was encoded using merely 4 bits, significantly reducing the memory footprint relative to the conventional 32-bit floating-point format. However, this reduction in precision can potentially degrade model performance, necessitating the adoption of Quantization-Aware Training (QAT) techniques. These techniques aim to minimize performance losses attributed to quantization by adapting the training process accordingly.

### 4.6.2 Comparison of LLama 2 7B and Mistral 7B

Our comparative experiments revealed that Llama 2 7B outperformed Mistral 7B, notwithstanding Mistral's reported superiority over Llama 2 13B in benchmark assessments. We hypothesize that this discrepancy in performance could be attributed to task-specific factors or the quantization process, with a possibility that Llama 2 benefits from more effective QAT methodologies.

A notable limitation of Mistral 7B, as observed in our experiments, was its inconsistent output formatting. Our methodology necessitated a specific format for keyword extraction, characterized by the segregation of each keyword with commas and encapsulation within square brackets of the whole set. While Llama 2 consistently adhered to this format, Mistral 7B exhibited variability, complicating the accurate parsing of output text into discrete keywords. This inconsistency was further compounded by the model's propensity for "hallucinations", a term denoting the generation of misleading or entirely fabricated content, unanchored in the provided input context.

In our experimental observations, we encountered instances where the proposed keywords either did not originate from the provided description or were derived from a previous description included in the few-shot examples prompt. Additionally, there were inaccuracies in the association between different sentence segments, leading to incorrect keyword identification.

| Description |
|---|
| Fred, a retired English composer and conductor on the threshold of eighty, and Mick, an elderly director still active, spend a spring holiday together in an elegant hotel at the foot of the Swiss Alps. For the two old friends it will be an opportunity to reflect on their future, which is quickly running out, but also to look with curiosity and tenderness at the confused existence of their children and the lives of the other hotel guests, who seem to be able to have of a time that is not given to them. While Mick struggles to complete the screenplay for what he thinks will be his last and most significant film, Fred has no intention of resuming his musical career. However, someone wants at all costs to see him conduct one last time and perhaps the events of this holiday will awaken in him the idea of being able to compose again. |
| Who didn't dream, as a child, of having a farm with lots of animals? A dream that was almost unattainable when the family did not own an agricultural estate, but which Michelle Moschella achieved through tenacity and audacity. |
| Majestic, capable of flying so high that you can imagine it dissolving in the sky, it is no wonder that it is considered the queen of the sky. The golden eagle has always been one of the most fascinating and even adored birds, so much so that Aristotle even considered it divine. Today the beliefs are very different, but her ability to hypnotize has remained intact and her magic still captures our gaze. Over the years, golden eagles have fought for the survival of their species, but fortunately now their numbers have recovered, also thanks to the incessant monitoring work of the Ticino cantonal hunting and fishing office. The documentary offers an intense look - as penetrating as the keen sight of the eagle and as strong as the grip of its claws - on the reality of our canton, rich in biodiversity and nestled between the majestic peaks of the Alps, highlighting how a thin thread inextricably binds the territory, the human being and the eagles in a single destiny. |

Rows A, B, C correspond to the three descriptions above.

| | Llama 2 7B | Mistral 7B | KeyBert |
|---|---|---|---|
| A | 'youth', ' retired composer and conductor', ' elderly director', ' spring holiday', ' elegant hotel', 'Swiss Alps', ' old friends', ' future', ' children', 'hotel guests', ' time', ' screenplay', ' last and most significant film', ' musical career' | 'Fred', ' retired English conductor', ' elderly director', ' spring holiday', ' elegant hotel', ' English Alps', ' spring holiday', 'confused existence', ' curious', 'tenderness', ' screenplay', ' last film', ' future', ' children's lives', 'musical career' | 'composer conductor', 'english composer','complete screenplay', 'mick struggles', 'holiday elegant', 'mick elderly', 'lives hotel', 'resuming musical', 'fred intention', 'fred retired' |
| B | 'farm', ' animals', ' agricultural estate', ' tenacity', ' audacity' | 'dream', 'farm', 'animals', ' audacity', 'tenacity' | 'animals', 'family did', 'agricultural estate','moschella achieved', 'dream unattainable' |
| C | 'golden eagle', ' queen of the sky', ' fascinating', ' adored birds', ' aristotle', ' divine', ' survival', ' species', ' monitoring', ' Ticino cantonal hunting and fishing office', ' biodiversity', ' Alps', ' destiny' | 'golden eagle', ' species', ' human being', ' inseparable', ' Alps', ' territory', ' nestled between', ' binding', ' thin thread', ' inextricably', ' destiny' | 'majestic peaks','years golden'.'eagles single'.'cantonal hunting'.'birds aristotle' |

● = good reference    ● = wrong reference    ● = hallucination

Figure 4.9: Comparison of keywords extracted using different methods: LLama 2 7B [110], Mistral 7B [48], and KeyBert [36]. LLMs demonstrate superior semantic understanding and flexibility. Notably, Llama 2 stands out for its absence of hallucination and accurate keyword composition, a quality lacking in Mistral 7B. The distinctiveness of these methods is highlighted through three different descriptions and extracted keywords, which are visually differentiated using colored boxes.

### 4.6.3 GPT 3.5

Given the results we obtained, our initial preference leaned towards Llama 2 7B. However, as the development of our tool progressed, Play Suisse opted for the utilization of OpenAI's GPT-3.5 model [12], accessed remotely via their proprietary API. This model, with its expansive 175B parameters, significantly surpasses the performance of prior methodologies.

A comparative analysis, illustrated in Figure 4.9, showcases the performance disparities between the aforementioned models and our preliminary experiments with KeyBert [36]. As indicated in the methodology section, KeyBert's rigidity and limited capabilities starkly contrast with the flexibility and power of larger language models (LLMs), demonstrating clear deficiencies in output quality.

#### Prompt engineering

To maximize the capabilities of the GPT-3.5 model, we delved into advanced few-shot prompting techniques, which were unfeasible with the constrained context windows of alternative models. In the realm of natural language processing (NLP) and machine learning, the term "context window" denotes the scope of contiguous words or tokens considered by a model during text analysis or generation. The context window's extent is pivotal, as it defines the volume of ancillary information the model can incorporate for comprehension or prediction of the current word or token. While both Llama 2 7B and Mistral 7B models are equipped with a context window of 4096 tokens, the GPT-3.5 model recently received an enhancement to support a context window of 16,385 tokens. This expansion enables the application of lengthier and more intricate prompts.

Our initial approach with open-source models entailed a few-shot prompt approximately 2k tokens in length for the fixed part, inclusive of examples, augmented by a section for the description of the video from which keywords were to be extracted. The prompt began with an elucidation of the task, instructing the model to anticipate a document containing a video's title and summary, from which it should distill keywords. The directive explicitly excluded personal nouns, such as names and cities, in favor of conceptual keywords, which could comprise single or multiple words. The instruction emphasized the importance of returning only the keywords, omitting any extraneous commentary. Following this, five examples of descriptions and their corresponding extracted keywords were provided.

Leveraging the capabilities of GPT-3.5, we expanded the prompt to accommodate up to 4000 tokens. Utilizing the LLM's API, we were able to guide the model in understanding the purpose behind the provided prompts as a separate prompt, labeled as "Role" prompt. We explained that the extracted keywords were intended as query terms to retrieve images using Clip embeddings. We established guidelines regarding the number of keywords to return and provided examples of what to include and what to avoid in a comprehensive list of requirements. Additionally, we outlined the structure of the prompt, emphasizing a few-shot approach through a series of illustrative examples.

In the actual prompt, we provided the same examples as for the open source LLMs' prompt. In addition, for each example we elucidated the rationale behind the chosen keywords, clarifying their significance while illustrating how they could effectively represent the content, and explaining why other words were not good options. Furthermore, we encouraged creativity, allowing for the condensation of entire sentences into succinct keywords. Additionally, we encouraged the invention of keywords that, while not explicitly mentioned in the summary, could effectively capture its essence and serve as an alternative title. Figure 4.10 shows a comparison of results obtained using the two different prompts.

## 4.7   Semantic consistency

In the methodology section of our research, we primarily focused on a comparative analysis between the CLIP [86] and BLIP 2 [65] methodologies. As illustrated in the comparative results presented in Figure 4.11, our findings indicate that CLIP exhibits a superior capability in accurately capturing the essence of images and assigning them to the most relevant keywords. This performance advantage was observed despite experimenting with both Image-Text Matching (ITM) and cosine similarity 3.4 scores within the BLIP 2 framework, which showed negligible differences in outcome. Notably, CLIP demonstrated a remarkable proficiency in recognizing textual content within images and understanding abstract concepts more effectively than BLIP 2. Additionally, our time efficiency measurements revealed that CLIP processes images at a faster rate, taking 7.1 seconds to compute the score between an image and text pair, compared to BLIP 2's processing time of 20.5 seconds per image, which further increases to 22.1 seconds when utilizing the attention score ITM method.

To incorporate CLIP into our research pipeline, we utilized the official, open-source Python library provided by OpenAI. This library offers comprehensive support for all the main models, although our experimentation did not extend to models employing the ResNet backbone. Instead, our focus was on the base model, ViT-B/32, which employs the Visual Transformer architecture for enhanced performance. Subsequently, we also used the most advanced model, ViT-L/14@336px, which is designed for processing high-resolution images and represents the pinnacle of performance among the available models. Our experimental results were highly satisfactory with the base models, but we ultimately chose to implement the most advanced model due to its superior performance capabilities and compatibility with our local computing resources.

It is important to acknowledge the existence of more recent versions of CLIP, such as OpenClip [19] and CoCa models [122]. These iterations offer improved performance on various benchmarks, attributed to the utilization of larger models and more extensive training datasets. Despite these advancements, our decision to exclude them from our testing was based on the adequacy of our current results in meeting the established needs and criteria for our research.

Figure 4.10: Comparison of keywords extracted using different prompts with OpenAI's GPT 3.5 [12]. While the basic prompt is the same used with previously tested LLMs, the advanced prompt offered enhanced guidance, furnishing additional explanations, hints, and examples for the desired output. Notably, the objective was to condense the keyword count to fewer than ten, excluding personal nouns and details. While the basic prompt outlined this requirement, the advanced prompt provided supplementary insights, evidently facilitating the process. Moreover, the advanced prompt encouraged the inclusion of more imaginative and conceptually diverse keywords. For instance, the prompt "Swiss Alps hotel" effectively amalgamates the concepts of "Swiss Alps" and "hotel," while "magic of eagles" incorporates "magic of," a phrase not explicitly present in the provided description.

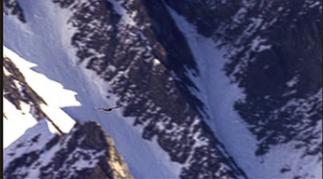

Figure 4.11: The image showcases the frame with the highest matching score between the given keyword and all the extracted frames. In the comparison, the first column employs BLIP-2's [65] embeddings along with an ITM attention-based score for computation. Meanwhile, the second column utilizes cosine similarity (ITC) on BLIP-2 embeddings, and the third column employs cosine similarity on CLIP [86] embeddings. Notably, CLIP demonstrates superior results, presenting images that are more closely related to the provided keyword.

## 4.8   Face detection

In the preliminary stages of our investigation into face detection, we employed RetinaFace [26], as delineated in the Methodology Section 3.10. For this technique, which boasts numerous open-source versions, our choice was an implementation that capitalized on GPU capabilities to expedite image processing. This particular implementation allowed for batch processing of images, achieving rates exceeding 100 images per second with larger batch sizes. However, we encountered several issues in our outcomes. Utilizing lower resolution images through resizing occasionally resulted in the non-recognition of smaller faces. Conversely, inputting images at full resolution often led to a proliferation of false positives, particularly with the erroneous detection of small faces scattered throughout the image, likely due to the misinterpretation of pixel groupings. Additionally, there were instances of multiple detections for a single face, evidenced by numerous bounding boxes demarcating the same facial area. We did not delve into whether these shortcomings stemmed from an inadequate implementation of the algorithm or from inherent limitations within the method itself, despite benchmarks indicating superior performance which was not reflected in our experience.

In contrast, we opted for an open-source implementation of the DSFD method [63]. DSFD demonstrated a superior capability in detecting faces of varying sizes across different images, including challenging scenarios involving occluded faces or those captured from unconventional angles, situations where RetinaFace faltered. Furthermore, the issue of false positives noted with RetinaFace was markedly improved with DSFD. For instance, in an experimental application on a video of 1h and 22mn, RetinaFace detected 2895 faces across 3752 frames, whereas DSFD identified 6931 faces, showcasing its efficacy. Consequently, DSFD was integrated into the final processing pipeline.

## 4.9   Landmark detection and EAR

Our initial implementation utilized the Practical Facial Landmark Detector (PFLD) [38], which encountered significant limitations, notably in its inability to accurately predict landmark positions in challenging scenarios such as profile views or even in straightforward cases involving large, frontally oriented faces. This imprecision adversely affected the performance of the Eye Aspect Ratio (EAR) algorithm, leading to incorrect determinations of eye closure. Instances of these inaccuracies are illustrated in Figure 4.12. Consequently, this step was initially excluded from our processing pipeline due to the high rate of false positives, which resulted in the unwarranted exclusion of viable frames, under the presumption of closed eyes. This was deemed acceptable as the final selection of thumbnail images, typically devoid of closed eyes, suggested a latent aesthetic scoring bias against such frames. Nevertheless, the final GUI tool's capacity to display similar images allowed for rapid manual correction by locating an alternate frame with open eyes when necessary.

On the other hand, in pursuit of enhancing our pipeline, we finally incorporated a more advanced facial landmark detection model, SPIGA [83], which demonstrated superior performance with significantly fewer errors. The SPIGA model delineates nine points per eye—eight along the contour and one at the pupil, compared to the six points identified by its predecessor. This augmentation facilitates a more refined computation of EAR, averaging three vertical lengths rather than two, thereby yielding a more precise measurement of eye closure with an established threshold EAR value of 0.2 to signify closed eyes.

Moreover, the inclusion of pupil position predictions, in conjunction with overall improved landmark detection, has enhanced the accuracy of face position scoring. This metric benefits from the refined calculation of the face center, defined as the midpoint on the line connecting the two pupils, located at the upper portion of the nose.

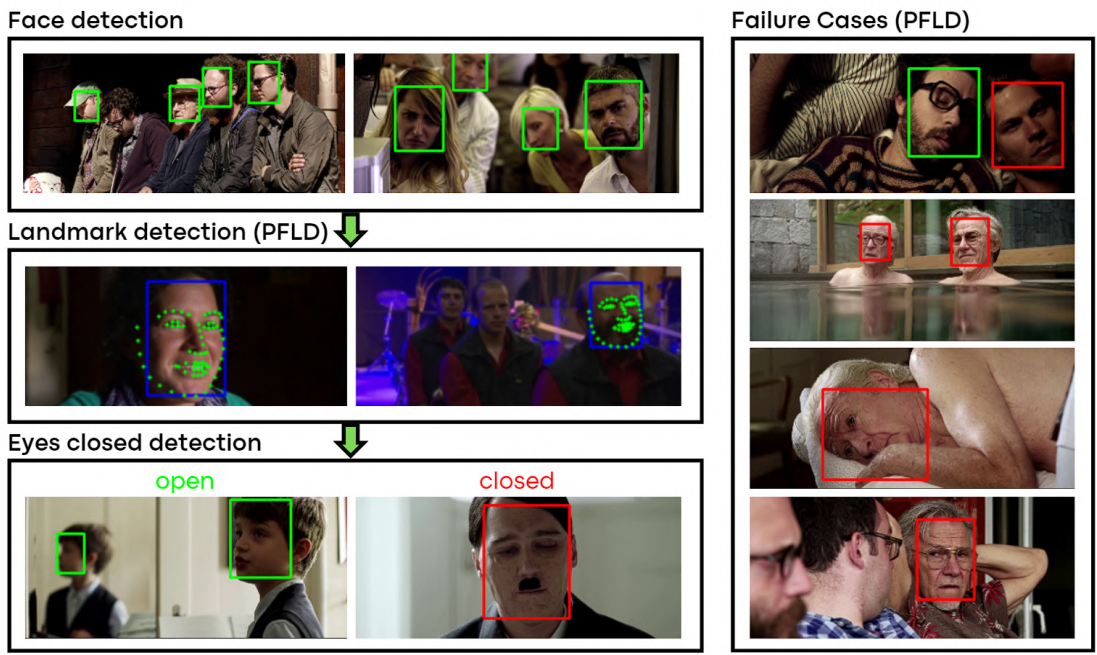

Figure 4.12: The illustration depicts the process for achieving closed eyes recognition through a combination of face detection, landmark prediction, and the use of Eye Aspect Ratio (EAR). On the right side, the diagram highlights the drawbacks of the PFLD method [38], which due to inaccurate face landmark prediction can lead to erroneous closed eyes detection (red box means the eyes are predicted to be closed, green means open). Such limitations have been solved by substituting PFLD with SPIGA [83].

## 4.10 Face identification

In the methodology section, we delineate the chosen approach for facial identification, which involves the computation of facial embeddings followed by unsupervised clustering. The initial step entails the augmentation of face detection bounding boxes by a factor of 1.2 on all sides. This adjustment aims to incorporate additional facial details, such as parts of the hair, which are often excluded in the tightly focused bounding boxes that typically encompass only the face, mouth, and nose areas. Such an enlargement strategy was important in enhancing the distinctiveness of each face by incorporating more detailed features into the image.

Subsequently, a filtration process was implemented to exclude diminutive faces. The rationale behind this decision stemmed from the observation that the model was optimized for faces resized to dimensions of 160x160 pixels. Faces smaller than this threshold would necessitate upscaling, leading to embeddings of compromised quality due to the insufficiency of detail. It is noteworthy that a face measuring 160x160 pixels occupies merely 1.2% of the total area in a 1920x1080 (full HD) resolution image, rendering the presence of such small faces relatively rare. Conversely, thumbnails typically feature faces that account for at least 10% of the image area. To strike a balance between excluding excessively small faces and retaining those of adequate size for thumbnail representation, we established a threshold of 5% area coverage. Faces falling below this threshold were classified as noise and excluded from the clustering process to enhance the accuracy of the outcomes by mitigating irrelevant noise.

Further explorations were conducted with two alternative clustering strategies: one treating each face as an individual data point and another considering each appearance of a face—across original, vertical, and horizontal crops—as separate data points. The latter method, although initially an unintended deviation, was observed to potentially improve performance in certain videos. This improvement, albeit not straightforwardly explicable, could be attributed to the increased density of data points in already dense regions, thereby facilitating the clustering algorithm's ability to discern and accentuate distinctions between larger and smaller clusters. Following a series of experimental analyses, the decision was made to adopt this strategy.

### 4.10.1 Hyperparameters finetuning

In the methodology section, we detail our approach of employing Principal Component Analysis (PCA) for dimensionality reduction prior to clustering with DBSCAN. Initially, the number of PCA components was fixed; however, subsequent analysis revealed significant variability in outcomes contingent on the dataset characteristics. Specifically, videos of shorter duration or those featuring a limited number of faces exhibited markedly different behavior in comparison to longer films. This observation led to the realization that maintaining a constant number of components was less effective than ensuring a minimum explained variance threshold. Empirical evaluation across diverse video datasets indicated that an explained variance of 0.74 optimally balances dimensionality reduction and data representation integrity.

On the other hand, in our exploration of optimizing the selection of Principal Component Analysis (PCA) component numbers for improved clustering outcomes, it became evident that a fixed value for explained variance was also insufficient for consistently achieving optimal cluster configurations. Despite initial findings suggesting a viable number of PCA components, further experimentation revealed that certain videos exhibited suboptimal clustering, which could be ameliorated through manual adjustments within a ±10 range of the initially determined component count. To systematize this adjustment process, we implemented a grid search across this 20-value spectrum to identify the component count that yields the most effective clustering result. This necessitated the development of a scoring mechanism for each clustering outcome to facilitate the identification of the superior arrangement.

Our methodology for scoring involved the computation of the minimum cosine similarity 3.4 between pairs of embeddings within the same cluster, a metric that inherently spans from -1 to 1, with 1 denoting identical vectors and -1 indicating completely opposite vectors. Clusters amalgamating multiple faces from different individuals typically registered cosine similarities approaching zero or dipping into negative values. Consequently, our initial scoring strategy comprised the summation of the product of each cluster's size (i.e., the number of faces it contained) and its minimum cosine similarity value. This approach ensured that clusters exhibiting negative values detrimentally impacted the overall score, thereby penalizing poor clustering.

However, this scoring mechanism inadvertently favored clustering configurations resulting from higher PCA's number of component selections, which tend to produce a greater number of clusters characterized by high similarity among included faces, often originating from identical scenes. Such outcomes, while achieving high internal cluster similarity, are not desirable due to the excessive fragmentation and lack of diversity among the faces within the clusters. Moreover, higher PCA's number of component correlated with an increased likelihood of embeddings being classified as noise. To mitigate this bias and promote a balance between the formation of coherent clusters and the minimization of overly granular clustering, we revised our scoring formula to deduct the count of faces identified as noise from the total score. This adjustment aims to prioritize clustering solutions that strike an optimal balance between cluster quality and size, minimizing the size of the noise cluster, thereby enhancing the overall utility of the clustering process. The final formula is defined as follows:

$$S = \sum_{i=1}^{N} (|C_i| \cdot \min_{j,k \in C_i} \cos(\vec{v}_j, \vec{v}_k)) - N_{\text{noise}} \tag{4.1}$$

Where $S$ denotes the final score of the clustering configuration, $N$ is the total number of clusters, and $|C_i|$ represents the size of cluster $i$, i.e., the number of faces within the cluster. $\min_{j,k \in C_i} \cos(\vec{v}_j, \vec{v}_k)$ calculates the minimum cosine similarity between all pairs of embeddings $\vec{v}_j$ and $\vec{v}_k$ within the same cluster $C_i$, while $N_{\text{noise}}$ is the number of embeddings flagged as noise.

Following dimensionality reduction, we empirically determined the optimal DBSCAN parameters to be an epsilon value of 0.5 and a minimum points threshold of 50, utilizing Euclidean distance as the metric. This configuration effectively categorizes groups with fewer than 50 faces as noise.

Figure 4.13 presents examples of the outcomes achieved with this technique. Despite sometimes the clustering output contains some imperfect clusters, the results are deemed satisfactory for our objectives. The primary aim is to swiftly identify the principal faces within a video for thumbnail generation, allowing for the exclusion of minor characters. Even in cases where major characters are amalgamated within a single cluster, this approach facilitates rapid image retrieval for these individuals, albeit without precise segregation. Occasionally, a significant character may be represented in both a large cluster and a smaller, separate cluster comprising images from a challenging angle, perceived as distinct entities. Typically, these smaller clusters originate from a singular scene or shot, rendering their exclusion from the final character proposal inconsequential due to their limited scope.

The most challenging scenario arises when all faces are amalgamated into a single cluster, complicating the differentiation of primary and secondary characters. To mitigate this, we have refined our hyperparameters to minimize the occurrence of such cases. Nonetheless, it is impractical to entirely eliminate the possibility of encountering datasets that defy the chosen parameters. For these outlier instances, we have incorporated a feature within the GUI tool that allows for rapid manual adjustment of clustering parameters. This inclusion ensures that face identification remains feasible, albeit not fully automated, representing a rare limitation of our method. Furthermore, the inclusion of the grid search approach as strongly reduced these failure cases, making them very uncommon.

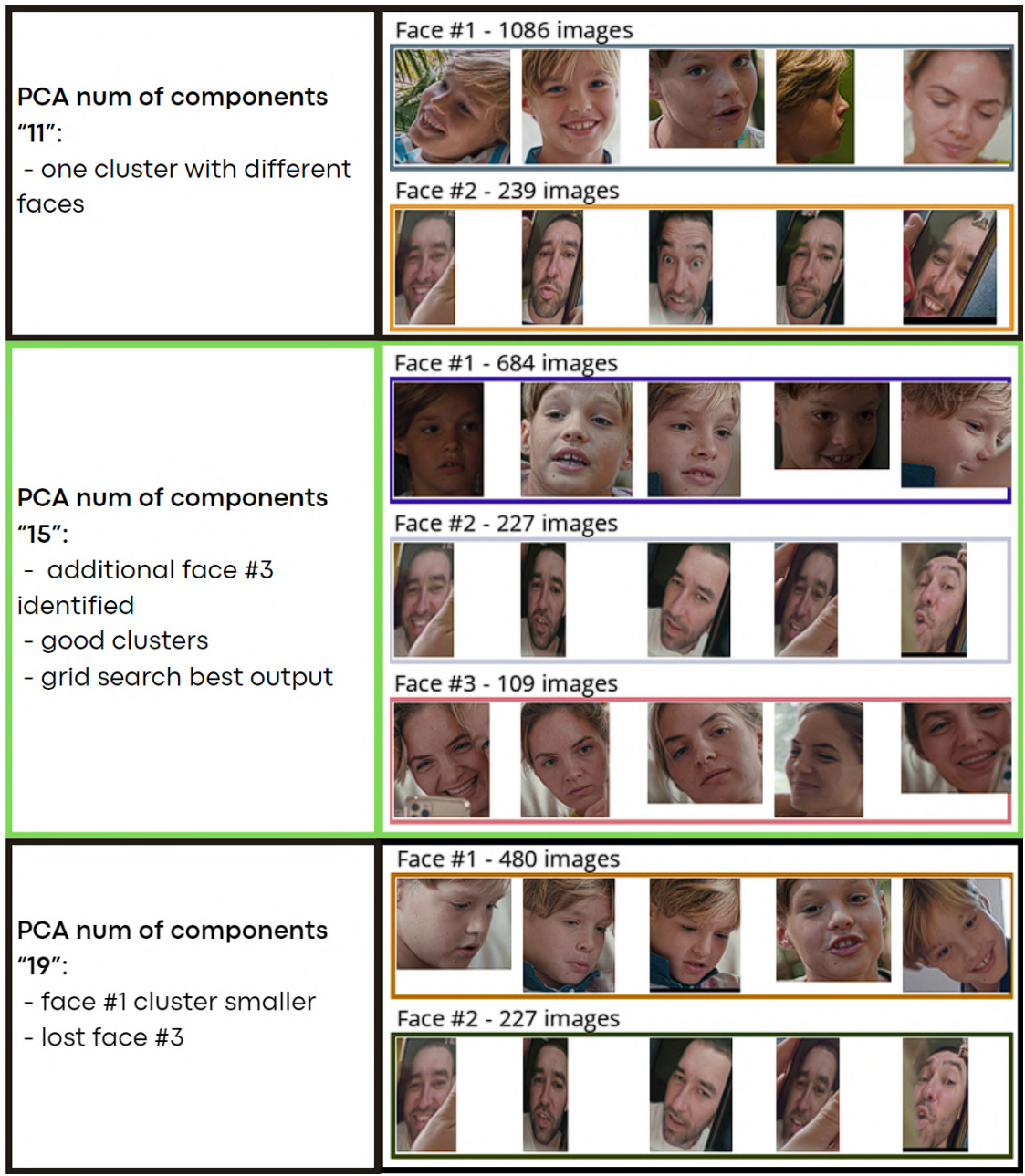

Figure 4.13: This image displays the results of face identification using Principal Component Analysis (PCA) with different number of components' values on a 14-minute video with few faces present. The implemented grid search algorithm determined that utilizing 15 components yielded the optimal output. Lower component values tend to amalgamate distinct faces into the same cluster, whereas higher values identify more images as noise, potentially removing legitimate faces and diminishing the prominence of the primary face cluster.

## 4.11    Time and memory requirements

The objective of this project is to design and implement a processing pipeline capable of automatically selecting a set of candidate thumbnails for new videos added to the Play Suisse database. This automated selection process facilitates the subsequent choice by professional thumbnail designers through a graphical user interface (GUI) tool. Given the temporal gap between the addition of a video to the database and its availability in the user catalogue, the thumbnail generation process benefits from a flexible timeline. Consequently, the project's methodology and approach were primarily influenced by considerations other than time constraints.

Utilizing Microsoft Azure's computing platform afforded the flexibility to scale hardware resources according to the demands of the processing pipeline, thereby mitigating concerns related to memory limitations. This adaptability ensured that memory constraints did not significantly influence the project's approach.

Nevertheless, the utilization of Azure's more powerful computational resources incurs higher costs, billed on an hourly basis. This economic factor necessitated the implementation of time and memory optimizations to mitigate expenses. Despite these optimizations, the primary focus remained on maintaining the integrity of the pipeline's performance, under the premise that the benefits derived from automating the thumbnail selection process significantly outweigh the associated costs. Detailed explanations of strategies used to optimize memory and time are explained in the Appendix.

## 4.12    Adjustements to the pipeline

In the methodology and introduction sections of our research, an essential component of our approach to determining the optimal pipeline was the engagement with the Play Suisse team of professional thumbnail designers. Through discussions regarding potential alternatives and focal areas, we were able to identify the most suitable pipeline to meet their specific requirements. These requirements became increasingly clear over time, particularly after the initial experimental outcomes were analyzed, allowing us to discern what aspects were truly beneficial and which required less emphasis. For instance, our preliminary strategy included the implementation of automatic color grading and the application of styles and color palettes based on tags that corresponded semantically. However, feedback from the design team highlighted that a predefined color palette could undermine the unique identity of each video, potentially diminishing the overall aesthetic appeal of the catalog. Consequently, our focus shifted away from this approach towards other elements of the design process.

Another area initially targeted for automation was the placement of titles within thumbnails. However, the design process typically involves selecting an image before creating a corresponding title, rendering the concept of automatic title placement impractical, as the title information would not always be available for input into the pipeline.

Our research also included an extensive review of existing literature to explore various methodologies for thumbnail generation, with an emphasis on creating non-clickbait thumbnails that accurately represent the video content and production quality while maintaining the recognizability of main characters. Feedback from the design team led to the realization that image generation should be reserved for less common scenarios. Instead, prioritizing the extraction and selection of video frames as the primary method for thumbnail creation emerged as the preferred approach. Thus, our efforts concentrated on optimizing the thumbnail extraction and selection process, while the image generation component was assigned a lower priority and subjected to fewer experiments to enhance performance.

The tools incorporated into our framework possess substantial capabilities that can also be exploited manually to achieve highly specific outcomes efficiently, significantly accelerating the process. For example, the segmentation and matting method, capable of processing text prompts to segment desired objects, and Stable Diffusion [92], an open-source text-to-image diffusion model, offer extensive functionalities. While our pipeline automates certain tasks using these tools' APIs, their vast array of features, not fully utilized in our automated process, could be manually leveraged to fulfill additional requirements if necessary.

## 4.13   Scoring system and variety purpose

In the initial phase of our investigation, we conducted a comprehensive analysis of each individual step within the process to evaluate its advantages, disadvantages, limitations, and potential failure scenarios. This approach enabled us to ascertain the relative robustness of each feature, identifying those that could be reliably utilized and others that, due to their inconsistent or less robust nature, might require alternative strategies to ensure optimal outcomes across a broad spectrum of videos. Upon integrating these components into a cohesive operational pipeline, we were left with a wealth of data extracted from the video frames. However, this data, in its raw form, lacked utility until it was effectively leveraged. To address this, we developed a sophisticated graphical user interface (GUI) specifically designed for the visualization, examination, and analysis of the data, distinct from the interface of the final tool.

### 4.13.1   Filtering panel with data visualizations

The primary functionality of this analytical tool was to facilitate the display of data distributions. For quantitative data, such as aesthetic scores, we employed histograms to visualize their distribution, whereas for categorical data, such as emotions, we utilized horizontal stacked bars to illustrate the proportion of each category. In the backend, the data was consolidated into a single dataframe, with various features segmented into separate columns. This setup was enhanced with interactive elements such as sliders and buttons, enabling users to filter data based on specific criteria—for instance, isolating images featuring closed eyes or selecting logos with scores within a specified range (e.g., 0.3 to 0.4). Additionally, we incorporated multi-choice dropdown menus to select desired emotions or facial clusters, offering the capability to apply these filters in combination.

Further refining our analysis, we introduced a feature to adjust the weight assigned to each score, thereby facilitating the isolation or amalgamation of scores to evaluate their impact on the final outcome. For the semantic consistency score, we implemented a multi-selection dropdown menu, allowing the selection of one or more keywords. The final semantic consistency score was calculated as the average of the cosine similarities 3.4 to the chosen words, with the provision to add custom keywords and dynamically recalculate cosine similarity to experiment with the possibilities of CLIP model.

In addition to the aforementioned functionalities, we incorporated interactive input fields within our system to enable the customization of parameters pertinent to the face clustering process, notably the epsilon and minSamples for DBSCAN, the number of components for PCA, and a threshold for the minimum area of faces, below which they would be considered noise. This customization capability was similarly extended to the clustering of images to eliminate redundancies in our dataset. Such an approach allowed for iterative experimentation with these parameters, facilitating the identification and selection of optimal values as elaborated in the respective sections above. Upon specifying these parameters, the initiation of the clustering process is triggered via a designated button, which consequently yields the explained variance from the PCA, alongside the clusters formed, each illustrated with five exemplary images. The five images are selected from the entire cluster to maximize cosine similarity. First, we obtain all the images and their corresponding embeddings within the cluster. Then, we select one embedding as a reference and calculate the cosine similarity with all other embeddings. Subsequently, we sort the images based on these similarities and choose five images that are evenly spaced among the sorted list. This approach provides an initial understanding of the diversity of faces within the cluster. It allows us to quickly assess whether the cluster predominantly contains images of the same face from the same scene (indicating low variety and potentially limited usefulness), multiple different faces (indicating failure), or a diverse set of the same face from different scenes (indicating a good cluster).

Furthermore, we introduced a feature enabling the selection between the maximum and the average values of scores for each face, particularly useful in scenarios where multiple faces are detected within a single image. The system also supports the creation and storage of named presets for these parameters, significantly enhancing the efficiency of testing across multiple videos. Visualization of clip embeddings is facilitated through a 3D point cloud, generated via t-SNE to reduce the dimensionality from 512 to 3 dimensions, with data points color-coded according to the keyword closest in semantic proximity to each image.

Subsequent to the configuration of these parameters, the execution of a "search" function filters the dataset according to specified filters and computes a final score based on predefined weights. The resulting high-ranking images are then displayed in a gallery format, with options to navigate through multiple pages, thus allowing a comprehensive examination from the highest to the lowest ranked images. An optional checkbox facilitates the reverse order display, from the lowest to the highest scores, aiding in the validation of the scoring system's efficacy in prioritizing or deprioritizing images according to the intended criteria.

The development of this system, leveraging the flexibility and rapid deployment capabilities of Jupyter Notebook's widgets, underscores its critical role in establishing the most effective presets for weights and filters. These presets were subsequently applied in an automated fashion within the final pipeline, culminating in the presentation of optimal thumbnails through the final GUI tool, as discussed in the Methodology Section 3.12. Despite the utilitarian design of this development tool, devoid of aesthetic embellishments, its practicality was paramount, given its intended use by developers rather than end users. Figures 4.14 exemplify the filters, distribution plots, and resultant outputs.

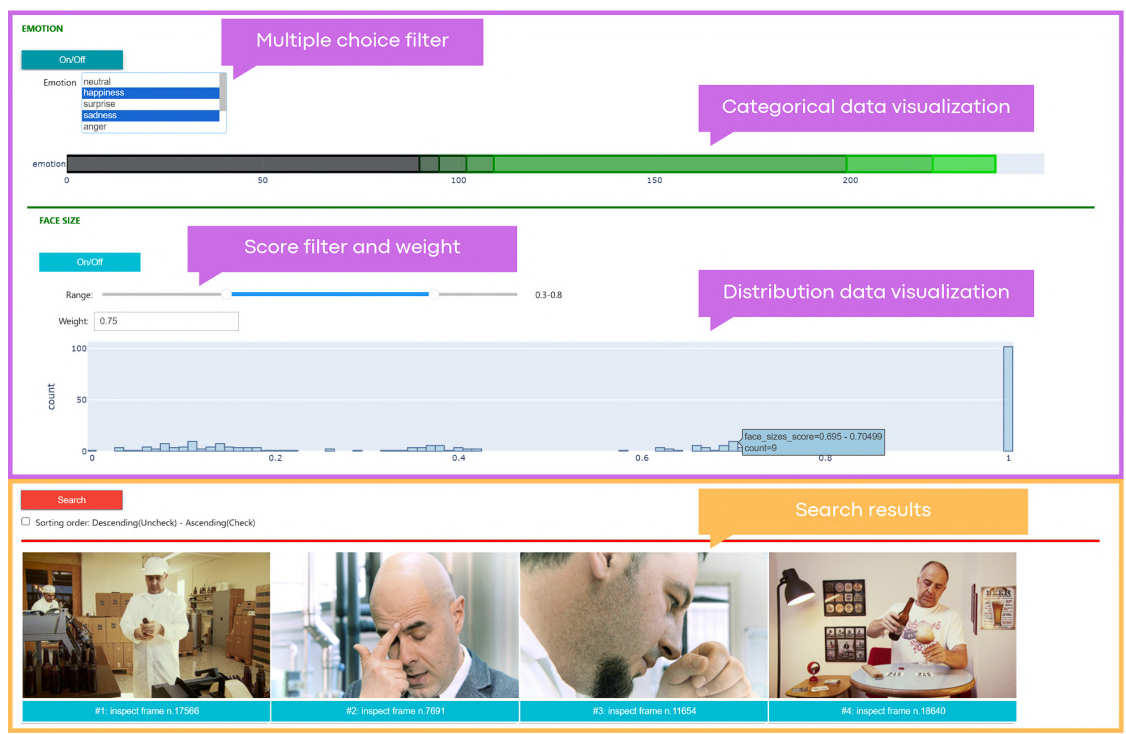

Figure 4.14: This image highlights examples of some of the different components that constitute the filtering panel on the data analysis GUI tool. The speech bubbles explain the purpose of the different parts.

### 4.13.2  Image panel for results visualization and analysis

To enhance the analytical capabilities of the graphical user interface (GUI) tool and to provide a deeper understanding of the criteria for the high-ranking of specific frames, additional functionalities were integrated. Upon displaying the search results, each image is designed to be interactive; selecting an image triggers the display of various plots and figures, elucidating the factors influencing its ranking based on the predefined parameters. Initially, a triptych of images is presented: the first delineates the original image, overlaying boxes on detected faces—colored red for faces with closed eyes and green otherwise. The second image incorporates the saliency prediction heatmap alongside the outlined face, offering a visual interpretation of the on-face focus score by highlighting the saliency concentration within the face box, with the score explicitly annotated in the upper left corner. The third image serves as a visualisation of the logo score, with the score mentioned in the upper left corner and the image depicting a probability map, from which saliency heatmap and face areas have been excluded as outlined in the Methodology Section 3.11.

Subsequent to these images, the analysis extends to report on the clustering of faces within the image, if applicable, accompanied by exemplars from the same cluster to validate the accuracy of the clustering process. Furthermore, the tool exhibits images originating from the identical shot, alongside those from the same scene, and those grouped via CLIP's embeddings. The culmination of this analysis is the presentation of images within the same group, achieved by amalgamating CLIP's clustering with shot information, thereby facilitating an understanding of which redundancy avoidance technique proved most effective.

Furthermore, the analytical platform encompasses an extensive suite of visualizations, including three pie charts delineating the distribution of face counts, emotional expressions, and shot scales within the curated image set, thereby providing insights into the typology of the selected imagery. Subsequently, a stacked bar graph delineates the original scores, whereas an additional bar chart elucidates the composition of the composite score, both with and without the application of personalized weighting factors. Moreover, for each score, the distribution of all data points is presented, accentuating those corresponding to images that were considered in the output of the search. This approach not only identifies the data points that were excluded but also emphasizes the relative positioning of the chosen image within the spectrum of all scores. This comprehensive visual representation facilitates a nuanced understanding of the factors contributing to the final image ranking, offering a comparative analysis of the selected image against others within each scoring criterion. Such a detailed analytical framework was instrumental in refining the algorithmic pipeline, enabling the precise calibration of weights across different presets and optimizing parameters for the clustering process. An illustrative example of the user interface activated upon image selection is depicted in Figure 4.15.

In summary, these features were designed to critically evaluate failure instances, ascertain the resilience of various features under typical conditions, and identify areas necessitating enhancement. This process enabled a thorough understanding of how to effectively utilize each feature, leading to the development of optimal presets for filtering and weighting. These presets were subsequently integrated into an automated process for the final selection phase.

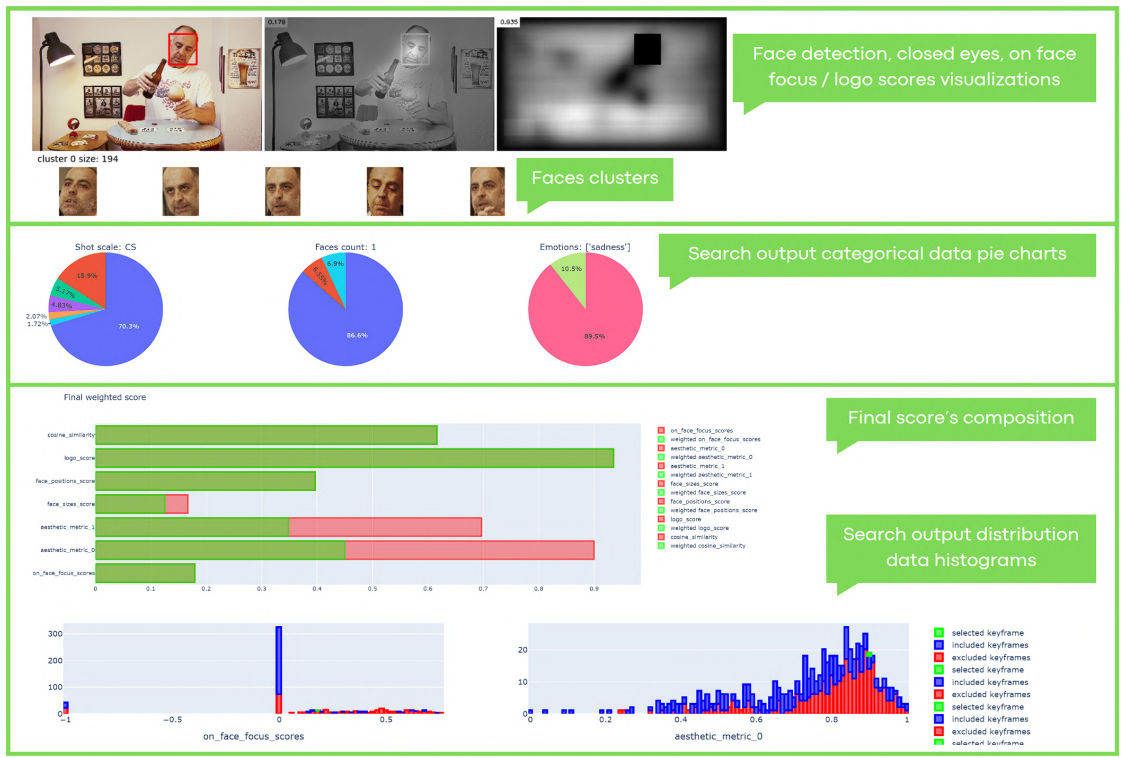

Figure 4.15: This image highlights examples of some of the different components that constitute the image result panel on the data analysis GUI tool. The speech bubbles explain the purpose of the different parts.

### 4.13.3 Quantile Transformation

Finally, the potential application of the Quantile Transformation methodology was thoroughly evaluated. This technique, designed for the modification of feature values, aims to remap these to a specified probability distribution, thereby facilitating a closer approximation of the transformed features to either a Gaussian (normal) distribution or a uniform distribution. The procedure achieves this by initially mapping the original data to a uniform distribution, which is subsequently mapped to the user-specified target distribution. Such transformations are particularly advantageous for algorithms predicated on the assumption of Gaussian-distributed data. Moreover, they contribute to the stabilization of variance and the enhancement of machine learning model performance by ensuring that each score is uniformly contributed to the final assessment. In our specific context, it could be leveraged to stabilize distributions of different scores so that each score would contribute to the final score equally. On the other hand, the utilization of this transformation was deemed unnecessary. A simpler approach, involving the normalization of each score to a 0-1 range, sufficed due to the roughly similar distribution of all scores under consideration. This approach's efficacy and the comparative impact of each score on the ultimate outcome are elucidated in Figure 4.16.

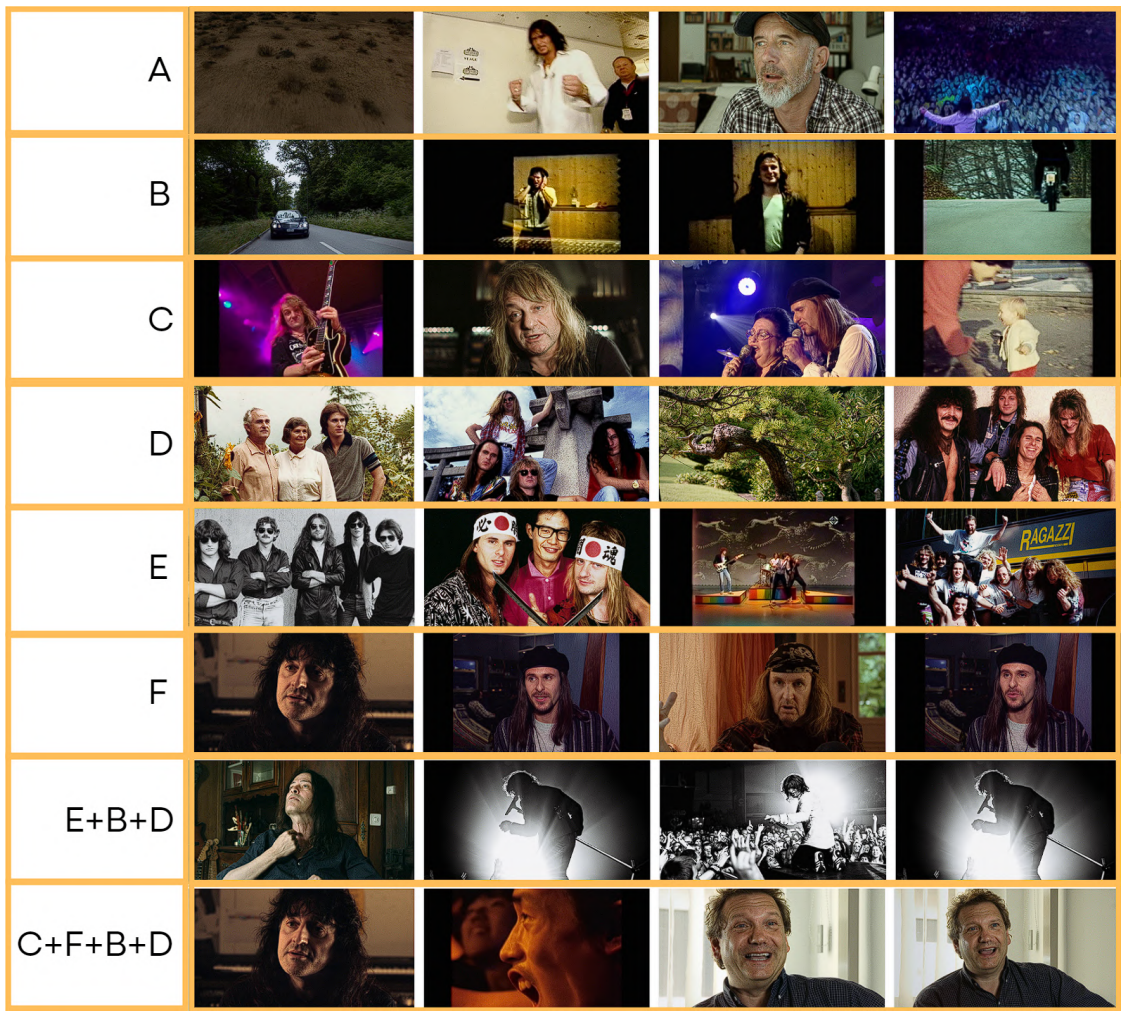

Figure 4.16: This image highlights the prioritization capabilities of the different scores. Each row represents the 4 images with the highest score images using different scores' weights. "A" has all weights set to zero, therefore the images are random. "B" enables only "logo score", "C" only "face position" score, "D" only "aesthetic score", "E" only "semantic consistency score" for keyword "swiss rock band", and "F" only "on face focus score". The two rows at the bottom are a combination of the other scores, identified by the corresponding letters.

## 4.14 Image enhancement

In the development of our image generation framework, as delineated in the Methodology Section 3.13, the initial strategy involved the selection of exceptional portraits of principal characters to serve as the foreground, integrated with picturesque and emblematic long shots for the background. This selection process mirrored the methodology employed for thumbnail selection in the final proposal, utilizing a preset similar to that which was employed for curating optimal portraits of each main character. Additionally, specific keywords were employed to guide the results towards semantic consistency. It is imperative to note that not all images are suitable for this purpose; certain images pose segmentation challenges. For instance, images featuring another individual in the background could inadvertently include that individual as part of the foreground, or images depicting individuals in unconventional positions, such as lying down. Consequently, we endeavored to prioritize images with a clear demarcation between foreground and background, aiming to achieve more consistently high-quality and stable results. In parallel, for background selection, we utilized a preset analogous to that used for aligning images with keywords, with a focus on wide shots to minimize the presence of faces, ensuring the background remains relevant without detracting attention from the foreground. By adhering to semantic consistency with keywords, we ensure that backgrounds are pertinent to the video, as every aspect of the image is crucial for effectively conveying the video's content.

### 4.14.1   Foreground segmentation

For foreground segmentation, our exploration encompassed various methodologies. The initial approach utilized a widely recognized Python library for background removal named Rembg, which is based on the U2-Net [85] architecture pretrained on diverse datasets, including one specifically designed for segmenting individuals from backgrounds. This method inherently produces a segmentation mask, albeit with the limitation of pixelated borders and the absence of a seamless transition between the foreground and background upon background replacement. To ameliorate this, a rudimentary algorithm for alpha matting was incorporated into the library. Despite these enhancements, the outcomes were frequently unsatisfactory, with numerous instances of failure, as illustrated in Figures 4.17.

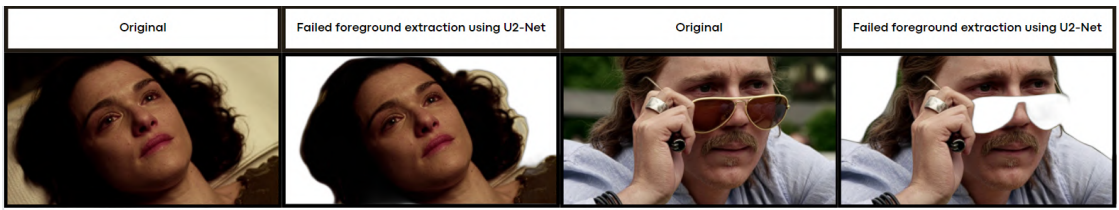

Figure 4.17: This image shows failure cases for foreground segmentation using U2-Net model [85].

In light of these limitations, we opted to adopt a more contemporary and sophisticated method, Matte Anything [121], which demonstrated superior accuracy in both segmentation and matting processes. Importantly, this process accepts a text prompt as input. Through our experimentation with keywords, we discerned that "person" keyword yielded more favorable results for our objectives rather than "foreground" to guide the segmentation process. Additionally, we present examples of results achieved through image harmonization together with failures cases (see Figure 4.18) for which code and pretrained models were made available on a GitHub repository. Originally designed for execution within a GUI tool, we adapted the code to facilitate automatic inference calls within our framework, obviating the need for a GUI server application. This adaptation was executed in a manner akin to our approach with the Matte Anything repository, which similarly provided code exclusively for GUI application execution.

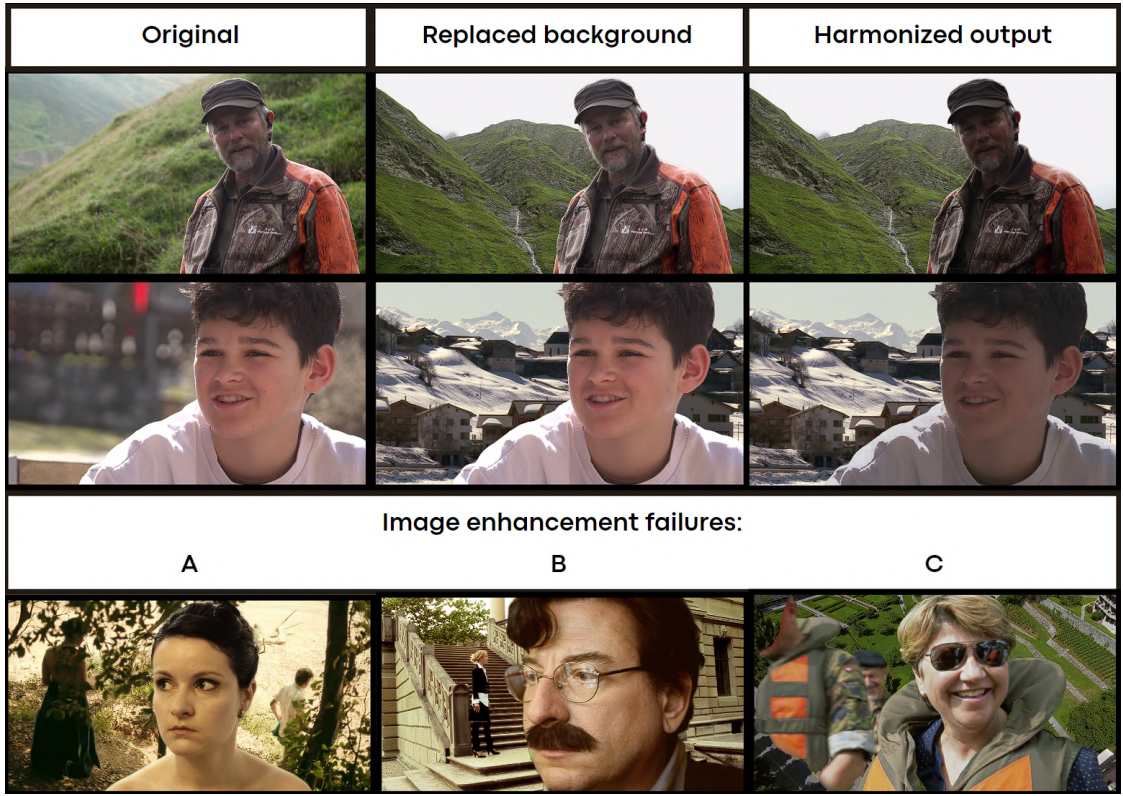

Figure 4.18: The initial two rows of the image exemplify successful applications of our image enhancement technique. Conversely, the final row highlights instances of failure. Each example in this last row suffers from an unsuitable background image selection, indicating a shortfall in our background selection mechanism. Image "B" demonstrates an inability to recognize and correctly apply matting to the transparent sections of glasses. Additionally, image "C" mistakenly includes individuals in the foreground, underscoring a lapse in our foreground selection process designed to exclude images containing multiple faces.

## 4.15   Image generation

In the process of generating backgrounds using stable diffusion, an extensive series of experiments were conducted, the details of which would be overly cumbersome to describe in entirety within this thesis. However, it is pertinent to note that these experiments largely did not meet the high-quality and consistency criteria set forth for our purposes, and were executed manually through advanced GUI tools for text-to-image generation leveraging stable diffusion, such as AUTOMATIC1111 webUI [7] and Kohya SS. Initially, the investigation involved testing various pretrained models of Stable Diffusion, including recent advancements such as Stable Diffusion XL (SDXL) [81] and SDXL Turbo [97], alongside finetuned versions geared towards the generation of hyperrealistic images, and models specifically trained for inpainting tasks. These models exhibited a significant improvement in the quality of photorealistic image generation in comparison to the original models detailed in prior publications.

### 4.15.1 Background generation with ControlNet

Subsequent experiments focused on generating composite images. The methodology initially involved creating new images of principal actors by either replacing only their faces or by finetuning a model to recognize the person as a learned object for generation in various scenes and poses. This approach, however, faced challenges in maintaining recognizability of the characters—critical for main characters in videos—and also posed issues for meeting clickbait criteria, leading to a cessation of further investigation in this direction. The strategy was then shifted towards background replacement, utilizing ControlNet [126] to guide the background generation process based on various inputs derived from a background frame of the video. These inputs included depth maps, edge detection, segmentation maps, and body poses, among others. Despite extensive experimentation, it was observed that strong conditioning with multiple combinations of the mentioned inputs often resulted in the production of artifacts, thereby prompting the decision to use a background image from the video as the only conditioning reference for generating a similar, yet suitably composite, background. For instance, providing an image of a mountain led the model to generate a correspondingly similar mountain scene that blended well with the foreground, enhancing the overall image quality beyond mere background substitution.

### 4.15.2 Background generation with text prompts

Alternatively, backgrounds could also be generated using text prompts. To streamline this process, a set of general prompts were predefined for generating abstract backgrounds, such as a plain red background, which, through the capabilities of diffusion models, would not result in a simple solid color but would adapt to complement the foreground, thus producing a higher quality image. This approach, while not suitable for all scenarios, proved effective in cases where the focus was primarily on the foreground character without the need for additional background context. However, this method may not be universally applicable for video thumbnail generation. Figure 4.19 illustrates some outcomes from these experiments.

Following the establishment of the background, adjustments to the foreground were considered to enhance compatibility with the background and to refine colors and overall image quality. As outlined in the Methodology Section 3.13, introducing noise to the face was deemed highly risky due to the potential alteration of facial features, rendering them unrecognizable. Among various alternatives tested, including the reduction of noise addition, decreasing the sampling steps, etc., the only successful method involved diminishing the InPaint conditional mask strength. This parameter determines the extent to which an image should adhere to its original configuration, addressing issues related to artifacts and the preservation of charac-ter recognizability. Figure 4.20 provides a visual comparison and elucidates the challenges associated with artifacts and loss of recognizability as mentioned.

## 4.16   GUI Tool

Numerous factors have influenced the decision to develop the final graphical user interface (GUI) tool as a web application, thereby ensuring its ease of access and utility for professional thumbnail designers. The specific requirements for this tool have been delineated in the Methodology Section 3.14. This section aims to elucidate the challenges encountered in implementing these requirements.

The necessity for the tool to be accessible via a web interface stems from the desire to obviate the need for individual members of the thumbnail design team to undergo the potentially intricate process of installing and configuring software on their local machines—an approach that is less than ideal. The execution of a web application mandates the development of a backend in conjunction with a frontend capable of supporting multiple users who may concurrently work on the same video project. Moreover, it is imperative to maintain a persistent state of the work progress within a database to ensure that selections made by any team member are visible to all others.

While the tailored development of both backend and frontend, along with the meticulous design of the user experience (UX) and user interface (UI) of the GUI tool, hold significance, they do not constitute the primary focus of this thesis. Consequently, we opted to utilize a GitHub repository named Gradio [1]. Gradio, a Python library, facilitates the rapid creation and deployment of web-based interfaces for machine learning models, enabling developers and data scientists to construct interactive UIs for their applications without necessitating extensive knowledge in web development.

Efficiency is pivotal to the implementation strategy, and is achieved by precomputing all data in the pipeline whenever a video is added to the Play Suisse database. Consequently, by the time a designer selects a thumbnail from the automated choices, the data is already prepared, stored in the database, and all real-time computations, except for calculating CLIP embeddings for newly added keywords (which are also expedited), are conducted swiftly. This design ensures a user experience that is overall satisfying and efficient. More details on other optimization strategies applied are explained in Appendix.

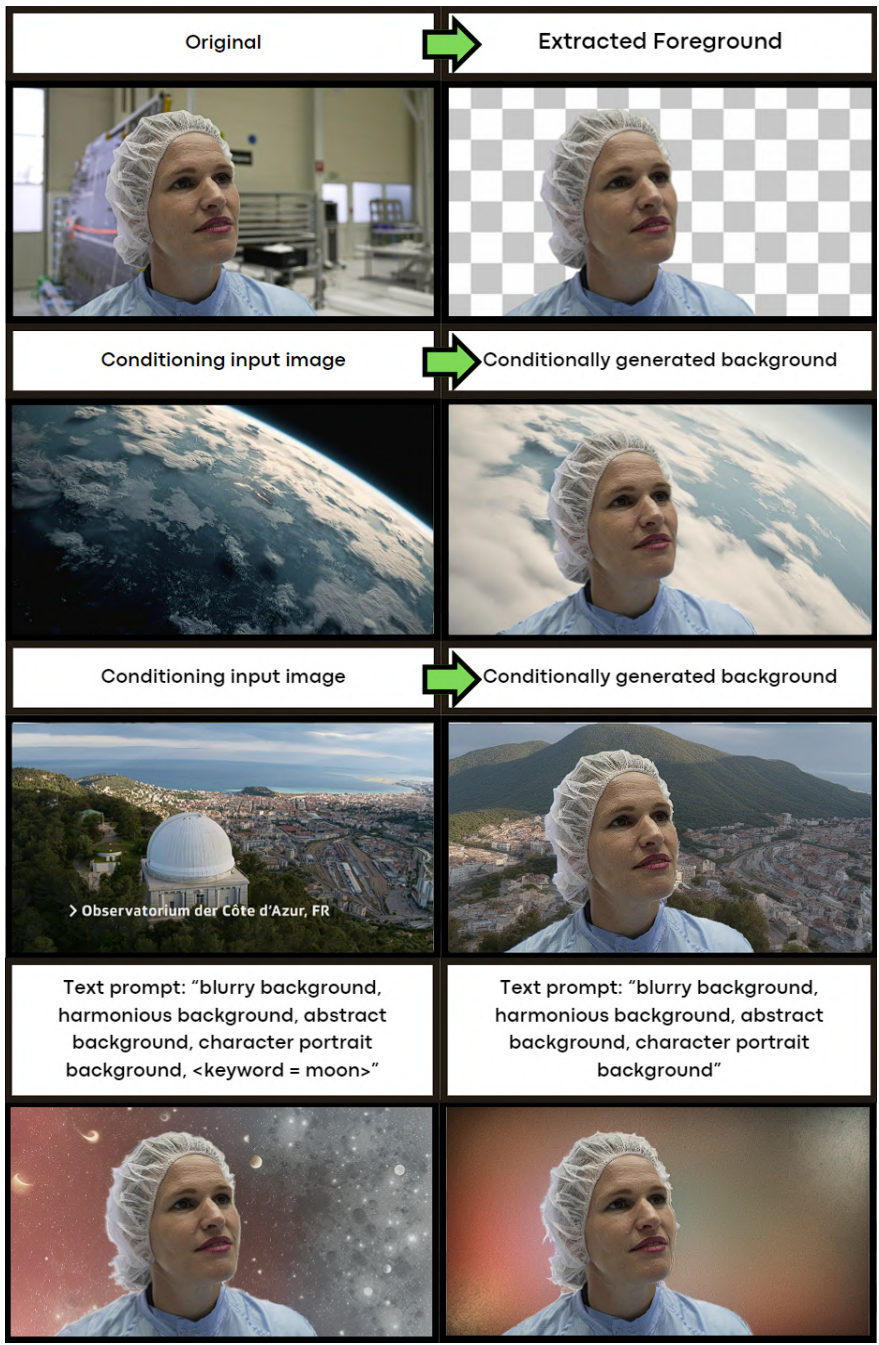

Figure 4.19: The first row of this image shows the original image and the extracted foreground. Moving to the second and third rows, on the right side they display the original foreground with a new background generated by conditioning the diffusion model with the image on the left side, using ControlNet [126]. This technique provides a better harmonization of the composite image compared to just replacing the background on the left. Finally, the fourth row depicts backgrounds generated without ControlNet, but only providing text prompts. The prompt on the left is dinamically modified by adding one of the extracted keywords "moon". Some artifacts are noticeable in regions in proximity of the mask's edges.

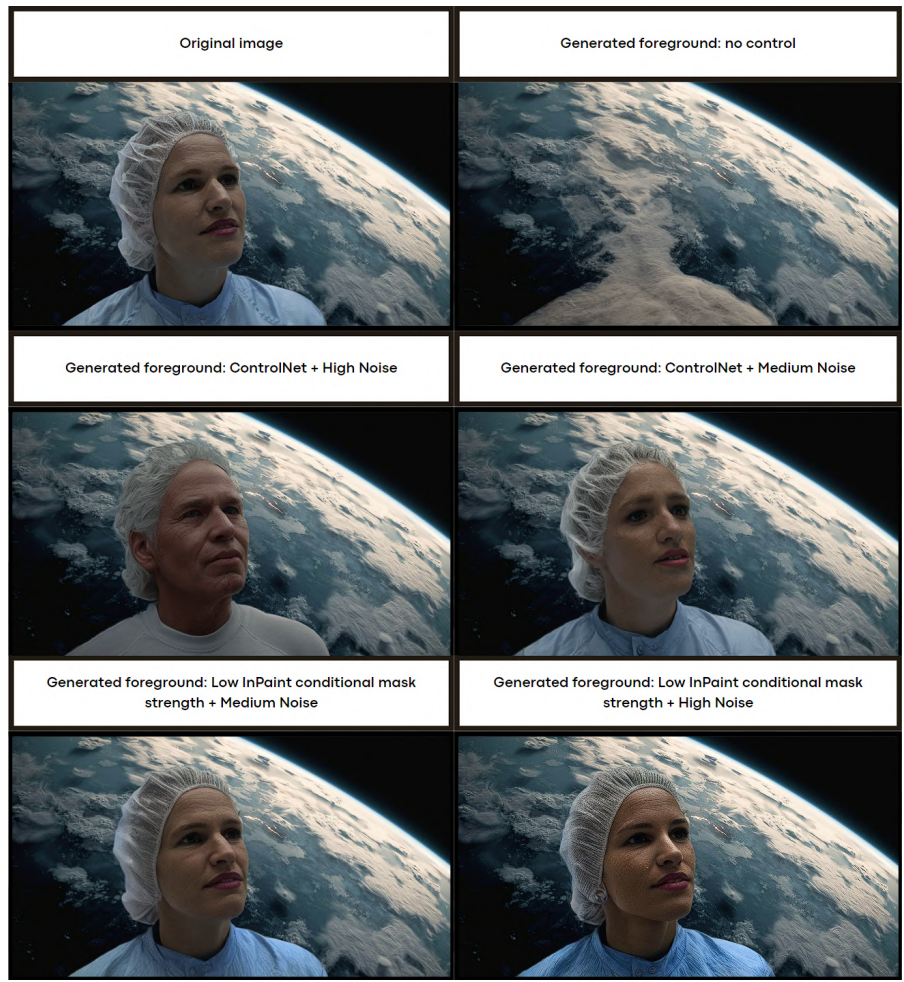

Figure 4.20: The image illustrates various experiments aimed at enhancing the compositional quality and aesthetics of the foreground, ensuring the depicted individual remains identifiable. Sole reliance on the Inpainting model leads to the removal of the person. Employing ControlNet [126] retains the individual but alters their appearance. Reducing the strength of the Inpainting conditional mask succesfully allows for small foreground adjustments while maintaining facial recognizability.

# 5 Results and Discussion

*This chapter presents a comprehensive evaluation and discussion on the thesis project's outcomes. It begins with the design and methodology for evaluating the thumbnail selection process in Section 5.1. Section 5.2 examines the performance of the proposed method with an experiment to determine if our proposal yielded the same or similar images to those selected manually. User preferences for the generated thumbnails are evaluated through a survey and statistically analyzed in Section 5.3, including detailed demographic insights in Subsection 5.3.1. The chapter also includes feedback from professional thumbnail designers to assess the practical application of the proposed method in the professional sphere in Section 5.4. This multifaceted analysis aims to validate the thesis objectives, demonstrating the proposed method's viability and efficiency in improving thumbnail selection processes.*

## 5.1   Evaluation design

In the last chapter we undertook an exhaustive analysis and discussion of each component within our pipeline, conducting rigorous experimentation on individual segments to discern their limitations, assess their acceptability or necessity for resolution, or determine their exclusion from the pipeline. Despite these insights, a comprehensive evaluation of the entire process's output remained unaddressed, leaving the fulfillment of our thesis's objectives in question. Establishing an objective framework for this evaluation proved to be a complex endeavor, particularly given the inherent subjectivity of thumbnail selection tasks.

Several studies in related fields attempt to mitigate this subjectivity by constructing training datasets comprising video and ancillary metadata to predict desired thumbnails or image sets for video summarization purposes. These approaches typically employ various metrics to measure the predicted outputs' proximity to the target outputs within their test datasets. However, such metrics are not without their limitations; a close approximation to a target output does not necessarily indicate the optimal possible output. Consequently, our evaluation methodology for the proposed method sought to adopt a more holistic perspective, aiming to objectively assess each aspect of the thumbnail selection task. This approach involves considering the multifaceted nature of thumbnail selection and attempting to quantitatively and qualitatively evaluate the efficacy of our method across these dimensions.

The primary objective of this project is to enhance the efficiency of thumbnail creation by automating numerous steps in the process without compromising the final product's quality. This innovation aims to significantly reduce the time required to produce multiple thumbnails, thereby catering to diverse user preferences more effectively. By streamlining this process, the initiative seeks to generate a variety of appealing thumbnails in a fraction of the time previously needed to create a single one. The ultimate goal of this endeavor, particularly relevant to Play Suisse, is to augment user engagement. Consequently, an essential aspect for analysis and evaluation in this chapter is the acceleration achieved through the proposed methodology.

Subsequently, our purpose (as detailed in Introduction Section 1.3) defines some key questions to be addressed, which we report again as follows:

- Can a tool leveraging different SOTA AI models automate most of the thumbnail creation process?

- How well does this AI-based approach perform in terms of accuracy and appeal?

- How does its performance compare to traditional manual methods?

- Do users and professional designers find the AI-generated thumbnails preferable or at least comparable to manually created ones?

- Can this AI-based tool improve the workflow of professional designers, and will it be adopted in their regular practices?

To answer these questions, we implemented different evaluation experiments.

We executed the code on a test set of 69 videos to conduct a preliminary evaluation. This assessment aimed to determine whether the thumbnails automatically generated by our system coincide with or resemble those manually created by professional designers for each video. The success of this alignment serves as an indicator of the method's accuracy and visual appeal, suggesting that the automated process is capable of selecting thumbnails that align with the choices made by human designers. Nevertheless, this test alone is insufficient for a comprehensive evaluation. The fact that the algorithm may or may not propose the exact thumbnail chosen by the designer does not conclusively indicate the presence or absence of other potential candidates that could be equally appealing or even superior. The traditional manual selection process involves reviewing the entire video content, a task often limited by time constraints, which could lead to overlooking other viable thumbnail options.

To address this limitation and further assess the effectiveness of our method, we conducted a public survey. This survey compared a thumbnail generated by our system against the current manually selected thumbnail and an additional thumbnail produced using the Hecate algorithm [105], providing an enriched dataset for comparison. This approach aims to evaluate the performance of the automated method in relation to the manual selection process by examining whether users perceive the automatically selected thumbnails as comparable, preferable, or at least closely aligned with the manually selected ones.

To assess the efficacy and workflow enhancements offered by the tool, we asked two professional thumbnail designers from Play Suisse team to conduct an extensive evaluation using over 70 assets. These experts were tasked with determining the number of viable thumbnails selected by our tool, and these findings were then benchmarked against results obtained from Hecate selection. This comparative analysis aims to highlight the potential proficiency of our method in generating a higher volume of valid thumbnail candidates and also underscore its capability to significantly accelerate the thumbnail selection process. Such a metric is crucial, given the objective to produce multiple thumbnails per video, thereby enhancing choice and flexibility in thumbnail selection.

Additionally, we share insights from our direct experience with the output generated by both the pipeline and the tool. This firsthand evaluation was instrumental in curating thumbnails for the survey mentioned, providing a comprehensive understanding of the tool's performance from both a user and developer perspective.

## 5.2   Evaluation of the method's selection performance

As planned, we executed our pipeline on 69 videos, assessing whether the automatic selections featured identical or similar images to the current thumbnails utilized in the Play Suisse catalog. Given that our approach generates proposals tailored to each required aspect ratio, we specifically examined proposals pertaining to the original frame's aspect ratio, excluding those subjected to cropping.

To obtain the closest matching image, we employed the CLIP embedding mechanism [86], to calculate the cosine similarity 3.4 between the proposed images and the existing thumbnail, selecting the image with the highest similarity measure. Considering that a video may comprise up to hundreds of thousands of frames, our downsampling strategy, as detailed in the Implementation Section 4.2, effectively reduces this number to a manageable few thousand frames, averaging to approximately 2.79% ± 0.2% of the total. From this subset, our algorithm further narrows down the selection to several tens of images, based on the detection of faces and the extraction of keywords. Our analysis revealed that, within the initially selected frames after downsampling, the average cosine similarity for the closest matching image stood at 0.886, indicative of a near-exact match barring minor edits and color grading adjustments. However, this average dropped to 0.799 for the final set of proposed thumbnails. As illustrated in Figure 4.1 in the Implementation chapter, a similarity score of 0.886 essentially denotes identical images, while the lowest observed score of 0.673 suggests images that are different yet strikingly similar. Thus, an average score of 0.799 is deemed satisfactory.

**Results and Discussion**

Further scrutiny of each video's current thumbnail against the closest matching image from our proposals revealed that, out of 69 videos, 13 utilized images sourced from the web. This practice can be attributed to the low quality of some videos, making it challenging to extract engaging frames that could entice viewership. In such cases, the selection of relevant stock images from the web, despite potentially compromising Play Suisse's objective to eschew clickbait thumbnails, appears as a pragmatic solution. Nevertheless, our innovative approach of generating thumbnails could address this issue for videos with problematic quality, ensuring the thumbnail content remains predominantly derived from the video itself.

Out of the total 56 videos where a perfect match was possible, our method's proposal contained matching images in 30 of them. Therefore, in our experiment, our method successfully provided matching images in 53.6% of cases where it was feasible to obtain the same image as the current thumbnail.

Out of the 39 remaining videos lacking a match or with thumbnails sourced from the web, we found that 21 of them had comparable images in the proposal, elevating the total to 51 out of 69 matches. This equates to a 73.9% occurrence rate where the proposal includes at least one similar image (not a perfect match). This analysis provides deeper insight into the notable average cosine similarity value obtained. Such outcomes meet our criteria effectively, as demonstrated in Figure 5.1, which illustrates visual examples of some of the matched pairs discussed.

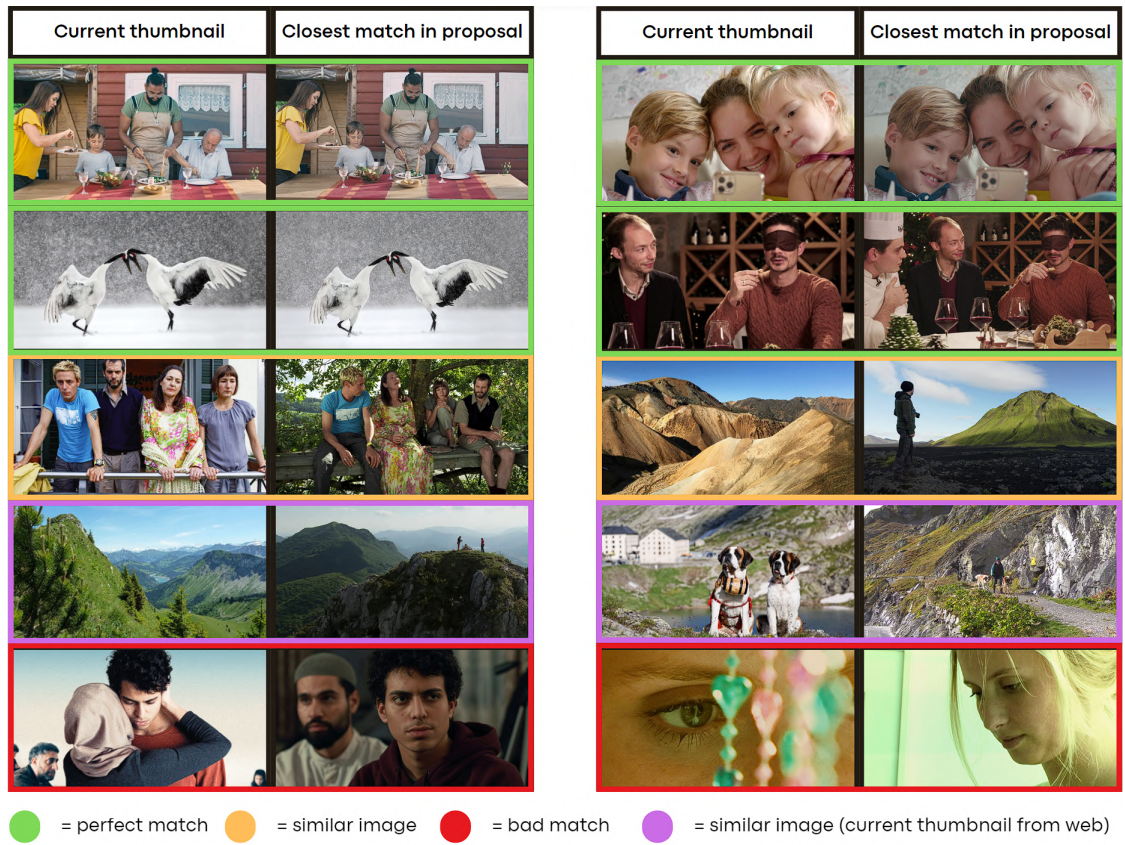

Figure 5.1: The image displays current thumbnails from the video catalog alongside the closest matches from the proposal generated by our method. Green boxes indicate perfect matches, while yellow ones signify instances where the proposal contains a similar image but not an exact match. Purple boxes also showcase where the matching image is similar, but the current thumbnail is sourced from the web, making a perfect match unattainable. Red boxes denote cases where no matching image was found in the proposal.

## 5.3   User preference evaluation

Despite the promising outcomes of the initial evaluation, these results alone are insufficient to definitively ascertain the efficacy and criteria fulfillment of the tool. Given the project's objective to accelerate the discovery and selection process of potentially multiple thumbnails, it becomes imperative to explore the possibility of identifying at least one alternative image, distinct from the current thumbnail, which might be equally or more appealing. To this end, conducting a public survey emerged as the optimal approach to quantitatively and qualitatively assess subjective preferences for thumbnails, which inherently vary according to individual interests and tastes.

In constructing the survey, we opted to include the existing thumbnail, an alternative image selected by our proposed method, and a third image selected via Hecate's method [105], dis-played concurrently for comparison. Participants were asked to rate, on a scale from 0 to 5, their likelihood of clicking on each thumbnail if encountered on their preferred streaming service dashboard, to gauge interest in viewing more content such as the summary. Additionally, a direct question solicited their preference among the three images, enabling an absolute choice. This two-pronged query design not only facilitates a deeper understanding of individual thumbnail preferences but also provides insight into the potential for enhancing user engagement with our tool. To mitigate any bias, thumbnails were presented without their titles, as they would have been available only on current thumbnails that are manually created. On the other hand, participants were informed of the original titles and encouraged to consider the relevance of the image to the title text in their evaluation, acknowledging that a title-congruent image could influence their preference.

We adopted a randomized approach to assign letters (A, B, C) to images derived from different sources, ensuring that any potential bias related to the source was minimized. To enhance the fairness and quality of the selection process for images to be put in the survey, each image was evaluated and chosen based on consensus between two individuals. We aimed to select images that were not only aesthetically pleasing but also distinct from the current video thumbnail while maintaining relevance to the video title. This criterion was particularly pertinent given that the current thumbnail could often be found in the automatic selection pool as well; selecting a similar or identical image would have been redundant.

The survey was designed with the dual objectives of maximizing participation and obtaining meaningful insights within a concise framework. To this end, we limited the number of image comparison questions to 17, estimating an average completion time of 8 minutes. This decision was informed by the rationale that a lengthy survey might deter potential participants, thus compromising the quantity and quality of the feedback collected.

Of the 17 comparisons, 5 featured a vertical thumbnail option, exclusively produced by our method, in contrast to the horizontal formats typically provided by Hecate and the existing thumbnails. This aspect of the survey aimed to explore the potential of vertical thumbnails in enhancing viewer engagement, despite the inherent format variation introducing a possible bias. It is noteworthy that the presentation order of the vertical thumbnails (among A, B, C) was varied, and the participants were unaware that only our method was capable of generating vertical thumbnails. This design choice allowed us to impartially assess the appeal of vertical thumbnails without influencing participant preferences based on format awareness.

Figure 5.2 illustrates two representative question formats used in the survey, including an example featuring the vertical thumbnails generated by our method.

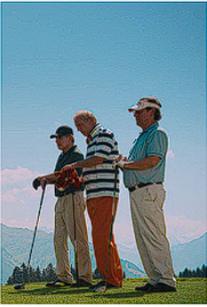

Figure 5.2: The image depicts two examples of the survey question format. The example on the right includes a vertical thumbnail.

### 5.3.1 Participant information

The survey garnered responses from a diverse cohort of 82 participants, predominantly hailing from three European countries: Switzerland, Italy, and France. Given the mandatory nature of the survey, the completion rate across all questions was uniform. Among the 82 respondents, the gender distribution was 35.8% female, 61.7% male, and 2.5% identifying as other genders.

## Results and Discussion

In terms of age demographics, the participants were largely young adults, with 69.1% aged between 20 and 30 years. The distribution for other age groups was as follows: 4.9% were under 20 years old, 8.6% were between 30 and 40 years, 7.4% fell in the 40 to 50 years category, 4.9% were between 50 and 60 years old, and an additional 4.9% were over 60 years of age.

The survey also inquired about the participants' professional and academic backgrounds, specifically their connection to image or content creation. A minority, 12.3%, reported a direct relation to image creation, while 19.8% indicated their involvement in content creation more broadly, leaving 67.9% unrelated to either field. An open-ended question aimed at those engaged in image-related fields garnered 19 responses, revealing a wide array of professionals including photographers, professional actors, innovation consultants, graphic designers, thumbnail creators, social media managers, music producers, web developers, UI/UX designers, brand managers, and videographers. This diversity underscores the multifaceted nature of the participants' engagement with image and content creation, as visualized in Figure 5.3.

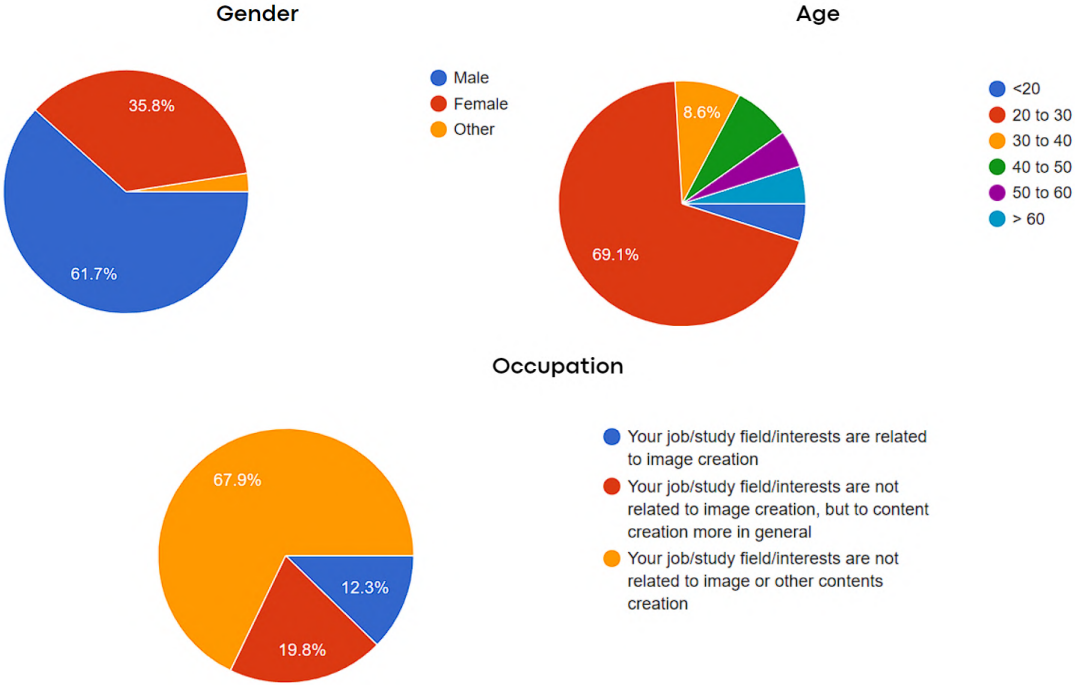

Figure 5.3: Pie charts illustrating the distribution of survey participants by gender, age, and occupation.

### 5.3.2   Survey results

**Analysis of image preference data**

In the evaluation of survey outcomes, a comprehensive statistical analysis was conducted on the collected data to discern user preferences for various image selection methods. The analysis revealed that our methodology (hereafter referred to as "Our Method" or "Proposed method") outperformed others, securing a preference rate of 45.77% among participants, surpassing the current thumbnail method which garnered 37.88%, and significantly outperforming the Hecate method at 16.36%.

To further elucidate these findings, we employed a combination of the Analysis of Variance (ANOVA) One-Way test followed by the Tukey Honestly Significant Difference (HSD) test. The ANOVA One-Way test serves as a critical tool for comparing mean scores across several groups—here, the different image selection methods—to ascertain the presence of any statistically significant disparities among them. This test is indispensable when the objective is to evaluate the efficacy of multiple methods or treatments relative to a singular variable, in this instance, user preferences for images. While the ANOVA test identifies if there exists at least one method with significant variance from others, it does not delineate which specific methods differ.

To bridge this gap, the Tukey HSD test is implemented as a post-hoc analysis, facilitating pairwise comparisons between groups to pinpoint the exact pairs of methods exhibiting significant differences in their means. It effectively manages the family-wise error rate, thereby ensuring the reliability of these multiple comparisons by mitigating the risk of Type I errors.

The results of the one-way ANOVA test reveal a significant disparity among the groups, as evidenced by an F-statistic of approximately 21.90 and a P-value of approximately $1.75 \times 10^{-7}$. Further examination through the Tukey HSD post-hoc analysis elucidates the specific pairs of methods between which these significant variances occur. It was found that there is a substantial difference ($p < 0.001$) in mean preference scores between the "Current thumbnail" and the "Hecate method," as well as between the "Hecate method" and "Our method", with the "Hecate method" underperforming in both comparisons. Conversely, the analysis between the "Current thumbnail" and "Our method" did not reveal a statistically significant difference ($p = 0.2102$), suggesting that the preference levels for these two methods are comparably similar when considering an alpha level of 0.05.

Subsequently, the collected categorical data, including age, gender, and occupation, were utilized to segment the results, facilitating a nuanced analysis of preferences within specific demographic categories. To ascertain the variations in preferences or perceptions across different gender groups, we employed the Wilcoxon rank-sum test (also known as the Mann-Whitney U test). This non-parametric test ranks all observations across both groups under comparison, thereafter assessing the statistical significance of the rank sum discrepancies between them.

## Results and Discussion

Furthermore, the Chi-Square test of independence was applied to evaluate the potential associations between two categorical variables. By comparing the observed frequencies within each category of a contingency table against the expected frequencies—assuming independence between the variables—this test elucidates significant relationships between demographic characteristics (such as gender, age, and occupation) and method preferences. This analysis is pivotal for identifying whether certain demographic segments exhibit distinct preferences for specific methods, thereby enhancing our understanding of method suitability across various population subsets. Finally, we also repeated the ANOVA One-Way and Tukey HSD tests.

Starting our analysis from gender, the leaderboard remains mostly consistent, with females showing a stronger preference for "Our method" over the "Current thumbnail." However, it's noteworthy that "Other" gender individuals, although with only two data points and limited statistical significance, favor "Current thumbnail" the most. Regardless of gender, "Hecate" consistently ranks last. The Chi-square test indicates that gender doesn't significantly influence preference for any evaluated method. However, ANOVA One-Way, Tukey HSD tests, and Wilcoxon rank-sum test demonstrate that females have a statistically significant higher preference level for "Our method" compared to "Current thumbnail" compared to males.

Likewise, analyses based on age and occupation reveal no significant influence on method preference. Despite different occupational groups agreeing on the method leaderboard, certain age groups—specifically those under 20 or between 50 and 60—alternate the positions of "Current thumbnail" and "Proposed method." However, due to the small sample sizes within these age brackets, this doesn't provide substantial statistical evidence to contradict the aforementioned findings for these specific parts of the participants.

### Analysis of click likeliness increase data

Next, we directed our focus toward examining the variations in user propensity to click on thumbnails, as gauged on a scale from 0 to 5. Contrary to concentrating on the absolute values assigned, our interest lay in the relative enhancements observed across different methodologies. For instance, should Question N have Method A to be assigned the lowest score 2, and Method B and C's thumbnails garner scores of 4 and 5 respectively, our primary concern is the incremental rise in user engagement: Method A demonstrates no improvement, while Methods B and C exhibit increases of +2 and +3 respectively. Upon applying this analytical framework across all queries, it emerged that, on average, our method manifested an uptick in the likelihood of user clicks by 1.44, surpassing the "current thumbnail" method's average increase of 1.21, and significantly outperforming the Hecate method, which only achieved a 0.59 average increase.

Further statistical scrutiny, employing ANOVA One-Way and Tukey HSD tests, corroborated the presence of statistically significant disparities in mean likelihood increase ($p < 0.001$) between the "Current thumbnail" and "Hecate method," as well as between the "Hecate method" and "Our method". In each instance, the Hecate method was outperformed. Conversely, the difference between the "Current thumbnail" and "Our method" was not statistically significant ($p = 0.1551$), suggesting comparable performance between these two approaches.

The analysis was extended to encompass categorical classes, with the objective of identifying any statistically significant variations in method preference across different categories. However, similar to the previous evaluations, this investigation did not yield any significant findings, indicating a uniform preference distribution across categories.

Tables 5.1 and 5.2 show the average scores attributed to each method for every question posed. An examination of these tables reveals a discernible correlation, particularly notable in the highest-scoring entries of each row. This observation aligns with the intuitive expectation that the thumbnail with the highest click probability is also likely to be preferred. Nevertheless, we examined some less common cases where this pattern did not occur.

### Analysis of data for different thumbnails orientation

Subsequent analyses were conducted to examine the influence of thumbnail orientation on user preferences, dividing the dataset into two distinct groups: questions featuring vertical thumbnails and those with exclusively horizontal images. For questions incorporating vertical thumbnails, a notable shift in user preference was observed. The current thumbnails emerged as the most favored, securing a 45.4% preference rate among participants, closely followed by the proposed method at 42.9%, and Hecate trailing with 11.7% preference. Statistical evaluation through ANOVA One-Way and Tukey HSD tests revealed no significant difference in mean preferences between the "Current thumbnail" and "Our method" (p = 0.9511), indicating comparability in user preference levels. A Chi-Square test further highlighted a significant gender-based association in preferences for the "Proposed method" (p = 0.012), with a pronounced preference among the Female category for our method when applied to vertical thumbnails, contrary to other demographic categories.

In terms of engagement metrics for vertical thumbnails, average click rates were calculated as 1.339 for our method, 1.348 for the current thumbnail, and 0.402 for Hecate. The statistical tests confirmed that the differences in mean click rates between the top two methods were not significant, mirroring the findings related to user preferences.

Conversely, analysis focusing solely on questions with horizontal images yielded results consistent with the overall findings of the comprehensive study. The preference distribution for this subset was 47.0% for our method, 34.8% for the current thumbnail, and 18.3% for Hecate. Similarly, engagement through clicks showed an enhancement of 1.49, 1.15, and 0.67 for our method, the current thumbnail, and Hecate, respectively.

These findings show the nuanced impact of thumbnail orientation on user preferences and engagement, offering valuable insights for optimizing content presentation strategies. Figure 5.4 shows some of the mentioned results.

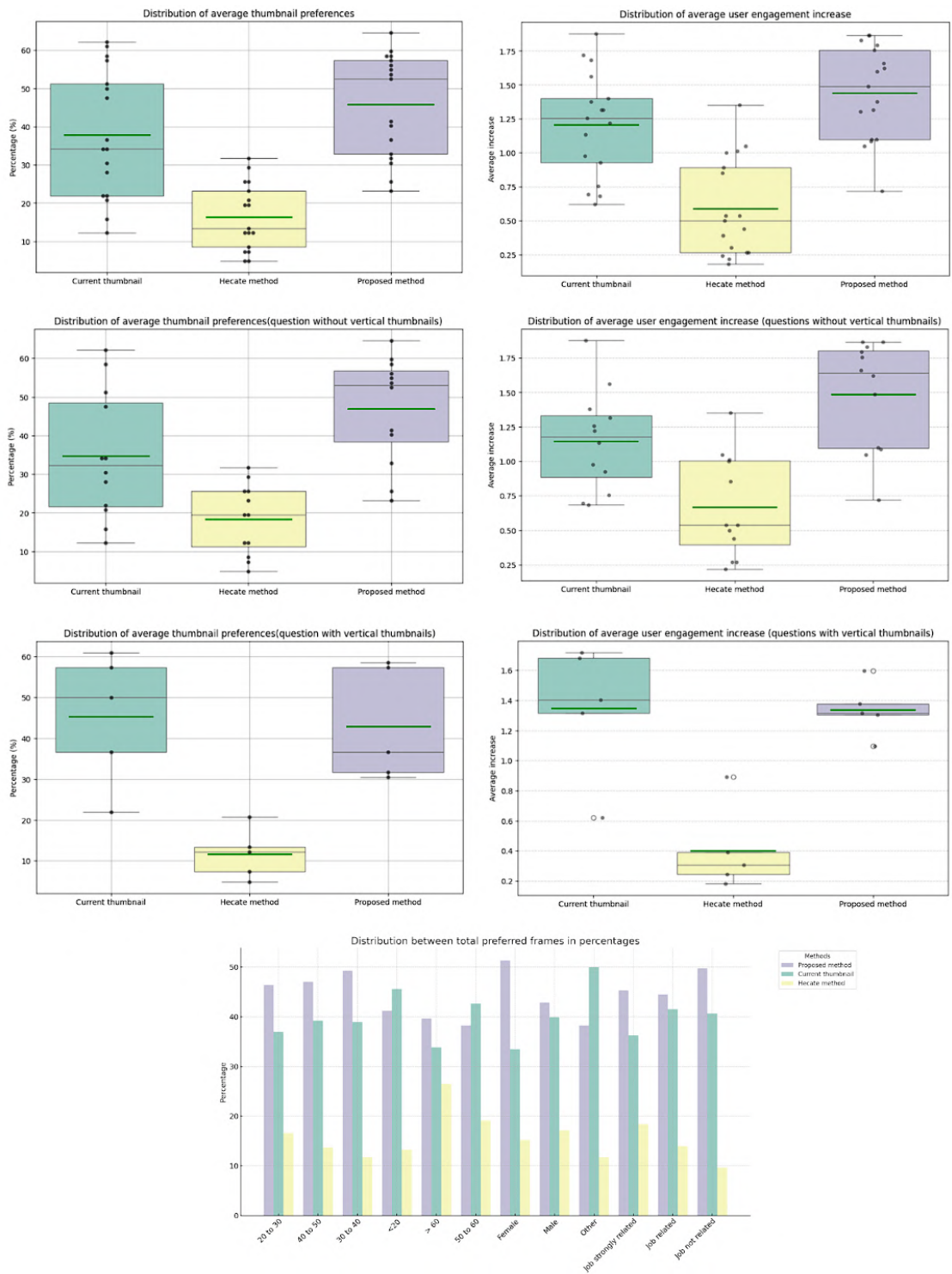

Figure 5.4: The figures consist of three sets of box plots displaying average survey scores: one for all questions, another excluding those with vertical thumbnails, and a third comprising only questions with vertical thumbnails. Mean and median values are depicted by green and black lines, respectively. Box boundaries represent the Q1 and Q3 quartiles, with whiskers indicating minimum and maximum values. A bar plot below compares preference percentages across gender, age, and occupation categories.

**Discussion**

Our comparative analysis demonstrates that our methodology garners a more favorable reception from users than Hecate [105] and slightly outperforms the current manual thumbnail selection in terms of general preference and user engagement, indicated by an increased likelihood of image clicks. Notably, our approach ranks second for vertical thumbnails. A deeper investigation, beyond mere average values, through statistical tests reveals that the data distributions for our method and the current manual thumbnail selection are similar, thereby affirming their comparability.

This equivalence with manual selection, coupled with the significant acceleration in the thumbnail selection process, underscores the superiority of our method over traditional manual approaches.

Vertical thumbnails are not universally suitable for all types of content; they are particularly advantageous for content featuring people, as this aspect ratio can be leveraged effectively. It's worth noting that vertical thumbnails will not replace horizontal ones, but cohexist. Consequently, a more meaningful comparison could have been drawn among vertical thumbnails for all three methods. Unfortunately, such a comparison was not feasible as these thumbnails were not present in Play Suisse for the analyzed videos, nor were they produced by the Hecate method.

The conducted survey successfully addresses several objectives set for our method, establishing its performance as comparable to, if not preferable over, traditional manual methods. This preference is echoed not only by general users but also by professional designers involved in image creation. Figure 5.5 shows some interesting examples.

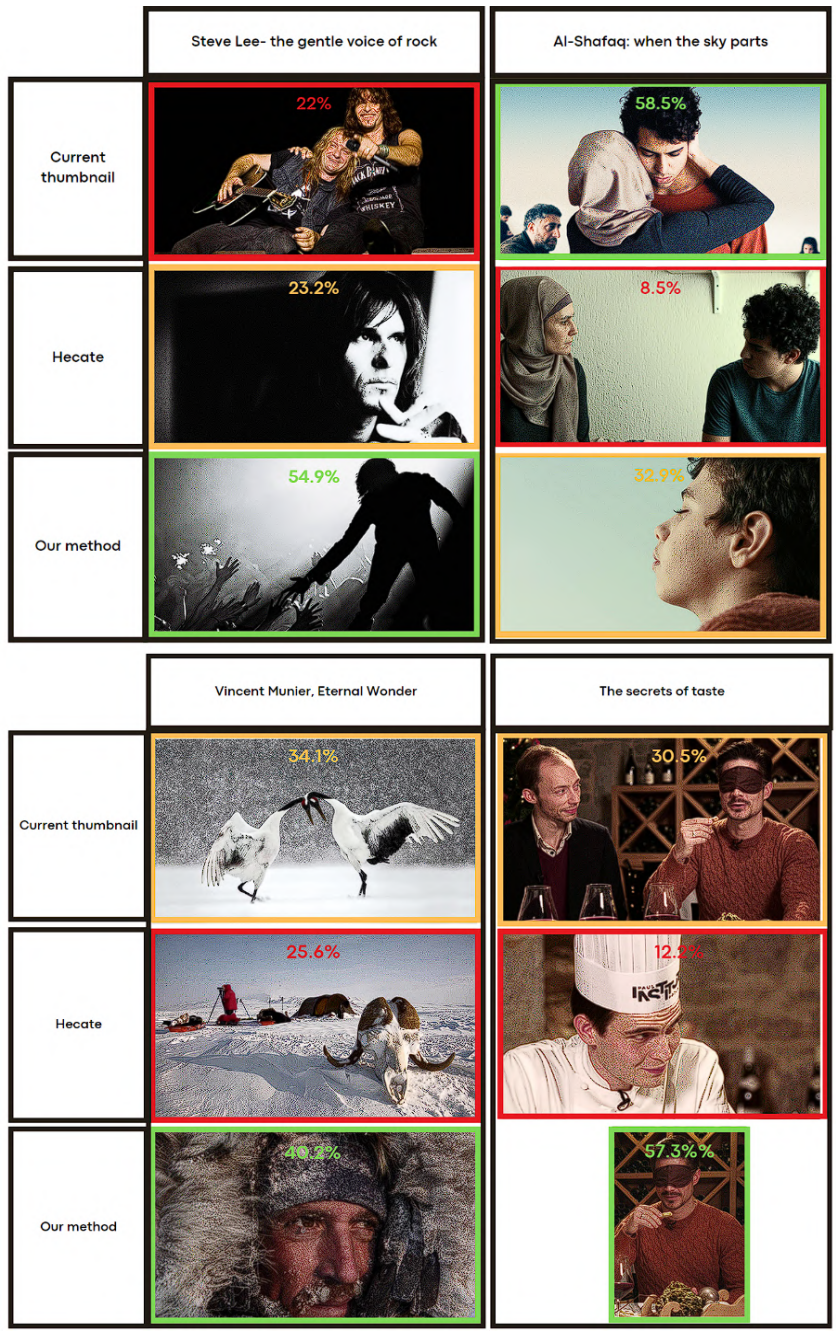

Figure 5.5: Examples of thumbnails' trio from survey's questions, with titles at the top and methods to the left. Each image is marked with the average preference percentage of users and color-coded (green=highest preference, yellow=medium, red=lowest). The top left example has the current thumbnail from the web. Interestingly, images from the video are more effective, so having a tool to find them fast is key. In the top right example, the current thumbnail, an edited composite of various images, emerged as the most favored on average, but demands a lot of time to create, while our method still demonstrates efficacy. The bottom right example has a vertical thumbnail similar to the current thumbnail. However, in this instance, the vertical format appears to have heightened user engagement more effectively.

| | Current thumbnail | Hecate method | Proposed method |
|---|---|---|---|
| The science of beer | 12.20% | 31.71% | 56.10% |
| The Swiss Way | 47.56% | 29.27% | 23.17% |
| 8x15. from KuFa Lyss | 21.95% | 20.73% | 57.32% |
| Steve Lee - the gentle voice of rock | 21.95% | 23.17% | 54.88% |
| Crime scene - hillside location with a view | 15.85% | 25.61% | 58.54% |
| Al-Shafaq: when the sky parts | 58.54% | 8.54% | 32.93% |
| An unwanted trip | 28.05% | 19.51% | 52.44% |
| The secrets of taste | 57.32% | 12.20% | 30.49% |
| The slide | 51.22% | 7.32% | 41.46% |
| Vincent Munier, Eternal Wonder | 34.15% | 25.61% | 40.24% |
| My little Tibet | 20.73% | 19.51% | 59.76% |
| The sunday quartet | 60.98% | 7.32% | 31.71% |
| Beauty and her beasts | 30.49% | 4.88% | 64.63% |
| On the trail of the damned of Daesh | 62.20% | 12.20% | 25.61% |
| Journey of Hope | 50.00% | 13.41% | 36.59% |
| Sparrows and snorkels | 34.15% | 12.20% | 53.66% |
| A Barber in Lugano | 36.59% | 4.88% | 58.54% |

Table 5.1: Average image preference percentage per method, for each video of the questions. Green shows the highest percentage of the three methods per video, yellow is the middle one and red is the lowest.

## 5.4   Professional thumbnail designers feedback

The concluding phase of our evaluation sought to elucidate how the implementation of the proposed tool influences the workflow of professional thumbnail designers, specifically in terms of process acceleration, workflow enhancement, and potential integration into their daily practices. To this end, we facilitated an extended trial period of the tool for two professional thumbnail designers, aiming to gather comprehensive feedback on its utility and impact.

The feedback from the first designer, who tried the tool for a limited amount of time, underscored the tool's efficacy in streamlining the thumbnail selection process. By exploiting the full range of functionalities offered by the tool, she was able to swiftly assess and choose from a diverse array of automated thumbnail suggestions for two distinct videos. Within minutes, she selected three thumbnails per video, each with a unique focus and intended purpose, thereby demonstrating the tool's capability to provide diversified content for different viewers preferences and tastes. Figure 5.6 shows the examples she selected with the corresponding different purposes of each thumbnail.

| | Current thumbnail | Hecate method | Proposed method |
|---|---|---|---|
| The science of beer | 0.76 | 1.35 | 1.87 |
| The Swiss Way | 1.22 | 1.00 | 0.72 |
| 8x15. from KuFa Lyss | 0.62 | 0.89 | 1.38 |
| Steve Lee - the gentle voice of rock | 0.68 | 1.05 | 1.62 |
| Crime scene - hillside location with a view | 0.70 | 1.01 | 1.79 |
| Al-Shafaq: when the sky parts | 1.38 | 0.44 | 1.05 |
| An unwanted trip | 1.26 | 0.54 | 1.66 |
| The secrets of taste | 1.68 | 0.30 | 1.30 |
| The slide | 1.88 | 0.22 | 1.83 |
| Vincent Munier, Eternal Wonder | 0.98 | 0.85 | 1.09 |
| My little Tibet | 0.93 | 0.54 | 1.49 |
| The sunday quartet | 1.72 | 0.24 | 1.32 |
| Beauty and her beasts | 1.13 | 0.27 | 1.87 |
| On the trail of the damned of Daesh | 1.56 | 0.50 | 1.10 |
| Journey of Hope | 1.40 | 0.39 | 1.10 |
| Sparrows and snorkels | 1.32 | 0.27 | 1.76 |
| A Barber in Lugano | 1.32 | 0.18 | 1.60 |

Table 5.2: Average user engagement increase per method, for each video of the questions. Green shows the highest value of the three method per video, yellow is the middle one and red is the lowest.

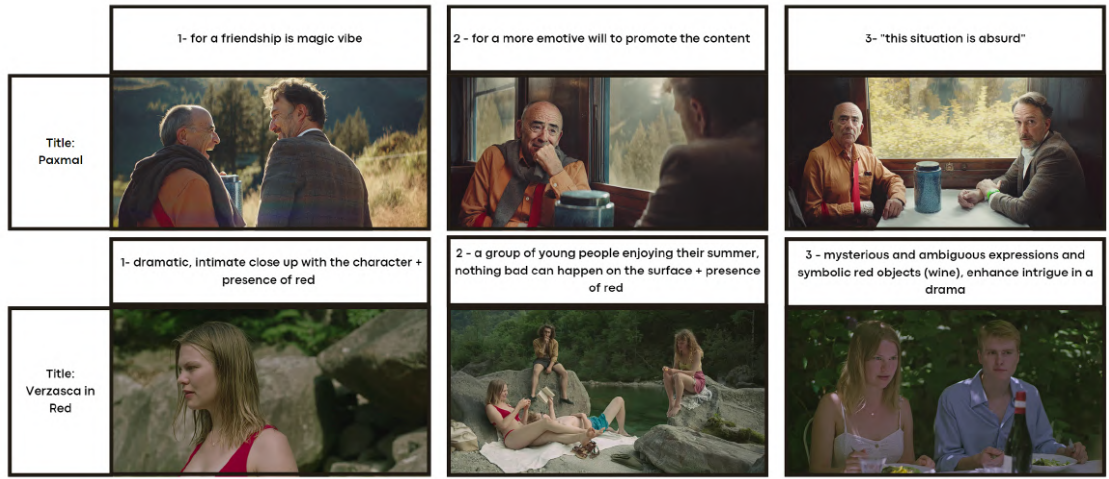

Figure 5.6: Examples of thumbnails trio with different communication purposes,individuated by the professional designer from the automatic proposal set.

The second designer's evaluation was markedly more detailed, offering an in-depth analysis and feedback on the tool's performance and its implications for professional thumbnail design practices.

Initially, the designer conducted a quantitative comparison between our tool and Hecate's method by assessing the number of valid thumbnail candidates proposed for a set of 20 videos. The results showed our method significantly outperforming Hecate, with an average of 2.5 valid candidates per video compared to Hecate's 0.7, marking a substantial 3.57-fold increase in efficiency.

The designer's feedback then delved into a qualitative assessment of the tool. Generally, the tool received positive remarks, particularly in its suitability for fiction content such as movies or TV series. It was noted that the selected thumbnails exhibited a wide variety and were representative, with a strong emphasis on capturing people and their emotions effectively. However, when it came to reportage-type documentaries, where the editorial focus is not primarily on portraits of individuals, the feedback indicated a less satisfactory outcome. These documentaries, often characterized by specific thematic focuses or multi-faceted subjects, rendered the initial selection of faces with emotions less relevant overall. This makes sense as in such cases the only relevant part of the proposal set would become the keyword-related images. Nonetheless, the inclusion of a Manual filter panel proved to be a crucial feature, allowing users to input keywords outside the tool's predefined scope. This capability facilitated the discovery of images that met the dual criteria of being both aesthetically pleasing and contextually relevant, thereby enabling the systematic identification of the best images for the video. Moreover, the commendable processing time of just a few seconds per keyword search was noted as highly acceptable. This Manual filter panel will be key also for having multiple thumbnails per video, enhancing the overall viewing experience with just a few clicks.

## Results and Discussion

To conclude, the feedback states that the tool has significantly improved workflow efficiency by accelerating the process, eliminating the need for manual video scrolling. The generated thumbnails are of comparable quality to manually selected ones, often providing at least one suitable option. However, when only one choice is available, it limits flexibility in text placement on the selected image.

The tool offers a wide range of usable images, surpassing the Hecate method in choice and diversity. Its keyword-based manual image search enhances diversification, with vertical thumbnails proving particularly effective for highlighting portraits and emphasizing facial expressions. However, future tests should assess image quality for vertical images, as they are frequently cropped and zoomed for better composition.

Parallel to the experiences of the professional designers, our own engagement with the tool during the survey creation mirrored the challenges and decisions intrinsic to professional thumbnail design. This process entailed selecting images that were not only representative of the video content and title but also distinct from existing selections made by Hecate or current thumbnails. Despite our non-expert status, the tool's design—characterized by its automatic segmentation into faces, emotions, and keywords—significantly facilitated our decision-making, allowing us to identify and choose preferred thumbnails swiftly. This aspect was corroborated by the survey outcomes, wherein the thumbnails selected via our tool were frequently favored over alternative options.

These observations collectively affirm that the tool substantially speeds up the thumbnail selection process, extending its accessibility to individuals beyond the realm of professional design. This feature is particularly relevant for Play Suisse, where the selection of efficient and effective thumbnails is not always exclusive to professional designers.

# 6 Conclusion

In this thesis we introduce a comprehensive pipeline designed to automate the process of selecting thumbnail candidates for video-on-demand (VOD) content, including movies, documentaries, concerts, and television series. This pipeline synergizes video processing with accompanying metadata, such as textual video summaries, to facilitate the rapid proposal of thumbnail candidates. An important part of the process is the development of a user-friendly graphical user interface (GUI) tool, which streamlines the thumbnail selection process. This tool offers dual functionality: it enables automatic selection for efficiency and incorporates a detailed manual filtering panel for more specific searches. Given the premium nature of the content targeted by our tool—wherein professional thumbnail designers traditionally manually select thumbnails—the selected candidates must meet high aesthetic and representational standards. Our objective is to speed up the thumbnail production process, allowing for the creation of multiple, user-tailored thumbnails per video to enhance viewer engagement.

To achieve these standards, our methodology incorporated a comprehensive set of criteria encompassing a breadth of diverse, representative proposals with superior aesthetic quality. These criteria included the generation of vertical portraits, meticulous consideration of logo placement, and a focus on capturing human emotions. Our initial step involved an exhaustive review of existing literature in the domains of video summarization and thumbnail selection. This review served to pinpoint the predominant solutions addressing common challenges within these fields, whilst concurrently highlighting the limitations and inapplicability of such methods to our specific context. Subsequently, we developed an innovative approach, comprising a sophisticated, multi-faceted pipeline in conjunction with an intuitive graphical user interface (GUI). This GUI is designed to facilitate user-friendly interaction with the data processed through the pipeline. Through rigorous research, iterative experimentation, and thorough evaluation of each component within the pipeline, we established a refined pipeline architecture that integrates the latest state-of-the-art (SOTA) models tailored to address each identified subproblem. The evolution of our pipeline's design was significantly influenced by iterative feedback from a team of professional thumbnail designers. This collaborative process was pivotal in identifying the essential elements of the pipeline and determining the aspects that required prioritization.

The culmination of our pipeline endeavors to meet the previously delineated criteria, finalizing with steps such as color grading and title placement being left to the discretion of the designer. We advocate for a thumbnail generation approach that incorporates automatic cropping to facilitate the automatic suggestion of thumbnails in any preferred aspect ratio. This is complemented by the integration of aesthetic quality and shot scale predictors, alongside diverse facial models for identifying principal actors, their emotional states, and the occurrence of closed eyes. To ensure the representativeness of the thumbnails, we incorporate a text-image semantic consistency check.

In an effort to foster diversity within the thumbnail selection process, we have developed presets that utilize the extracted data in various ways, as determined through meticulous data analysis facilitated by a bespoke GUI tool crafted for this specific purpose. These presets are designed to prioritize different content types, enhancing diversity. Moreover, we have introduced two innovative techniques for image composition to further augment diversity: one technique involves the replacement and harmonization of backgrounds, while the other employs conditionally generated backgrounds through diffusion models. Due to the delimited amount of use cases where such generated images apply in Play Suisse catalog, we put less focus on this aspect.

To rigorously assess the efficacy of our proposed method, we devised several evaluation strategies that align with the objectives of our thesis. This involved conducting a public survey to gauge thumbnail preferences among our method, an alternative method, and the current approach of manual selection. Additionally, we sought feedback on the GUI tool from professional thumbnail designers following its extensive use and executed an experiment to determine if our proposal yielded the same or similar images to those selected manually.

The evaluations consistently returned favorable outcomes, demonstrating that our proposed method effectively met all specified criteria. It accelerated the process, improved workflow efficiency, and achieved high levels of accuracy and visual appeal. Furthermore, it was comparably favored by users over thumbnails created through conventional manual techniques and successfully increased user engagement.

## 6.1 Future work

In the course of implementing our methodology, we identified several aspects that, while not immediately critical, presented opportunities for future enhancement. These aspects were subsequently deprioritized due to time constraints.

One such area of potential development involves the aesthetic of images. Despite our satisfaction with the performance of the aesthetic prediction model currently integrated into our system, we have employed a flexible Python library designed to seamlessly incorporate new models as they become available. This approach ensures that should a more effective model be released, we could readily evaluate its impact and consider its integration to potentially replace the existing model.

Our initial efforts have less prioritized the development of thumbnail generation techniques involving image composition and background modification. Future research could delve more deeply into this area, especially improving the selection of images for foreground and background extraction to mitigate observed failure cases. The current methodology does not always yield images suitable for use as thumbnails, indicating a need for further investigation into achieving greater consistency. Additionally, the outcomes presented in this thesis were achieved using the first model of Stable Diffusion [92], selected for its rapid generation capabilities and modest VRAM demands, which aligned with the constraints of our VRAM-limited testing environment. With the advancements in AI technology, such as the introduction of the SDXL Turbo model [97], there is potential for more efficient background generation. Although at the time of this thesis, the inpainting models for this newer method are not yet available, nor is there extensive support for their usage with ControlNet [126], their anticipated availability promises significant improvements in the speed and quality of background generation.

In terms of user interface development, we have utilized the Gradio Python library for its rapid deployment capabilities and ease of use. However, future work could explore the use of more sophisticated libraries that offer greater flexibility and advanced GUI features, potentially enhancing the user experience.

Play Suisse has not specified requirements regarding the processing time for the pipeline, given the limited volume of daily video content. However, our pipeline is adaptable for use in other contexts where video content volumes are substantially higher. In such cases, optimizing the pipeline to reduce processing time, possibly by simplifying steps and functionalities, could yield a lightweight version. This streamlined iteration would maintain the core capabilities of the original tool, ensuring its effectiveness for large-scale operations while accommodating the need for efficiency.

The designer's feedback underscored the necessity for enhancing the level of detail in certain vertical thumbnails. It's acknowledged that when the cropped image focuses on a small portion, there's a natural loss of resolution. We've already explored and successfully experimented with advanced upscaling methods leveraging diffusion models, although we haven't yet integrated them into our pipeline. Future work could easily integrate this additional step.

Additionally, the feedback prompted a request for a more pertinent proposal concerning Reportage-type documentaries. When such content is identified, a slight adjustment to the scoring system could suffice to prioritize images less focused on faces and more on the specific themes of the documentary. These adjustments should be easily implementable and could be considered in future iterations.

# A Appendix

## A.1 Automatic cropping code adaptation for time optimization

The implementation of the automatic cropping method was adapted from an existing GitHub repository that had already integrated the cropping method [125], along with a feature for filtering cropping proposals based on face detection, utilizing DSFD [63]. We modified the code from this repository in several ways. Initially, we devised a function to extract and store face detection predictions, which could also be utilized in other segments of the pipeline as required. Subsequently, after identifying a significant amount of processing time per image, we employed a profiler tool to determine which code sections were most time-consuming. Despite time not being a stringent constraint for our use case, this phase was the most time-intensive, and we aimed to optimize it wherever possible. Our optimizations were twofold: firstly, by employing a more efficient sorting function for the obtained tensor rather than the slower algorithm applied to a standard list; and secondly, the original code required running most steps twice to achieve both the best crop and the best crop with the face centered. We streamlined this process by labeling crops with centrally located faces, applying the crop quality prediction en masse, and subsequently selecting the top n labeled and unlabeled crops. This optimization significantly reduced the processing time by a factor of 14.8x ( an average of 0.58 seconds per image, compared to the original 4.28s which had to be repeated twice).

## A.2 Memory optimization

### A.2.1 Saving extracted data

The process of memory optimization within our pipeline involves the strategic extraction of voluminous data from frames, which is often requisite for subsequent stages of the pipeline. To mitigate the extensive RAM requirements posed by retaining the extracted data throughout the process, we devised a solution that involves the immediate serialization and storage of data from each specific pipeline step into a dictionary encapsulated within a pickle file. This approach not only facilitates the immediate deletion of data from RAM post-storage but also allows for the eventual re-loading of data into RAM when necessitated by future pipeline steps. Furthermore, this methodology affords a mechanism for determining whether a pipeline step has already been executed by verifying the presence of the extracted data within the pickle file. This robust design ensures continuity and resilience against interruptions and crashes by enabling the pipeline to bypass completed steps and resume operations from the point of interruption. Upon completion of the process, the data is consolidated into a single pandas dataframe, which is stored as an h5 file. This file format offers the advantage of environmental and system independence compared to pickle format, thus ensuring compatibility and accessibility across different systems, an essential feature given the operation of the GUI tool on a dedicated server application. Additionally, a JSON file is generated to encapsulate supplementary information that is not amenable to storage in a database format akin to pandas dataframes. All temporary files are subsequently deleted to optimize storage space.

### A.2.2 Optimization of loaded data

The extensive RAM requirements are not limited to the extracted data but also encompass the images and other data sources from which information is extracted. The initial strategy of loading images at full resolution proved impractical due to the prohibitive memory requirements, leading to the adoption of an approach where each image is downscaled such that the longest edge measures 256 pixels. This preliminary measure was refined upon further evaluation of the pipeline, wherein it was discerned that most models necessitated image resizing to specific resolutions, while only a minor subset could process images of any resolution. This insight prompted an experimentation phase with the latter models, which revealed that lower resolution images did not significantly impair the quality of the predicted output. Consequently, the CLIP model, specifically the ViT-L/14@336px version, was identified as the model that downscales images to the highest resolution by adjusting the shortest edge to 336 pixels. This procedure involves the removal of letterboxing from each image to restore the original video frame resolution as intended by the production company, followed by the identification and downscaling of the shortest edge to 336 pixels. This approach ensures minimal loss of input data when images are further downscaled by models requiring lower input resolutions, thereby preserving image detail and optimizing RAM usage without compromising performance.

By implementing these strategies, it is feasible to store in excess of 5000 images— a volume indicative of frames extracted from extended videos—within the confines of the 16GB RAM capacity provided by the chosen Microsoft Azure cluster. For cropped images, only the bounding boxes are retained in memory, with cropping operations conducted on-the-fly from the original images. This approach precludes the need to store multiple cropped images in RAM, effectively avoiding to quintuple the amount of data to be managed and thus optimizing both memory utilization and computation time.

In contrast, this resolution reduction method is limited to components of the pipeline that process the image in its entirety. Specifically, for segments focused on facial images, employing the strategy of cropping detected faces from previously downscaled images proves suboptimal. For instance, if the shortest edge of an image is reduced to 336 pixels, and a face occupies approximately 12% of the total image area, cropping at this stage results in a significantly low-resolution facial image. Although initial experiments adopting this approach yielded acceptable outcomes, the final pipeline was refined to reload each image at its original resolution before cropping faces, thereby maximizing facial image resolution. This adjustment is crucial, particularly for enhancing landmark detection accuracy, and it marginally benefits other processes involving facial images as well.

Alternative strategies, such as loading only specific image segments in batches, were considered. This method could be advantageous in scenarios where available RAM is insufficient for loading the complete image dataset. However, this approach necessitates repeated loading of models into memory for each batch, thereby extending processing time. In contrast, loading models a single time and applying batch-wise processing could mitigate RAM limitations, yet this was not adopted in our context due to sufficient RAM availability, thus prioritizing time efficiency.

### A.2.3 VRAM optimization

Optimization of VRAM, the memory utilized by GPUs, is another critical aspect. The most resource-intensive operation involves loading substantial models, such as Large Language Models (LLMs), for tasks like keyword extraction. Through quantization, we managed to accommodate the model within an 8GB VRAM constraint. Nevertheless, this issue was ultimately addressed by leveraging OpenAI's proprietary GPT 3.5 model, accessible via web API calls, thus circumventing the VRAM limitation. To further minimize VRAM usage, we ensured the exclusion of gradient calculations during model inference, promptly deallocating tensors when redundant, or transferring them from the GPU's VRAM to RAM. Although most models in our pipeline support batch processing of images—represented as multidimensional matrices—this approach significantly increases VRAM demands, often leading to pipeline instability. To enhance stability, we processed images individually, with limited exceptions such as shot detection, without incurring substantial time penalties.

## A.3 Time optimization

To enhance efficiency, batch processing emerges as a principal strategy. Despite the avoidance of batch processing in several stages due to potential memory overflow incidents leading to system failures, the selection of models integrated within our processing pipeline was deliberately aligned with their compatibility for batch processing. This alignment not only facilitates the incorporation of batch processing to expedite the workflow when the available hardware and VRAM permit but also signifies a commitment to optimizing time efficiency wherever feasible. Furthermore, numerous minor enhancements have been implemented to accelerate the process. Among these, the optimization of the automatic crop library, as delineated in the preceding section, stands out for its contribution to process acceleration. Notably, the pipeline segment most susceptible to time consumption is the downsampling process, followed by the autocrop functionality. The latter encompasses a variety of tasks, including but not limited to the computation of the logo score with saliency prediction. The temporal allocations for each segment of the pipeline are outlined in Table A.1. Notably, the most time-intensive stages are downsampling and face detection with autocropping, presenting opportunities for further optimization before deploying this tool extensively in a production environment.

Downsampling is currently optimized, leveraging C++ and parallelization techniques. However, there's potential to explore the parameter enabling video parsing at half the frames per second (fps), a feature introduced by our code. This adjustment, currently utilized primarily for longer videos, is expected to reduce both time and memory requirements. On the other hand, the autocropping step operates on a per-image basis. While adapting the code to process images in batches could offer a solution, initial experiments suggest potential challenges due to significant memory demands, which may exceed current hardware capabilities.

Overall, the total processing time varies significantly based on factors such as video length and the number of faces detected, as the absence of faces translates directly to reduced face processing requirements. Thumbnail generation hasn't been incorporated as it hinges on the quantity of thumbnails required, which is a value yet to be established in production stage. However, it's anticipated to be less time-consuming than the aforementioned downsampling and autocropping steps.

| Process | Percentage |
|---|---|
| 1. Downsampling | 36.806% |
| 2. Metadata retrieval | 0.038% |
| 3. Keyword extraction | 0.157% |
| 4. Letterboxing removal, keyframes loading and resizing | 7.448% |
| 5. Shot scale classification | 2.038% |
| 6. Face detection with autocropping, logo and on face focus scores | 25.953% |
| 7. Aesthetic prediction | 5.687% |
| 8. Semantic consistency prediction | 9.03% |
| 9. Cropping images on faces and position score computation | 3.578% |
| 10. Face embeddings computation | 0.682% |
| 11. Images and faces clustering | 0.418% |
| 12. Emotions detection | 0.289% |
| 13. Ethnicity, gender, age prediction | 6.642% |
| 14. Saving data to dataframe | 0.172% |
| 15. Automated selection | 1.062% |

Table A.1: Time percentage required to compute each step of the pipeline.

## A.4   Gradio Python library

Gradio [1] allows for the swift deployment of a web application, enabling us to concentrate on the tool's practicality and functionality while the management of server operations and backend, as well as the intuitiveness of the UI interface, are automatically handled. However, this ease and speed of implementation come at the cost of reduced flexibility for bespoke use cases, presenting challenges for the incorporation of advanced functionalities. For instance, Gradio does not facilitate the incorporation of images into buttons for multiple-choice checkbox options. Additionally, it lacks support for the dynamic creation of UI components, permitting only the dynamic update of their properties. This limitation can be circumvented by pre-instating more elements than necessary and rendering them visible as required, thereby exploiting this functionality to simulate the dynamic creation of such elements.

### A.4.1   Optimizations

Despite its limitations, the flexibility of the proposed system enables the development of a sufficiently complex prototype for daily use by the design team, which can be further customized in the future using more specialized, low-level libraries. The use of Gradio, supported by Microsoft Azure, facilitates deployment on their servers, allowing for straightforward implementation. The system architecture is designed such that the client runs a lightweight version of the code, with the majority of computational tasks being executed on the server. Initially, the pipeline was configured to store only the video frames as images in Play Suisse's cloud storage, obtaining cropped images by storing the coordinates of the bounding boxes. This approach would necessitate loading both the image and the crop coordinates into RAM, performing the cropping operation, and then transmitting the cropped image to the client. Such computational demands are deemed undesirable on the server side, given the need for servers with high computational and memory capabilities, which are costly. Consequently, the optimal solution involves storing all cropped images in storage and sending URLs to these images to the client for download. This method significantly reduces the server's computational load but increases the storage requirements for each asset, which is, nevertheless, more cost-effective for the company.

The computation of final scores has been optimized for temporal efficiency relative to the initial prototype, employing matrix operations to facilitate vector-based processing of all computations. This optimization enables real-time updating of scores based on filters selected in the GUI, thus minimizing the user's wait time to the duration required for image downloads upon initiating a search, which is typically brief. Transitioning from data related to one video to another is slightly slower, as it involves loading the video's database into the server's RAM, resetting the UI with default values, and facilitating the client's download of all images of automatic proposals, but not slower than a couple of seconds.